\documentclass[a4paper,twoside]{article}

\usepackage{amsmath}
\usepackage{amssymb}
\usepackage{amsthm}
\usepackage{appendix}
\usepackage{cite}
\usepackage[T1]{fontenc}
\usepackage[margin=3.5cm]{geometry}
\usepackage{graphicx}
\usepackage{indentfirst}
\usepackage[utf8]{inputenc}
\usepackage{lscape}
\usepackage{makecell}
\usepackage{multirow}
\usepackage{scalerel}
\usepackage{tabularx}
\usepackage{times}
\usepackage[hyphens]{url}
\usepackage{xltabular}

\newcommand{\q}[1]{``#1''}

\makeatletter
\def\th@plain{%
  \thm@notefont{}
  \itshape 
}
\def\th@definition{%
  \thm@notefont{}
  \normalfont 
}
\makeatother

\DeclareMathOperator*{\concat}{\scalerel*{\Vert}{\sum}}
\newtheorem{definition}{Definition}[section]

\begin{document}
\thispagestyle{empty}
\begin{center}
\includegraphics[width=\textwidth]{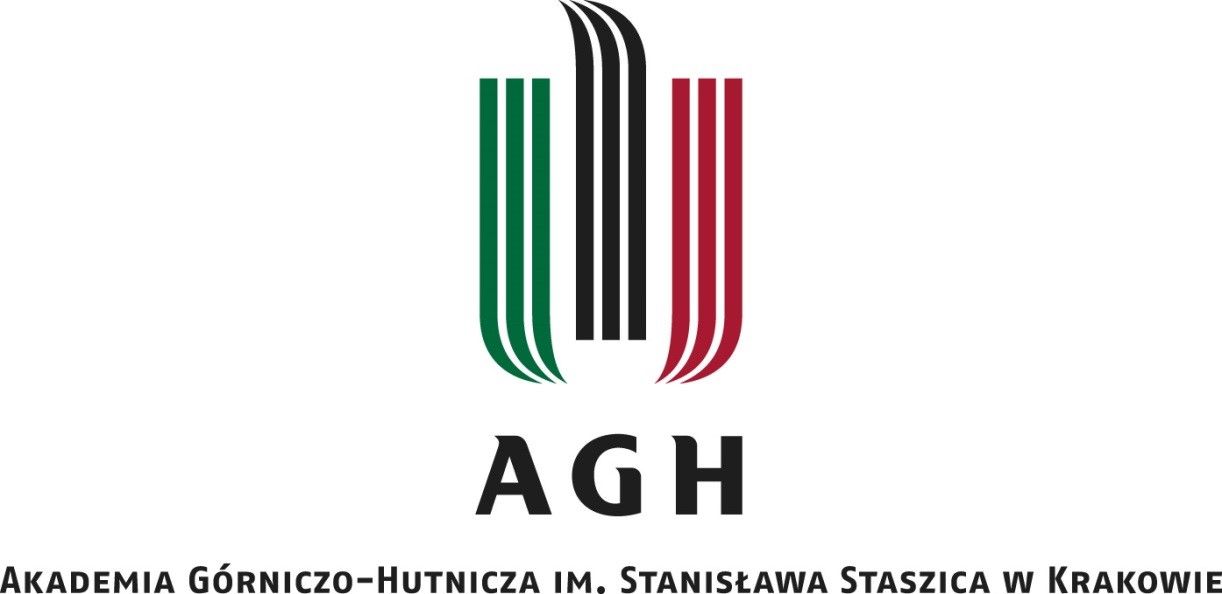}\\
\bf{FACULTY OF COMPUTER SCIENCE, ELECTRONICS AND TELECOMMUNICATIONS}\\[5mm]
\bf{INSTITUTE OF COMPUTER SCIENCE}\\[14mm]
\huge Master's Thesis\\[12mm] 

\Large Application of Graph Neural Networks and graph descriptors for graph classification\\[2mm]

\large Zastosowanie grafowych sieci neuronowych oraz deskryptorów grafowych w klasyfikacji grafów\\[2mm]

\end{center}

\vspace*{\fill}

\begin{tabular}{ll}
Author: & Jakub Adamczyk\\
Field of study: &	Computer Science - Data Science\\
Supervisor: & Wojciech Czech, PhD\\
\end{tabular}

\begin{center}
Krakow, 2022
\end{center}

\newpage\null\thispagestyle{empty}\newpage

\section*{Acknowledgements}
\thispagestyle{empty}

I would like to sincerely thank my supervisor, Wojciech Czech, for his guidance. I am thankful for his suggestions, including the proposed subject of this thesis. His hard work and vast knowledge of graph methods have been a great source of inspiration and motivation for my work.

I also want to thank AGH University of Science and Technology, and ACK Cyfronet in particular, for computational resources, which aided my research.

In particular, I would like to thank my grandparents and parents for teaching me discipline and diligence, owing to which I could finish my thesis.

\clearpage

\section*{Abstract}
\thispagestyle{empty}

Graph classification is an important area in both modern research and industry. Multiple applications, especially in chemistry and novel drug discovery, encourage rapid development of machine learning models in this area. To keep up with the pace of new research, proper experimental design, fair evaluation, and independent benchmarks are essential. Design of strong baselines is an indispensable element of such works.

In this thesis, we explore multiple approaches to graph classification. We focus on Graph Neural Networks (GNNs), which emerged as a de facto standard deep learning technique for graph representation learning. Classical approaches, such as graph descriptors and molecular fingerprints, are also addressed. We design fair evaluation experimental protocol and choose proper datasets collection. This allows us to perform numerous experiments and rigorously analyze modern approaches. We arrive to many conclusions, which shed new light on performance and quality of novel algorithms.

We investigate application of Jumping Knowledge GNN architecture to graph classification, which proves to be an efficient tool for improving base graph neural network architectures. Multiple improvements to baseline models are also proposed and experimentally verified, which constitutes an important contribution to the field of fair model comparison.

\clearpage

\newpage
\pagenumbering{gobble}
\setcounter{tocdepth}{3}
\tableofcontents
\addtocontents{toc}{\protect\thispagestyle{empty}}
\clearpage
\pagenumbering{arabic}

\section{Introduction}
\label{section_introduction}
\setcounter{page}{1}

Many practical applications require solving problems represented as tasks on graphs. They arise in many areas, such as theoretical computer science, bioinformatics, linguistics, and social sciences. Graph structure is a very abstract and general representation and, as such, can serve as a framework to formulate many distinct problems in a unified way. Graph theory, pioneered by Euler in the 18th century, has very strong mathematical foundations. A multitude of practical applications, combined with a strong theoretical background, resulted in the dynamic development of this branch of mathematics with many sub-disciplines. One of them, particularly interesting from a computational perspective, is graph comparison and matching, which aims to infer the properties of a graph and use them to answer questions such as:
\begin{itemize}
    \item How can we define the similarity between graphs?
    \item How can we represent graphs in metric spaces?
    \item How can we use graph representations combined with their similarity for predicting the properties of other graphs?
\end{itemize}
Tasks of this discipline can also be formulated in terms of machine learning, which has long traditions in computer science and computational learning theory. This resulted in the creation of graph representation learning, a field of study which aims to learn graph properties from the gathered data and use them for supervised or unsupervised learning tasks on graphs \cite{graph_representation_learning}. The tools used include graph descriptors combined with classification algorithms, graph kernels directly defining graph similarity, and graph neural networks.

\subsection{Motivation}
\label{section_introduction_motivation}

Most real-world graph datasets are far too large for manual analysis, and instead require scalable algorithms. They can be divided into three groups \cite{graph_representation_learning,gnn_survey_comprehensive_survey_wu,gnn_survey_review_of_methods_zhou}: node-level, edge-level, and graph-level. We can perform both supervised and unsupervised tasks on each level, e.g. node classification, node clustering, edge regression, graph classification. Graph-level tasks are arguably the most well studied in mathematics, as they use many graph features very well studied theoretically, e.g. algebraic connectivity \cite{descriptor_graph_algebraic_connectivity}, efficiency \cite{descriptor_graph_efficiency} or chemical indices \cite{descriptor_index_wiener_index,descriptor_graph_zagreb_index,descriptor_index_randic_index}. Graph kernel methods are also used at this level, as they describe whole graph structures to compare them to each other. There are many practical applications of this type of graph task in both supervised (when we know target labels) and unsupervised (we do not) settings:
\begin{enumerate}
    \item Graph classification: supervised learning, when the target is a class from a finite, unordered, discrete set. Applications include, e.g. predicting mutagenicity \cite{datasets_mutagenicity}, classifying molecules with activity against HIV, \cite{datasets_ogb} or detecting malicious software based on analysis of its control flow graph \cite{dataset_software_control_flow}.
    \item Graph regression: supervised learning, when a target label is a continuous number. Applications are, e.g. inferring solubility of molecules \cite{datasets_solubility} or predicting quantum properties of compounds, such as atomization energy \cite{datasets_qm9}.
    \item Graph clustering: unsupervised learning, when we aim to group graphs into a given number of clusters. The example application is clustering substrate-reaction metabolic networks for constructing phylogenetic trees \cite{czech_doktorat}.
\end{enumerate}

Out of those three groups, graph classification and regression are more popular. It is because a plethora of problems from bioinformatics and related domains (pharmacology, molecular biology, biochemistry, quantum physics, and chemistry) are naturally expressed as supervised learning on molecular graphs. Those are often binary classification (with two classes, whether a molecule has or does not have a given property) or regression (magnitude of physical property, e.g. solubility). There are multiple datasets available in this area, and the interest in the scientific community as well as in the private sector is growing. In silico experiments using machine learning-based predictions are becoming an important tool, e.g. in novel drug discovery \cite{bayer_admet}.

Most works on machine learning on graphs focus on undirected, unweighted, homogenous simple graphs, possibly with attributed nodes (vertices), and do not consider edge attributes, as this type of graphs is the most common in general. Directed graphs have asymmetric adjacency matrices, which require dedicated GNN architectures to handle them, as it results in undesirable numerical properties, such as potentially complex eigenvalues. Heterogeneous graphs (with nodes with different sets of attributes) can be represented as homogenous graphs with high-dimensional node feature vectors, or we can work on them directly with dedicated GNN architectures. They are much more common in the context of recommendation systems using graph-based approach, where users and items are distinct types of nodes. Edge attribute information can be incorporated with Levi graph transformation \cite{handbook_of_graph_theory,levi_transform} and regular GNN architectures can be applied then, or we can use dedicated architectures. Multigraphs (graphs with more than 1 edge between any pair of nodes) are much less frequent than simple graphs and not applicable in many areas (e.g. in molecular structures). However, many of the ideas in this thesis apply to graphs of any type.

Traditional machine learning methods, including deep learning, have been used for structured data, such as tabular datasets, audio, or images. This data has a well-defined metric space, which makes them easier to analyze with both manual feature engineering and deep learning. Graphs are combinatorial structures and, as such, require dedicated algorithms and feature engineering. However, all approaches work under the hood in the same way: they extract a graph embedding, i.e. a vector of typically continuous numbers, which represents a graph as a point in some high-dimensional metric space. Given this representation, the problem is just a tabular classification. Therefore, the most common approaches really differ in the way they create the graph embedding.

The traditional approach calculates features selected by the researcher and uses them as a direct input for the classifier. Another common name for those features is graph invariants, as the vast majority of them depend solely on the structure of the graph, not on any particular labeling or ordering of the nodes or edges in the graph, or their permutation (they are isomorphism invariant). They are most often based either on graph theory (e.g. node centrality, clustering coefficient), or algebraic graph descriptors \cite{czech_graph_investigator,graph_representation_learning}) or on domain knowledge, e.g. using molecular fingerprints in chemistry (e.g. Extended Connectivity Fingerprints (ECFP) \cite{fingerprints_ECFP}). They can describe both the local and global structure of the graph, but typically they require domain knowledge and choosing the right set of descriptors is not straightforward \cite{fingerprints_descriptor_spaces_comparison}. Some of them are learned from data, e.g. ECFP, and some are predetermined by expert knowledge, e.g. MACCS keys \cite{fingerprints_MACCS}.

Historically, graph kernels \cite{graph_kernel_history} were often used to capture the similarity between graphs using their structural properties (e.g. using random walks or subtrees \cite{graph_kernels}, or isomorphism testing heuristics \cite{wl_kernel}). They are an example of shallow learning, as they learn from data, but do not learn representations in an end-to-end fashion, like deep learning. While they are highly expressive, they come at a high computational cost, both in terms of their calculation (typically at least $O(n^3)$) and later use for similarity search (at least $O(n^2)$), making their usage infeasible for large datasets.

Deep learning applied to non-euclidean domains, known as geometric deep learning \cite{geometric_deep_learning_2016,geometric_deep_learning_2021}, has seen a surge in popularity in recent years. A large part of that revolution is graph neural networks (GNNs), designed specifically to learn node and graph embeddings in an end-to-end fashion. An important feature of those architectures is preserving input invariance, i.e. the order of nodes does not matter, which was not the case for traditional deep learning models \cite{gnn_scarselli_2009,gnn_survey_comprehensive_survey_wu}. The key advantages of GNNs are relatively low computational cost (compared to graph kernels) and end-to-end training (learning from data for efficient graph representation tailored to a given task). However, they are based on the message-passing paradigm \cite{gnn_scarselli_2009,graph_representation_learning,datasets_qm9}, in which neighboring nodes exchange information. This means that capturing more global features and explicit structural graph features is problematic, requiring sophisticated architectures, deep networks, and solving problems arising from them (e.g. exploding/vanishing gradients, overfitting, oversmoothing \cite{oversmoothing,oversmoothing_2}).

\subsection{Research goals}
\label{section_introduction_research_goals}

The aim of this work is to investigate techniques for graph classification, on tasks requiring capturing both local and global features of graphs. To this end, we focus on modern, general-purpose GNN architectures (not limited to any particular domain), and on graph descriptors, such as molecular fingerprints. Those approaches extract both local and global features, exploiting structural graph knowledge as well as available node features. The specific goals of the research project presented in this thesis are as follows:

\begin{itemize}
    \item Investigate GNN architectures for graph classification and their building blocks, and how the architectural choices (e.g. depth, sophistication, information aggregation strategies) affect the quality of models.
    \item Compare the efficiency of GNNs and graph descriptor-based models, especially on tasks requiring extraction of hierarchical information, such as molecular property prediction.
    \item Test solutions on a varied collection of state-of-the-art benchmark datasets, and compare models in a unified way, following experimental procedures designed with recommended fair evaluation and fair comparison techniques.
\end{itemize}

\subsection{Contributions}
\label{section_introduction_contributions}

Our contributions can be summarized as follows:
\begin{enumerate}
    \item We designed and created a fair evaluation benchmarking environment for GNNs. A collection of challenging datasets with varied properties has also been selected.
    \item We performed numerous experiments with GNN architectures, obtaining one of the largest sets of results in the literature. Furthermore, we analyzed those results from different perspectives, giving many insights into GNN architectures.
    \item In addition to GNN architectures, as a part of fair comparison, we propose 4 new baselines for graph classification:
    \begin{itemize}
        \item Simple Graph Convolution (SGC), applied to this task for the first time
        \item Local Degree Profile (LDP) with additional graph descriptors
        \item Molecular Fingerprints (MFPs) and Deep Multisets, combined with node features embedding
    \end{itemize}
    \item We performed experiments with graph descriptor-based LDP model and with molecular fingerprints, showing that the latter outperform GNNs in most cases. We also proposed using a new model, based on concatenation of multiple fingerprints, obtaining a more robust classifier.
\end{enumerate}

\subsection{Thesis overview}
\label{section_introduction_thesis_overview}

This thesis is organized in the following way. Firstly, in Section \ref{section_introduction_notation} we introduce the notation and definitions from graph theory used throughout this text.

Further, in Section \ref{section_literature} we present the review of the literature relevant to the thesis. In particular, it is divided into three subsections, following the main research goals of the work. In Section \ref{section_literature_gnn} we review GNN architectures, their particular building blocks, and strategies for using them for graph classification. In Section \ref{section_literature_descriptors} we describe the available graph descriptors from various domains. Finally, in Section \ref{section_literature_evaluation} we introduce evaluation strategies for ML in graph classification.

Section \ref{section_theory} presents the detailed theoretical background of the methods relevant to performed experiments. In particular, Section \ref{section_theory_gnns} focuses on building blocks of GNN architectures, which are explained and compared based on their theoretical properties and the capability of capturing graph properties. In Section \ref{section_theory_descriptors} we describe graph descriptors, algorithms for their calculation, their interpretation, strengths, and weaknesses. We outline the details of evaluation methodologies, good practices of fair evaluation, and their applications to the graph domain in Section \ref{section_theory_evaluation}.

In Section \ref{section_experiments} we provide descriptions of experiments preparation, results, and their discussion. Experimental setup, including architectures used, datasets, hyperparameter tuning, and evaluation methodology are in Section \ref{section_experiments_setup}. We present results and discuss them in Section \ref{section_experiments_results_and_discussion}.

The thesis summary, drawn conclusions, possible future work, and final remarks are presented in Section \ref{section_summary}.

\subsection{Notation and definitions}
\label{section_introduction_notation}

In this section, we recall the basic mathematical notation and definitions used further in this work. Like most works on GNNs, we consider primarily undirected, unweighted, homogenous simple graphs, possibly with attributed vertices (nodes). We do not consider edge attributes for GNNs, since they require graph transforms or dedicated architectures; however, graph descriptors may make use of this information. We do not consider directed or heterogeneous graphs or multigraphs, as they require dedicated GNN architectures and the problems requiring them are more specific and less popular. However, most of the ideas in this thesis also apply to them. Because of this, we will assume in the rest of this thesis that the graphs possess appropriate properties of undirected, unweighted, homogenous simple graphs, e.g. symmetric adjacency matrix.

First, we present the mathematical notation used. The special sets of numbers are represented with a blackboard bold font, e.g. $\mathbb{R}$ is a set of real numbers.

Vectors are represented with lowercase Latin or Greek alphabet letters. Matrices and tensors are written as uppercase letters. For example: $x$, $A$, $\Theta$. Scalars at particular indices are denoted as $x_1$, $A_{1,2}$. This notation is also often used to denote different vectors/matrices, e.g. $G_A$ vs. $G_B$, however, the meaning should be always obvious from the context in those cases. For notational convenience, commas in indices will be sometimes omitted, e.g. $A_{ij}$. In the case of matrices, if the entire dimension is selected, a colon is used, e.g. $A_{i,:}$ means selecting the entire $i$-th row. Vectors are indexed with integers from 1 to $n$, and matrices are indexed similarly and are of shape $(m, n)$ ($m$ rows and $n$ columns), unless otherwise specified. Tensors are indexed similarly to matrices. Shapes of vectors, matrices and tensors are written in upper index as an abbreviation, e.g. $x^n \Leftrightarrow x \in \mathbb{R}$, $A^{m,n} \Leftrightarrow A \in \mathbb{R}^{m \times n}$. Series of matrices (e.g. weight matrices in neural network layers) are marked with index in parentheses in upper index, e.g. $W^{(i)}$, $H^{(i + 1)}$. The identity matrix is denoted as $I$. A diagonal matrix with vector $x$ on diagonal is marked as $diag(x)$.

The dot product between vectors is marked as $\cdot$, e.g. $z = x \cdot y$. Scalar, matrix, or tensor multiplication is marked as two letters next to each other, e.g. $C = AB$. Elementwise vector or matrix multiplication, also known as Hadamard product, is denoted with $\odot$, e.g. $C = A \odot B$. Matrix multiplication with itself $k$ times, also known as matrix $k$-th power, is written as $A^k$, e.g. $AAA = A^3$. Matrix or vector concatenation is marked $\concat$ operator, with vectors/matrices being concatenated in squared brackets, e.g. $c = \left[ a \concat b \right]$. The concatenation dimension for matrices should be obvious from the context, and most often is the horizontal dimension (increasing the number of columns).

Functions are marked with lowercase Latin or Greek alphabet letters, e.g. $f(x)$. It is assumed that arguments are vectors, unless otherwise specified. In particular, $\sigma(x)$ denotes the sigmoid function, not just any function in general, like in some works on GNNs:
$$\sigma(x) = \frac{1}{1 + e^{-x}}$$

\noindent Below, we present the basic graph theory definitions and notation.

\begin{definition}[Undirected graph]
An undirected graph $G$ is defined as an ordered pair $G = (V_G, E_G)$, where $V_G$ is a set of vertices and $E_G$ is a set of edges. An edge $e_{ub} = \{ u, v \} \in E_G$ is an unordered pair of vertices ($u \in V_G$, $v \in V_G$). Two vertices $u$ and $v$ are adjacent ($u \sim v$) if they are joined by an edge.
\end{definition}

\noindent If the context of a graph is obvious, the indices are dropped for all definitions concerning it, e.g. $G = (V, E)$. Edges are written as $(u, v)$ and it is assumed that the order of vertices in the pair does not matter. Vertices are also called nodes, which is a more standard naming in GNN literature. Graphs are interchangeably called networks. Following standard notation from GNN literature, $|V|$ will be also written as $n$ if the context is obvious. For example, we use $O(n^2)$ instead of $O(|V|^2)$, as typically the graph is sparse, i.e. $|V| >> |E|$, and the complexity bound depends only on $|V|$, while in computer science the main complexity variable is traditionally denoted as $n$.

\begin{definition}[Simple graph, multigraph]
A graph is simple if there is at most one edge between any two vertices $u$ and $v$. Otherwise, if there is at least one pair of vertices $u$ and $v$ connected with two or more edges, the graph is a multigraph.
\end{definition}

\begin{definition}[Attributed graph]
A graph is attributed if its vertices or edges have feature vectors (attributes) attached to them. Elements of those vectors are typically numerical, either real numbers or integers. Graphs with all vectors with the same attributes are homogenous, while graphs with at least two distinct types of feature vectors are heterogeneous.
\end{definition}

\begin{definition}[Walk]
A walk of length $k$ from vertex $u$ to vertex $v$ is a sequence of $k$ edges connecting $u$ and $c$. The closed walk is a walk where $u = v$.
\end{definition}

\begin{definition}[Path]
A path $p_G(u, v)$ between vertices $u$ and $v$ is a walk without repeated vertices.
\end{definition}

\begin{definition}[Vertices distance]
The distance $dist_G(u, v)$ between vertices $u$ and $v$ is a length of the shortest path between $u$ and $v$. If a path does not exist, then $dist_G(u, v) = \infty$.
\end{definition}

\begin{definition}[Vertex neighborhood]
The (direct) neighborhood $N(v)$ of vertex $v$ is a set of vertices adjacent to $v$. $k$-th order ($k$-hop) neighborhood $N^k(v)$ of a vertex $v$ is a set of vertices of distance to $v$ lesser or equal to $k$.
\end{definition}

\begin{definition}[Vertex degree]
The degree $d(v)$ of vertex $v$ is the number of edges incident to $v$, equal to the size of $N(v)$ for simple graphs.
\end{definition}

\begin{definition}[Permutation invariance]
A function $f$ of a vector argument $x$ is permutation invariant if the value of $f$ does not change when the order of vector elements changes.
\end{definition}

\noindent Unordered sets, such as vertices and edges, may be represented with indexable structures, e.g. with vectors or matrices. However, they are still permutation invariant. Only for implementational purposes, the indexing of nodes is fixed after loading the graph as a matrix.

\begin{definition}[Graph isomorphism]
Let $G$ and $H$ be simple graphs. A graph isomorphism between $G$ and $H$ is a bijection: $\alpha: V_G \rightarrow V_H$ such that:
$$(u, v) \in E_G \Leftrightarrow (\alpha(u), \alpha(v)) \in E_H$$

\noindent Isomorphic graphs $G$ and $H$ are denoted as $G \simeq H$. Isomorphism preserves the structural information of the graph represented by its edges. It constitutes an equivalence relation on graphs that partitions the set of graphs into equivalence classes.
\end{definition}

\begin{definition}[Adjacency matrix]
Let $G$ be a simple graph. Adjacency matrix $A_G$ is a matrix for which

$$A_{u, v} =
\begin{cases}
1, & \mbox{if } {u, v} \in E_G \\ 
0, & \mbox{if } {u, v} \notin E_G \\ 
\end{cases}$$

\noindent Adjacency matrix is symmetric for undirected graphs. For graphs without self-loops, the diagonal entries are zeros. For weighted graphs, a non-zero entry $A_G(u, v)$ represents the weight of the edge between $u$ and $v$. Those matrices are typically (very) sparse.
\end{definition}

\begin{definition}[Degree matrix]
For a simple graph $G = (V, E)$ with $n$ vertices, the degree matrix $D \in \mathbb{R}^{n \times n}$ is a diagonal matrix:
$$D_{ii} = d(v_i)$$
\end{definition}

\begin{definition}[Laplace matrix]
For a simple undirected graph $G = (V, E)$ with $n$ vertices, its Laplace matrix (Laplacian) $L \in \mathbb{R}^{n \times n}$ is defined as:
$$L = D - A$$

\noindent The elements are defined as:
$$L_{ij} =
\begin{cases}
d(v_i), & \mbox{if } i = j \\ 
-1, & \mbox{if } (v_i, v_j) \in E \mbox{ and } i \neq j \\ 
0, & \mbox{otherwise}
\end{cases}$$

\noindent $L$ is a positive semidefinite matrix. It is (very) sparse for sparse graphs, which is often the case, as are all other Laplacian matrices (see below).
\end{definition}

\begin{definition}[Symmetric normalized Laplacian]
The symmetric normalized Laplacian $L^{sym} \in \mathbb{R}^{n \times n}$ is defined as:
$$L^{sym} = D^{-\frac{1}{2}} L D^{-\frac{1}{2}} = I - D^{-\frac{1}{2}} A D^{-\frac{1}{2}}$$

\noindent The elements are defined as:
$$L^{sym}_ij =
\begin{cases}
1, & \mbox{if } i = j \mbox{ and } d(v_i) \neq 0 \\ 
-\frac{1}{\sqrt{d(v_i)d(v_j)}}, & \mbox{if } (v_i, v_j) \in E \mbox{ and } i \neq j \\ 
0, & \mbox{otherwise}
\end{cases}$$

\noindent $L^{sym}$ is a positive semidefinite matrix.
\end{definition}

\begin{definition}[Vertex descriptor]
A vertex descriptor is a scalar or vector attached to the given element, calculated using its structural properties (e.g. neighbors) or attributes vector.
\end{definition}

\begin{definition}[Graph descriptor]
A graph descriptor (graph invariant, graph feature) is a scalar or vector attached to the graph. It is calculated using some properties of the graph (e.g. its structure) or using a distribution of vertex (node) descriptors (e.g. histogram of values of a  particular descriptor for all vertices of the graph).
\end{definition}

\begin{definition}[Graph embedding]
A graph embedding is a vector attached to the graph, representing it as a point in a metric space.
\end{definition}

\clearpage

\section{Literature review}
\label{section_literature}

\noindent In this chapter, we present an overview of the literature relevant to the thesis. We review works on GNN, graph descriptors, and the evaluation methodologies for ML models.

\subsection{Graph neural networks}
\label{section_literature_gnn}

This section describes GNN architectures. First, we provide a historical perspective and works describing the general framework of GNNs. Then, the taxonomy of different architectures used in the literature is reviewed. Later, we present the related works on convolutional and attentional architectures, as well as on readout layer.

\subsubsection{Introductory works and general architecture}
\label{section_literature_gnn_introductory_works}

Applying neural networks to graph data requires the models to be able to work with graph matrix representation (e.g. adjacency matrix). It also requires the model to be invariant to the input permutation, so that the nodes of the graph can be seen by the network in any order. Most GNN models aim to learn a node embedding for each node, which (possibly after many layers, which may also be of many types) are aggregated for the graph classification as a graph embedding. Due to this formulation, most works on GNNs focus on node classification, but those architectures can be easily applied to graph classification, as it just requires one additional step of node embedding aggregation and classifying the graph embedding, typically with a multilayer perceptron (MLP).

The first works for applying neural networks to graphs and introducing the GNN concept are \cite{gnn_scarselli_2004,gnn_scarselli_gori_2005,gnn_scarselli_2009}. Those works are a series by the same authors, with the most mature model presented in \cite{gnn_scarselli_2009}, which is sometimes called vanilla GNN in literature. This model is limited to undirected homogenous graphs, but can use both node and edge features. It introduces the concept of learning node embeddings through message passing, where each node exchanges its information (messages) with its neighbors. Information from neighbors is aggregated and used to update the node embedding, and this process is repeated multiple times, extending the radius of information exchange for each node. First, the authors propose first using a local transition function to learn a node embedding, and then using a local output function to learn an output state of the node using the input node features and its embedding. Both functions are parametrized and are MLPs, and their parameters are learned through backpropagation. This is also called an encoding network, as this encodes node states and is a recurrent neural network (RNN). Therefore, this type of network is an example of RecGNN (Recurrent Graph Neural Network) \cite{gnn_survey_comprehensive_survey_wu}. Computing the embedding of all nodes is done through an iterative scheme with Banach's fixed point theorem. While this model allows learning with graph data, it has significant limitations. The representation is quite limited, as it does not learn hierarchical representations. Moreover, the learning scheme is inefficient, requiring both obtaining a fixed point solution and backpropagation in each iteration.

The general framework of neural message passing for graphs presented in \cite{gnn_scarselli_2009} has been generalized in \cite{datasets_qm9} and is known as a message-passing paradigm \cite{introduction_to_graph_neural_networks} or as a Message-Passing Neural Network (MPNN) model (also called neighborhood aggregation strategy \cite{gnn_GIN}). It is used by a vast majority of GNNs. The single layer model in this framework represents message passing between a node and its neighborhood, therefore in $k$ layer GNN each node captures the information of its $k$-hop neighborhood. Formally, GNNs are parametrized with two functions reflecting how the node neighborhood function is aggregated ($a_v = AGGREGATE(N(v))$) and how this information is used, along with its embedding, to obtain the new, updated embedding ($h_v = COMBINE(a_v, h_v^{prev}$). Additionally, for graph classification \cite{datasets_qm9} a layer with $h_G = READOUT(H_v)$ function is used (often called global pooling layer \cite{introduction_to_graph_neural_networks,gnn_survey_comprehensive_survey_wu,pytorch_geometric} or graph-level pooling layer \cite{gnn_GIN}), which aggregates the node embeddings matrix and outputs the graph embedding vector. Different GNN architectures can therefore be defined in terms of choice of those functions and their properties. For example, the difference between families of spatial convolutional and attentional architectures lies in a different weighting of neighbors in $AGGREGATE$ function. Such layers are known as message-passing layers \cite{pytorch_geometric} or computational/propagation modules \cite{gnn_survey_review_of_methods_zhou}. In GNN literature, layers are also often called modules, blocks, operators or message-passing functions.

\subsubsection{GNN taxonomy}
\label{section_literature_gnn_taxonomy}

In recent years, numerous surveys on GNNs have been conducted. Most of them propose a particular taxonomy, since the classification of architectures into a few distinct families makes the comparison of technical details easier.

In \cite{geometric_deep_learning_2016} convolutional GNNs are reviewed and divided into two groups, depending on the approach to deriving and implementing the convolution operator: spatial and spectral. The work 
\cite{gnn_survey_representation_learning_on_graphs} contains an overview of many graph representation learning approaches, including traditional \q{shallow} approaches, but also features GNNs, divided into two groups: neighborhood aggregation (convolutional) and message-passing (vanilla GNN, MPNN). 

GNNs can also be decomposed into basic building blocks, as proposed in \cite{gnn_survey_inductive_biases}. GNN architectures are analyzed as compositions of 6 basic building blocks, called Graph Networks (GNs), and proposes classifying the GNN architectures in this framework.

The authors of \cite{gnn_survey_review_of_methods_zhou} observe that GNNs can be classified on many axes: working on directed vs. undirected graphs, homogenous vs. heterogeneous graphs, static vs. dynamic (changing in time) graphs, and the choice of computational modules, which is further divided into computational modules, sampling and pooling methods. General GNN architectures are divided by message passing architecture into two main groups: convolutional and recurrent. Convolutional aggregation is further defined as either spectral or spatial, where spatial convolution can be either \q{basic} convolution or attentional. Recurrent operators can be convergence-based or gate-based (using a gating mechanism known from, e.g. Long-Short Term Memory (LSTM) \cite{lstm} or Gated Recurrent Unit (GRU) \cite{gru} networks).

Alternatively, as described in the work \cite{gnn_survey_comprehensive_survey_wu}, GNNs are divided into four categories: recurrent (RecGNNs), convolutional (ConvGNNs), graph autoencoders (GAEs), and spatial-temporal (STGNNs). Convolutional architectures are classified as either spatial or spectral. Basic attention-based architectures are treated as a variant of spatial convolutional architecture, but it is noted that there are interesting attentional GNNs that do not fit this classification. GAEs are divided by their application, as either used to learn network embeddings or for graph generation, and are built in practice using either recurrent or convolutional blocks.

The work \cite{gnn_survey_deep_learning_on_graphs} provides an overview of deep learning on graphs, but focuses on GNNs specifically. It categorizes deep learning methods on graphs into five categories: graph recurrent neural networks (Graph RNNs), graph convolutional neural networks (GCNs), graph autoencoders (GAEs), graph reinforcement learning (Graph RL), and graph adversarial methods. Recurrent architectures are divided into node-level or graph-level, based on the task they were created to perform (e.g. node classification vs. graph generation). Convolutional layers are further specified as either spatial or spectral and are also differentiated by the convolution filter type (e.g. first-order vs. polynomial), readout mechanism, or computational complexity. Attention-based modules are included in the convolutional approach, but are detailed separately as an improvement.

While \cite{gnn_survey_neural_symbolic_computing} focuses on exploring the relationship between GNNs and Neural-Symbolic Computing (NSC), authors in their analysis divide GNNs into three groups: convolutional, message-passing (vanilla GNN, MPNN), and attentional (using attention mechanism).

In the work \cite{geometric_deep_learning_2021}, not only a division of GNNs into groups is proposed, but the authors also organize them into a strict hierarchy: convolutional, attentional, and message-passing. Each group is a subset of the previous one, however, the authors specify that using a more general group of GNNs is not always advantageous, as simpler architectures are naturally regularized (and therefore prevent overfitting). 

The authors of \cite{gnn_ml_on_graphs_taxonomy} propose a taxonomy for graph encoding methods: shallow embedding, graph regularization, graph auto-encoding, and neighborhood aggregation (GNNs). GNNs are further classified as either global filtering (spectral) or using local message-passing, which can be general frameworks, spectrum-free methods, or spatial. Graph convolutional methods are categorized separately as spectrum-based, spectrum-free, spatial, or attentional. Spectrum-based methods compute the full graph spectrum and spectrum-free methods rely on its approximation, typically using some polynomial base.

In turn, the goal of \cite{gnn_survey_gnn_in_NLP} is applying GNNs to graphs for natural language processing (NLP). The authors present different GNNs as different implementations of graph filter learning and divide them into four groups: spectral convolutional, spatial convolutional, attentional, and recurrent. Pooling layers are also considered and divided into global and hierarchical.

As the focus of this work is graph classification, layered feedforward architectures are of the most interest, as they can easily learn using supervised learning with labels for the whole graphs. While GAEs can be used for graph classification, they do not learn end-to-end for this specific task (since they optimize the encoding error during training, not misclassification error), therefore their performance is expected to be worse, and for this reason, they will not be considered. We will analyze the following categorization of architectures, inspired by the most of the works described above: recurrent GNNs (RecGNNs), convolutional GNNs (ConvGNNs), attentional GNNs (AttGNNs); ConvGNNs can be either spectral or spatial. We will focus on convolutional and attentional approaches. Recurrent architectures are mostly used in the pioneering works on GNNs, have high computational costs, and in general have been outperformed by convolutional and attentive architectures for node and graph classification \cite{gnn_survey_comprehensive_survey_wu,gnn_survey_review_of_methods_zhou}, therefore they will not be considered further. We treat the attentional architectures as a separate group, next to the convolutional layers, as the attention mechanism is a sophisticated weighting scheme and many recent works focus on it specifically \cite{attentional_gat,attentional_gat_v_2,attentional_gaan,attentional_agnn,attentional_understanding_attention}. However, it should be stressed that attentional architectures also use graph convolution, i.e. they simply implement attention-based graph convolution.

\subsubsection{Convolutional architectures}
\label{section_literature_gnn_convolutional}

Convolutional GNN (ConvGNNs) architectures have been the most popular ones in recent years, as they are efficient, convenient to compose, and their similarities to CNNs can be exploited \cite{gnn_survey_comprehensive_survey_wu}. They are divided into spectral-based or spatial-based, depending on how the graph convolution is defined, as for graphs the definition is not as obvious as for images. Spectral approaches stem from graph signal processing, where the convolution operation is interpreted as smoothing the graph signal, i.e. removing the noise from it \cite{gnn_graph_signal_processing}. It operates on the eigendecomposition of the Laplacian and uses graph Fourier transform to transform the signal, convolving it in the spectral (frequency) domain with learnable filters. Spectral approaches are sometimes considered more suited for the transductive setting, where the test set is known and consists of a set number of cases; this is a common case for node classification, where a given subset of nodes in each graph has unknown labels. Spatial ConvGNNs are more similar to CNNs, as they define graph convolutions based on a node's spatial relations, which can be also seen as direct information propagation with vertex neighborhood \cite{gnn_survey_comprehensive_survey_wu}. While this approach aims to have both lower computational cost (not requiring the eigendecomposition) and better generalization properties (not depending on a particular eigenbasis), defining the convolution for differently sized neighborhoods and maintaining the local invariance property is not straightforward. They aim to perform well in inductive settings, i.e. to learn general rules and apply them to test data having the same distribution as the training data.

One of the first works on ConvGNNs is \cite{convgnn_bruna}, which introduced the Spectral Network architecture and basic ideas of spectral analysis of graphs: Laplacian eigendecomposition, treating convolution as signal smoothing, and learnable localized filters with multiple channels. A spatial interpretation is also provided. This pioneering work has four major limitations \cite{gnn_survey_comprehensive_survey_wu,introduction_to_graph_neural_networks}, all related to computing the exact eigendecomposition of graphs. Firstly, any change to the graph changes the eigenbasis, which heavily affects stability and therefore the generalization properties. This can also be interpreted as inherent overfitting. The learned filters are domain-dependent, so transfer learning similar to CNNs is not possible. Additionally, the filters may be non-spatially localized. Lastly, the eigendecomposition itself is computationally costly, having $O(n^3)$ complexity.

Further works on spectral ConvGNNs focused on simplifying the ideas from \cite{convgnn_bruna}. The approximations both reduce the computational complexity and regularize the model. The work \cite{convgnn_hammond} introduced the general idea of approximating convolutional filters with truncated expansion using $K$-th order Chebyshev polynomial basis. ChebNet \cite{convgnn_chebnet} is an implementation of this idea for ConvGNNs. This convolution has the advantage of being $K$-localized, since it is a $K$-th order polynomial in the Laplacian, therefore the filters can extract local features independently of graph size \cite{gnn_survey_comprehensive_survey_wu}. It also does not require computing the eigenvectors of the Laplacian, only the largest eigenvalue, which greatly reduces the computational load. Additionally, pooling is used, employing the Graclus algorithm \cite{pooling_graclus} for graph coarsening and node ordering algorithm to transform the graph to a 1D signal. Another parametrization, with Cayley polynomials (which are parametric rational complex functions), which aims to capture narrow frequency bands in the spectral domain, is presented in \cite{convgnn_cayleynet}. An even stronger simplification is presented in well-known work \cite{convgnn_gcn} as Graph Convolutional Network (GCN). This architecture limits the layer-wise convolution to just $K = 1$, since previous architectures had the tendency to overfit on local neighborhood structures for graphs with wide degree distributions \cite{introduction_to_graph_neural_networks}. For this reason, this first-order approximation became a popular technique for ConvGNNs. Additionally, the largest eigenvalue is approximated as $\lambda_{max} = 2$ and free parameters are reduced to just one vector. Those vectors are composed into a matrix to allow multichannel input and output. Numerical instability (vanishing/exploding gradients) of deep multilayer GNNs is also addressed through the introduction of \q{renormalization trick}, in which self-loops are added to the graph (i.e. identity matrix is added to adjacency matrix). The resulting model is well regularized, localized, and greatly simplified. It can be also treated as a spatial method \cite{introduction_to_graph_neural_networks,gnn_survey_comprehensive_survey_wu}, which in each layer aggregates the information from the direct node neighborhood.

Further works on spectral ConvGNNs mostly focused on expanding the structure learned in the convolution layer. While all previous models use the original graph structure to propagate information, Adaptive GCN (AGCN) \cite{convgnn_agcn} additionally learns the underlying implicit relations that may exist in the graph. Additional \q{residual} edges added this way shorten the paths in the graph between nodes important to each other, increasing the flow of information between them. To this end, AGCN computes \q{residual} the Laplacian matrix using the learned adjacency matrix, which is in turn computed via a learned metric. Mahalanobis distance and Gaussian kernel are used to compare nodes and, as a very generic framework, create a large space of learnable solutions. However, this results in a dense adjacency matrix, which greatly increases the computational cost of processing all layers in the graph.

Spatial approaches define convolutions directly using the topological structure of the graph, and have been largely developed in parallel to recurrent and convolutional spectral approaches. Neural Network for Graphs (NN4G) \cite{convgnn_nn4g}, proposed in parallel with vanilla GNN, learns graph dependencies through a compositional layered architecture, where the convolution directly sums up a node's neighborhood information. Residual and skip connections are used to use information from different layers. The downside of this early approach is the usage of an unnormalized adjacency matrix (which may cause stability problems). Neural Fingerprints (Neural FPs) \cite{convgnn_neural_fp} architecture takes a similar approach, as it also sums up the embeddings of the node and its entire neighborhood. However, it also multiplies that sum with a weight matrix, where each node degree has a separate matrix. Unfortunately, this parametrization makes it infeasible to apply this model for large-scale graphs with many node degrees.

Diffusion Convolutional Neural Network (DCNN) \cite{convgnn_dcnn} interprets graph convolution as a diffusion process, where information is transferred between nodes with a certain transition probability, so that information distribution can reach an equilibrium. It is efficiently implemented as a probability tensor power series, where the probability transition matrix is computed using degree and adjacency matrices, so the graph topological structure drives the probability distribution. Hidden representation (embedding) matrices from each diffusion step are concatenated as the final model outputs. For graph classification, the average across nodes is taken, and the averaged vectors are concatenated instead. Diffusion Graph Convolution (DGC) \cite{convgnn_dgc} takes a similar approach, but sums up the outputs of each diffusion step instead of concatenating them. In both of those models, the power of a transition probability matrix is used, therefore very little information is contributed by distant neighbors. While this may be beneficial for applications where direct neighbors are exponentially more important (e.g. social networks), this means that more global structure is not considered, which in turn may be harmful in applications where long dependency chains exist (e.g. biology).

Some spatial convolutional models use the neighborhood sampling approach, where a subset of $k$ neighboring nodes is taken into consideration as a convolutional patch. In those approaches node ordering is often introduced, where nodes (either all from the graph or in each neighborhood separately) are ranked by their importance, and only a subset of fixed size $k$ is used.

PATCHY-SAN \cite{convgnn_patchy_san} takes a 4-step approach: node selection, neighborhood assembly, graph normalization, and convolution with traditional CNN architecture. First, all nodes in the graph are ordered using some labeling procedure (e.g. node degree, centrality, or Weisfeiler-Lehman color \cite{gnn_survey_comprehensive_survey_wu}), and $w$ nodes are selected using stride $s$ (padding with zeros if required). For each one of the selected nodes, the neighborhood is assembled using breadth-first search, so that each node has exactly $k$ neighbors; note that BFS node selection order is known, as the order of nodes is fixed. Graph normalization is a mathematically involved step, in which the graph is transformed into a regular Euclidean grid structure. The idea is that nodes from different graphs that have similar structural roles should be assigned similar positions. After this step, a regular CNN architecture can be used, as the normalized neighborhoods act as receptive fields and node and edge attributes are regarded as channels. This architecture inherently leverages both node and edge attributes and is able to apply regular CNN architectures to graph data. However, it is computationally expensive (optimal neighborhood normalization is even NP-hard, so approximations have to be used \cite{convgnn_patchy_san}), the multistep method is complicated, and the method does not really work on graph data - it merely transforms it into an easier problem in Euclidean domain, which may lose some important information and makes layered composition much harder.

GraphSAGE \cite{convgnn_graphsage} proposes two innovations: a general inductive framework for different ConvGNNs and a neighborhood sampling method. Node neighbors are uniformly sampled to create fixed-size sets, which keeps computational cost constant, which is especially advantageous for graphs with varying node degrees, e.g. social networks. However, this comes at the cost of lowered accuracy. Neighbors are then aggregated using a chosen aggregating function, leading to different GraphSAGE subtypes. Mean aggregation can be interpreted as an approximation of a transductive GCN approach, or as a form of skip connection \cite{introduction_to_graph_neural_networks}. LSTM aggregator has a larger expressive capability but is not permutation invariant and node neighbors are permuted to simulate this. The pooling aggregator uses a fully connected layer with max pooling of outputs. In addition to those mechanisms, an unsupervised loss function is proposed to encourage nearby nodes to have similar representations and distant nodes to have different representations.

A quite special type of spatial ConvGNN is Graph Isomorphism Network (GIN) \cite{gnn_GIN}, which is constructed using very different principles compared to many other architectures. Authors show the connection between GNNs and Weisfeiler-Lehman isomorphism test, showing that typical GNNs (e.g. GCN, GraphSAGE) are at most as powerful as WL graph isomorphism test. They also provide examples of graph structures that cannot be distinguished by such architectures and propose GIN model to solve this problem, which is proved to be as powerful as WL test. The main requirement for this is using injective $AGGREGATE$ and $COMBINE$ functions, which are summation and 2-layer MLP, respectively. For graph classification, the choice of readout function is also studied in this work, and the sum function is proved to be more powerful in distinguishing graph structure than mean and max.

\subsubsection{Attentional architectures}
\label{section_literature_gnn_attentional}

Attention mechanism \cite{attention_bahdanau,attention_self_attention,attention_is_all_you_need} can be considered an additional learned weighting scheme, using which the more \q{important} input features receive higher weights. This contrasts with plain convolutional architectures, which assume either the identical contribution of neighbors to the central node (e.g. GraphSAGE) or with predetermined weights (e.g. GCN) \cite{gnn_survey_comprehensive_survey_wu}. A special type of attention is self-attention \cite{attention_self_attention}, most often used in GNNs, where input attends to itself, identifying the most important parts. Attentional architectures are inherently spatial, as they operate on a node neighborhood to assess the importance of neighbors.

Graph Attention Network (GAT) \cite{attentional_gat} applies the self-attention mechanism (a single layer feedforward neural network) to node neighbors as an attention layer to weight them. Additionally, multihead attention \cite{attention_is_all_you_need} stabilizes the learning process \cite{introduction_to_graph_neural_networks} and increases the model capacity \cite{gnn_survey_comprehensive_survey_wu}; $k$ independent attention mechanisms are used in parallel and their outputs are concatenated or averaged to obtain the node embedding. This method deals with the problem of different node degrees through learned weights, so that wide node distributions are not problematic. It is also efficient in terms of parallelization capabilities and can be easily applied to inductive learning problems. The downside is that weighting may sometimes result in highly varying predictions and high standard deviations \cite{evaluation_pitfalls_of_gnn_evaluation}. The authors of \cite{attentional_gat_v_2} observe that applying linear layers one after another in GAT means that the ranking of attended nodes is unconditioned on the query node. To deal with this problem, they propose GATv2 architecture, in which every node can attend to any other node. They show that this way the model is more resistant to structural noise (e.g. small changes in graph structure) and requires fewer attention heads to perform as good or better than plain GAT. Since this method requires just a simple reordering of operations, it does not increase the computational cost and retains the advantages of GAT.

Gated Attention Network (GaAN) \cite{attentional_gaan} also uses the multihead attention mechanism, but in contrast to GAT it uses key-value attention and dot product attention. Moreover, while in GAT heads have uniform weights, in GaAN they are weighted using an additional soft gate, called gated attention aggregation, which is implemented as a convolutional network. This allows differentiating more, and less important subspaces created by attention heads, which may be especially different between nodes, e.g. with different structural roles in the graph.

\subsubsection{Readout}
\label{section_literature_sampling_pooling_readout}

Readout layer aggregates the node representations flatly into a single graph embedding vector. It is also called global pooling \cite{pytorch_geometric}, global aggregation, \cite{gnn_GIN} or direct pooling \cite{gnn_survey_review_of_methods_zhou}.

The most basic and, at the same time the most popular, readout techniques are mean, max, and sum \cite{gnn_survey_comprehensive_survey_wu,pytorch_geometric}, which are applied as the last layer before MLP. As shown in \cite{gnn_GIN}, the sum function is the most powerful of the three, as there exist graph structural patterns that mean and max aggregators cannot differentiate, but the sum can. Alternatively, a weighted average can be used, with weights computed using attention, called global (soft) attention \cite{pooling_global_attention_pool,pytorch_geometric}.

SortPool \cite{pooling_sort_pool} proposes sorting the node embeddings according to the structural roles of the nodes. WL coloring is used to this end as a graph labeling technique since it provides an ordering consistent among graphs, i.e. if nodes have similar structural properties in different graphs, they will be assigned similar relative positions. Such structure can be fed to 1-dimensional CNNs and max pooling layers, and their output is classified with MLPs.

Sequence-to-sequence models have been adapted to sets \cite{pooling_set2set} and are known as Set2Set \cite{pytorch_geometric, gnn_survey_review_of_methods_zhou}. They can be used for global pooling, as a graph is an unordered set of nodes. It uses an LSTM-based method and special attention, called content-based attention, which has the property that the vector retrieved from memory using the attention will not change if the memory is randomly shuffled.

\clearpage

\subsection{Graph descriptors}
\label{section_literature_descriptors}

Graph descriptors, also called graph invariants or features, have been developed historically to capture the topological structure of the graph. Since they are hand-crafted and based on mathematical properties or domain-specific observations, they are highly interpretable and their advantages and disadvantages in particular situations are known.

A distinction can be made based on which graph element is described: the whole graph or its element - vertex, edge, or a pair of such. Graph invariants describe the graph as a whole, typically reflecting a single topological property. For this reason, they cannot be used to reconstruct the graph univocally, but are also largely independent (uncorrelated), meaning that multiple can be used together to form a good feature set for graph classification. They directly describe the global structure of the graph. Examples include density, diameter, radius, efficiency, graph clustering coefficient, and multiple features derived from graph spectrum, such as spectral radius or algebraic connectivity \cite{descriptor_graph_dynamical_processes_on_complex_networks, descriptor_graph_spectral_radius,descriptor_graph_algebraic_connectivity,czech_doktorat,handbook_of_graph_theory}.

Vertex descriptors determine vertex importance or its topological role in the graph. Edge descriptors are similar to vertex descriptors, but they describe edges. Pair descriptors operate on two vertices or edges. They often describe the similarity or dissimilarity of the given objects. For graph classification, the values of the given element descriptors are aggregated, similarly to the node feature aggregation in GNNs. Typically, a histogram of values with a given number of bins (which is a hyperparameter) is calculated. Element descriptors include:
\begin{itemize}
    \item vertices: degree, clustering coefficient, betweenness centrality, PageRank, communicability betweenness \cite{czech_doktorat,descriptor_vertex_watts_strogatz, descriptor_vertex_pagerank}
    \item edges: edge connectivity, range of edge, edge betweenness \cite{czech_doktorat,descriptor_edge_features}
    \item pairs: shortest path length, longest path length, commute time \cite{czech_doktorat,descriptor_pair_commute_time}
\end{itemize}

Many complex graph descriptors, called molecular fingerprints, were created in the field of cheminformatics (area between computer science and computational chemistry). They are domain-specific, as they have been designed precisely to capture the physiochemical properties of molecules. Their calculation typically requires chemical graphs, which are node-attributed with atom types and edge-attributes with bond types. They have been used mostly in scenarios requiring graph classification, such as virtual screening (high-throughput screening (HTS), discarding inactive molecules in large molecular databases) or structure-activity relationship modeling (SAR modeling, predicting bioactivity of molecules).

From a mathematical point of view, they are often similar to bag-of-words, where a number of topological substructures are first counted and then hashed into the fixed-length vector (representing a dictionary/hashmap). This makes those descriptors lie somewhere between node and graph descriptors. Arguably the most famous descriptor is an Extended Connectivity Fingerprint (ECFP) \cite{fingerprints_ECFP}, also known as circular or Morgan fingerprint. It relies on counting atom types in node neighborhoods with increasing radii (in a fashion similar to Breadth-First Search (BFS)). RDKit fingerprint \cite{fingerprints_RDKit}, introduced by RDKit open-source library for cheminformatics, has been inspired by the proprietary Daylight fingerprint. It identifies all subgraphs with a given size in a molecule, taking into account atom and bond types, counts them, and hashes them into fixed-size vectors. MACCS keys \cite{fingerprints_MACCS} is a unique fingerprint because its 166 features are not learned from data, but instead were chosen by domain experts as useful in molecular property prediction. They are mostly node feature counts or subgraph counts.

\subsection{Evaluation methodology}
\label{section_literature_evaluation}

Proper model selection and assessment are crucial for selecting the classifier that will really perform the best for future data. Additionally, a clear and concise description of the model is required to understand the formal basis for proposed algorithms. Ablation studies and the analysis of particular elements are also required to assess the broader applicability of the model and its sensitivity to domains, dataset sizes, and other variables. This makes the task of proper model evaluation a hard and important problem in machine learning in general.

In work \cite{ESL}, the descriptions of typical model validation methods are provided. In particular, authors describe 3 groups of methods: holdout, cross-validation (CV), and bootstrapping. In holdout, a single test set it is used to assess the model performance and is used when the data is sufficiently large. Cross-validation uses multiple test folds, but each sample from the dataset is used only once as a test sample. This method is more useful for small datasets, and it also provides means to calculate the standard deviation of the test score. Bootstrapping methods rely on creating multiple (tens or hundreds) test sets, drawn with replacement from the dataset, and as such, they require sophisticated improvements to not be overly optimistic. Typically, authors precisely differentiate test and validation sets, as evaluation methods can be used in many configurations, still the final results should always be reported on the test set.

The authors of \cite{evaluation_experimental_study_of_ml} focus on the design of experiments and their evaluation in machine learning. In particular, they highlight the importance of train-test division and the importance of baselines. The final model metrics should be compared to simple baselines, and other algorithms from relevant literature and also take into consideration class imbalance, which is common for many domains, including molecular property prediction.

The rapid development of GNN algorithms raised a concern about the quality of ML practices and evaluation methodologies specifically in this domain. The authors of \cite{evaluation_pitfalls_of_gnn_evaluation} analyze architectures for node classification on standard datasets and also introduce 4 new datasets. The last part is important, as new models may be optimized toward standard benchmark datasets, and testing their performance on previously unseen datasets from a similar domain leads to a challenging and fair comparison. Indeed, the authors find out, e.g. that GAT has a considerably high variance on 2 new datasets, a behavior that has not been seen before on the standard datasets. The authors also follow good practices from \cite{evaluation_troubling_trends_in_ml}, where for fair comparison the model architectures and many parameters are fixed, hyperparameter grids are provided, and the same training, validation, and model selection procedure is applied to all compared algorithms. For each dataset, 100 random training/validation/test splits are used, where each test result is obtained as an average over 20 random initializations for that split. 100 results are averaged as a final result for the dataset. For model comparison, instead of performing paired t-tests (which compare one model to another), a model ranking scheme is proposed to compare every model to every other model.

Additionally, the work \cite{evaluation_gnn_fair_comparison} concerns the fair evaluation of graph classification and is, therefore, the most relevant to this thesis. The authors compare 5 GNN architectures and a strong baseline, which turns out to be the best model in many cases. Adding node degree as an additional node feature is also considered, giving better results on social datasets. The following evaluation scheme is proposed: 10-fold cross-validation for testing, where for each test fold an average of 3 random initializations is used as a result. Inner validation with holdout is used for hyperparameter tuning.

The works above raise a question of proper evaluation practices in GNNs and the importance of comparing the results between works. To assess the possibility of referring to the results from other works, we perform the overview of evaluation methodology in the studies about graph classification with GNNs.

Multiple works report only the performance of models on node classification tasks and not for graph classification. It is nevertheless important to check their methodology for a comprehensive overview. GCN \cite{convgnn_gcn} uses a holdout for testing, repeated 10 times, and holdout for validation. AGCN \cite{convgnn_agcn} uses holdout for testing and a 5-fold CV for validation. DCNN \cite{convgnn_dcnn} contains results both for node and graph classification, and for node classification, holdout is used for both validation and testing. GraphSAGE \cite{convgnn_graphsage} and GAT \cite{attentional_gat} use holdout for validation and testing.

Other models provide results for graph classification and typically use the same or similar datasets from biology and chemistry. Neural FPs \cite{convgnn_neural_fp} use holdout for testing and validation. DCNN \cite{convgnn_dcnn} for graph classification uses holdout for validation and testing, and it should be noted that holdout parts are unusually large, both consisting of 1/3 of the dataset. PATCHY-SAN \cite{convgnn_patchy_san} has an unclear description, but as we understand it, the hyperparameters on the same datasets were used for reporting results, meaning that no test set was really used, and only the validation results are provided, using 10-fold CV. GIN \cite{gnn_GIN} only provides 10-fold validation results and does not use the test set.

Some GNNs are evaluated only on non-standard tasks, which makes the direct comparison of those approaches with other works impossible. DGCN \cite{convgnn_dgcn} evaluates the model on datasets from image segmentation, using holdout for testing and validation. DGC \cite{convgnn_dgc} is used for spatio-temporal time series forecasting, with a holdout for testing and validation.

Nevertheless, using sophisticated evaluation methodologies is problematic from the implementational and computational points of view. Retraining neural networks multiple times can be prohibitively costly, even on small datasets. Using larger and more challenging datasets makes evaluation easier, as the holdout method on a properly challenging test set is often enough. Due to the rising popularity of GNNs, such benchmarks have been created, e.g. MoleculeNet \cite{datasets_moleculenet} or Open Graph Benchmark \cite{datasets_ogb}. Standardized benchmarks also often feature precalculated training-validation-test splits for the holdout method to compare algorithms in a controlled environment.

\clearpage

\section{Theoretical background}
\label{section_theory}

\noindent In this chapter, we introduce the theory behind the main elements of the thesis. We explain the mathematical foundations of graph neural networks, graph descriptors, and graph classification algorithms. Lastly, we describe approaches to model evaluation.

\subsection{Graph neural networks}
\label{section_theory_gnns}

The goal of a GNN is to learn a function assigning each graph a label from a set of classes. This is typically realized as a two or three-step process, highly resembling how classic convolutional neural networks (CNNs) work:
\begin{enumerate}
    \item Node embedding - each node from the graph is represented as a vector, i.e. embedding
    \item Graph embedding (readout) - node embeddings are combined with a readout function, creating an embedding vector for the whole graph
    \item Graph classification - a multilayer perceptron (MLP) is used to classify the graph vector
\end{enumerate}
This is a typical end-to-end deep learning process, as backpropagation is used to learn the weights of both MLP and embedding parts. The 3-step process stems from adapting GNNs for node classification to graph classification tasks, since node classification works very similarly, albeit without the readout step (only steps 1 and 3).

\begin{figure}[h]
\makebox[\textwidth][c]{\includegraphics[width=\textwidth]{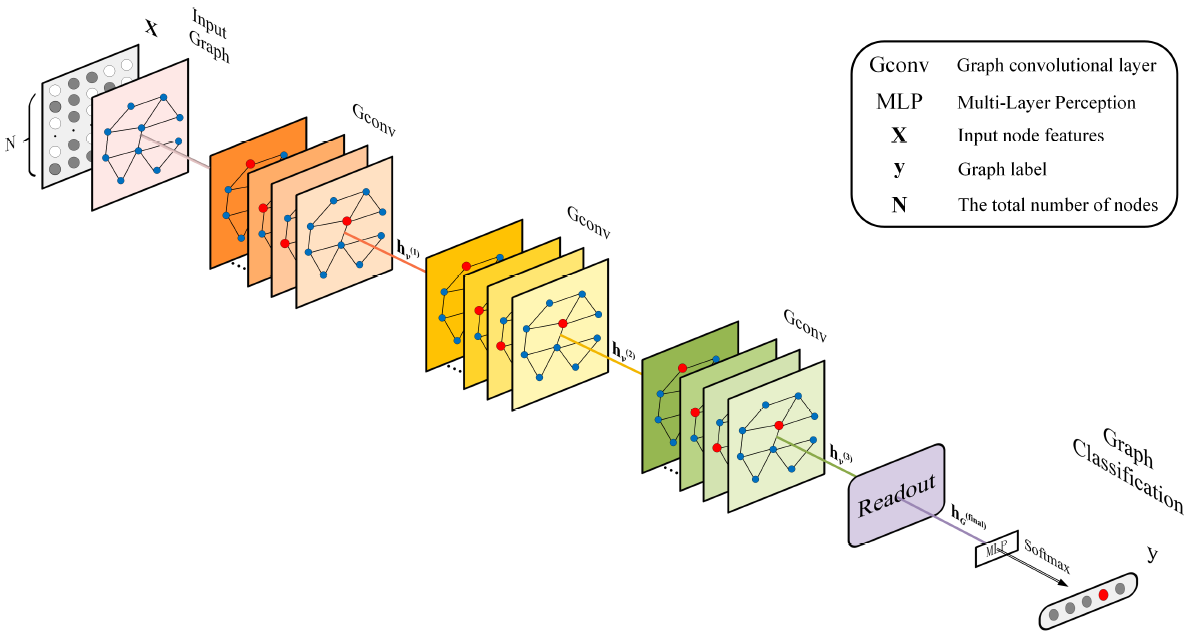}}
\caption{Visualization of general convolutional GNN architecture. Source: \cite{flat_gnn_architecture_graphics}}
\label{flat_gnn_architecture_graphics}
\end{figure}

The architecture of a typical GNN is similar to CNNs, consisting of stacked graph convolutional layers, each passing its output to the next one, as shown in Figure \ref{flat_gnn_architecture_graphics}. At the end, the readout function aggregates the output of the last layers to a vector.

GNNs use both graph topology and node features to learn node embeddings. Graph structure is represented as an adjacency matrix $A$, while node features are stacked as horizontal vectors into the matrix $X$. Convolutional layers output transformed feature vectors, i.e. $H^{(i)}$ is the output of $i$-th layer, where $H^{(0)} = X$. A layer is therefore a function $H = f(X, A)$.

In the rest of this section, we emphasize the most important GNN architectures by writing them \textbf{in bold}.

\subsubsection{Graph convolutional layers}
\label{section_theory_gnns_graph_convolutions}

First, we introduce spectral ConvGNNs, as they have a solid theoretical foundation in graph signal processing and were pioneering works in ConvGNNs. This section closely follows multiple related works \cite{convgnn_bruna,convgnn_gcn,gnn_survey_representation_learning_on_graphs,gnn_survey_comprehensive_survey_wu} and the works about respective architectures.

For degree matrix $D$ and adjacency matrix $A$, we can represent the graph with symmetric  normalized Laplacian matrix $L = I - D^{-\frac{1}{2}} A D^{-\frac{1}{2}}$. As this matrix is positive semidefinite, its eigendecomposition is $L = U \Lambda U^T$, where $U$ denotes the matrix forming an orthonormal space. The graph Fourier transform of a signal $x$ (e.g. a vector of a single graph feature, $x \in \mathbb{R}^N$) is $\mathcal{F}(x) = U^Tx = \hat{x}$, and the inverse Fourier transform is $\mathcal{F}^{-1}(\hat{x}) = U\hat{x}$.

Spectral graph convolution is then defined for convolutional filter $g_\theta = \text{diag}(\theta)$ as:
\begin{equation}
g_\theta \star x = U g_\theta U^T x.
\end{equation}

The choice of $g_\theta$ can vary. Most naturally it can be parametrized with eigenvalues of the Laplacian $\Lambda$, so we have $g_\theta(\Lambda)$. Using the exact definition has two considerable downsides. First, computing the eigendecomposition has $O(n^3)$ complexity, which is prohibitively high in practice. Additionally, any change to the graph changes the eigenbasis, and it is heavily domain-dependent, meaning that it will not generalize well to graphs with even moderately different structures than seen during training; this is similar to overfitting.

In the work \cite{convgnn_hammond} it is shown that $g_\theta(\Lambda)$ can be approximated by truncated series of Chebyshev polynomials $T_k(x)$ up to $K$-th order. This results in convolution, which is used in \textbf{ChebNet} \cite{convgnn_chebnet}:
\begin{equation}
\begin{split}
& g_\theta \star x \approx \sum_{k=0}^K \theta_l T_k(\tilde{L})x \\
& \tilde{L} = \frac{2}{\lambda_{max}}L - I,
\end{split}
\end{equation}

\noindent
where $\lambda_{max}$ is the largest eigenvalue. The convolutional filter $\theta$ is now a vector of Chebyshev coefficients, which are recursively defined as $T_0(x) = 1$, $T_1(x) = x$, $T_k(x) = 2xT_{k-1}(x) - T_{k-2}(x)$. This formulation is $K$-localized, i.e. for each node it depends only on its $K$-th order neighborhood. This is a desirable feature, as this means that filters can extract local features independently of graph size \cite{gnn_survey_comprehensive_survey_wu}. Also, for this reason, this method is also a precursor of spatial methods, which are characterized by their locality.

However, the previous approach still overfits, especially on local neighborhood structures for graphs with very wide node degree distributions, common e.g. in social networks. It is also hard to stack layers and is not fully localized. \textbf{Graph Convolutional Network (GCN)} \cite{convgnn_gcn}, arguably the most commonly known and used ConvGNN (or GNN in general), proposes a few simplifications.

Firstly, we set $K=1$ and learn a function linear on the graph Laplacian spectrum in each layer. This limits the range of possible filters, reducing overfitting, but removes the need for explicit parametrization of $g_\theta$, e.g. by the Chebyshev polynomials, and allows easy stacking of multiple layers. Furthermore, we approximate $\lambda_{max} \approx 2$, as we expect other parameters to adapt to this during training. Setting $K=1$ leaves us with 2 filter parameters $\theta_0$ and $\theta_1$, and to reduce overfitting even further, we restrain the number of parameters by setting $\theta = \theta_0 = -\theta_1$. After introducing all those simplifications, we have:
\begin{equation}
g_\theta \star x \approx \theta \left(
I + D^{-\frac{1}{2}} A D^{-\frac{1}{2}}
\right) x.
\end{equation}

Since $I + D^{-\frac{1}{2}} A D^{-\frac{1}{2}}$ has eigenvalues in range $[0, 2]$, stacking multiple such layers could lead to numerical instabilities and exploding/vanishing gradients. For this reason, we add self-loops to the graph (an edge from each vertex to itself) in the so-called renormalization trick, replacing $I + D^{-\frac{1}{2}} A D^{-\frac{1}{2}}$ by 
$\tilde{D}^{-\frac{1}{2}} \tilde{A} \tilde{D}^{-\frac{1}{2}}
$, where $\tilde{A} = A + I$ and $\tilde{D}_{ii} = \sum_j \tilde{A}_{ij}$. This graph convolution can easily be formulated for $D$-dimensional node features (i.e. $D$ channels):
\begin{equation}
Z = \tilde{D}^{-\frac{1}{2}} \tilde{A} \tilde{D}^{-\frac{1}{2}} X \Theta,
\end{equation}

\noindent
where $X \in \mathbb{R}^{N \times D}$ is the graph feature matrix and $\Theta \in \mathbb{R}^{D \times F}$ is a matrix of filter parameters. Since the filters are learnable weights, $W$ is often used instead of $\Theta$ to mark this matrix. The full layer-wise propagation rule is therefore defined as:
\begin{equation}
H^{(k+1)} = \sigma \left( \tilde{D}^{-\frac{1}{2}} \tilde{A} \tilde{D}^{-\frac{1}{2}} H^{(k)} W^{(k)} \right),
\end{equation}

\noindent
where $\sigma$ denotes an activation function, most often $ReLU(x) = \max(0, x)$.

It is also a first-order approximation of ChebNet, which is a localized spectral filter, therefore making GCN $1$-localized, i.e. each node aggregates messages from its direct neighborhood. For this reason, this method can also be interpreted as a spatial convolution. This notion is further reinforced by observing that GCN has node-wise formulation \cite{convgnn_gcn,graph_representation_learning}:
\begin{equation}
h^{(k + 1)}_u = \sigma \left( W^{(k)} \sum_{v \in N(u) \cup \{u\}} \frac{h_v^{(k)}}{\sqrt{d(u) d(v)}} \right)
\end{equation}

\noindent
where $d(v)= |N(v)|$ is the degree of node $v$. This weighted average normalizes the influence of neighbors using their degrees. It can be interpreted as smoothing the hidden representations locally, along the edges of the graph, encouraging similar predictions for close nodes. Additionally, as noted in \cite{convgnn_gcn}, this can be loosely interpreted as a differentiable and parametrized generalization of the Weisfeiler-Lehman algorithm with $k=1$.

In \cite{oversmoothing} it is shown that the convolution formulation of GCN is basically a form of Laplacian smoothing. This also explains why a small number of layers (e.g. 1-3 in \cite{convgnn_gcn}) often works best for ConvGNNs - having too many layers causes oversmoothing, a negative phenomenon when embeddings of neighboring nodes become indistinguishable.

One of the latest developments of the line of reasoning presented above is \textbf{Simplified Graph Convolutional network (SGC)} \cite{convgnn_sgc}. It is based on the hypothesis that nonlinearity $\sigma$ between layers is not critical, but rather the results of GCN stem simply from local averaging. Denoting $S = \tilde{D}^{-\frac{1}{2}} \tilde{A} \tilde{D}^{-\frac{1}{2}}$, in GCN we have:
\begin{equation}
H^{(k + 1)} = \sigma \left( \tilde{D}^{-\frac{1}{2}} \tilde{A} \tilde{D}^{-\frac{1}{2}} H^{(k)} W^{(k)} \right) = \sigma \left( S H^{(k)} W^{(k)} \right).
\end{equation}

\begin{figure}
\makebox[\textwidth][c]{\includegraphics[width=1.2\textwidth]{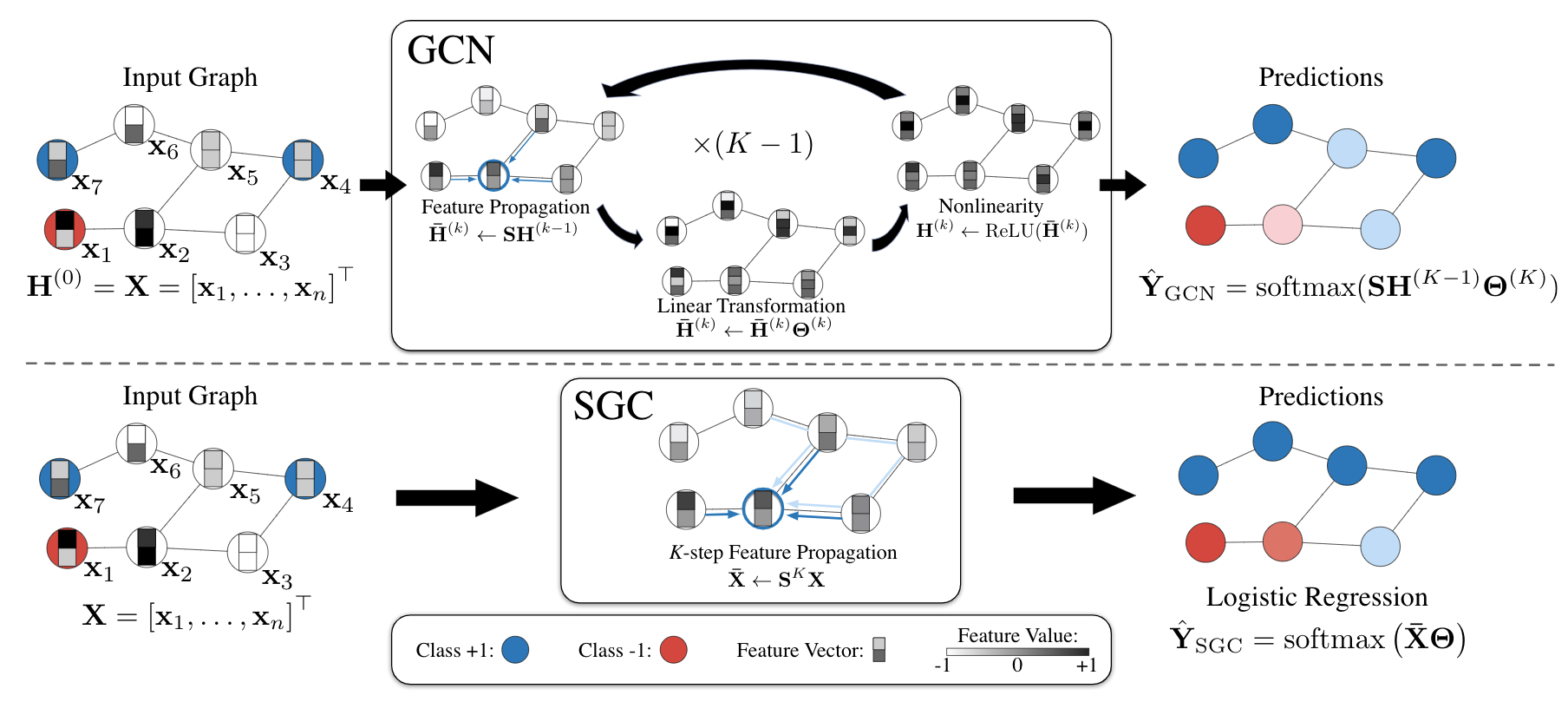}}
\caption{The visualization of GCN and SGC for node classification. Source: \cite{convgnn_sgc}}
\label{convgnn_sgc_graphics}
\end{figure}

In SGC, on the other hand, we remove the $\sigma$ function, and the expression for $K$-layer GNN becomes:
\begin{equation}
H^{(K)} = S^K X W,
\end{equation}

\noindent
where matrix $S$ is raised to the $K$-th power. This model has the receptive field of $K$ hops, same as similar GCN, but has only one weight matrix instead of $K$, which makes it extremely fast. The authors of SGC also show that SGC corresponds to a fixed low-pass filter on the graph spectral domain. While SGC is aimed at node classification, it can be used for graph classification as well as a simple baseline model. For a visual comparison between GCN and SGC, see Figure \ref{convgnn_sgc_graphics}.

The majority of modern ConvGNNs are spatial approaches, since their spatial locality typically corresponds to lower computational cost and better predictive performance due to capturing the properties of node neighborhoods. Two arguably most well-known architectures in this group are GraphSAGE  and Graph Isomorphism Network (GIN).

\textbf{GraphSAGE} \cite{convgnn_graphsage} proposes two changes compared to GCN: neighborhood sampling and parameterizable aggregation function. The first technique enforces using the same, constant neighborhood size for all nodes, no matter their degree, which makes the computational cost predictable and constant, which is not the case for other GNNs. This can be used for any GNN to reduce the computational cost, but is reported \cite{convgnn_graphsage} to reduce the performance quality as a consequence. The propagation scheme in GraphSAGE is highly parameterizable and explicitly models the $AGGREGATE$ and $COMBINE$ functions. Each node first aggregates the representation of neighboring nodes, but not including itself, as $h_{N(v)}^k = AGGREGATE(N(v))$, where the aggregating function can be any function invariant to input permutation (i.e. symmetric) and with sufficiently high representational capacity, for example:
\begin{enumerate}
    \item Mean - we take the average value of each feature (or hidden dimension), which is similar to GCN. However, in GraphSAGE we do not include the node itself in this averaging. While not presented in the paper, it is possible to use another permutation invariant function, such as sum or max, which has been implemented, e.g. as a part of the PyTorch Geometric \cite{pytorch_geometric} library.
    
    \begin{equation}
    h_{N(v)}^{(k + 1)} = \text{mean} \left( \{ h_u^{(k)}, \forall u \in N(v) \} \right)
    \end{equation}
    
    \item LSTM - it is a trainable aggregator, representing a more complex function, with high expressive capability. Since LSTMs process input sequentially, the neighbors are randomly shuffled before aggregation, to simulate permutation invariance.
    
    \begin{equation}
    h_{N(v)}^{(k + 1)} = \text{LSTM} \left( \text{permute} \left( \{ h_u^{(k)}, \forall u \in N(v) \} \right) \right)
    \end{equation}
    
    \item Pooling - neighboring vectors are first transformed by MLP (typically a 1-layer network with ReLU activation), and then pooled with max function. This approach aims to capture the different aspects of neighboring nodes and implicitly weight them \cite{skip_jumping_knowledge}.
    
    \begin{equation}
    h_{N(v)}^{(k + 1)} = \max \left( \{ \text{MLP}(h_u^{(k)}), \forall u \in N(v) \} \right)
    \end{equation}
    
\end{enumerate}

The $COMBINE$ function concatenates the node vector with its aggregated neighbors vector and feeds the result to a single layer MLP, creating the output hidden representation for that node:
\begin{equation}
h_v^{(k+1)} = \sigma \left( W^{(k)} \left[ h_v^{(k)} \concat h_{N(v)}^{(k)} \right] \right),
\end{equation}

\noindent
where $\sigma$ is almost always the ReLU function. The most important difference between GCN and the mean variant of GraphSAGE (the most similar one) is the unique treatment of the node itself in the latter - in GCN it is averaged just like its neighborhood. In GraphSAGE this can be interpreted as a type of skip (residual) connection, enabling putting more weight on the node itself, compared to its neighborhood. Formally, the difference is:
\begin{equation}
\begin{split}
& \text{GCN:} \quad h_v^{(k+1)} = \sigma \left( W^{(k)} \cdot \text{mean} \left( \{ h_u^{(k)}, \forall u \in N(v) \cup \{ v \} \} \right) \right) \\
& \text{GraphSAGE:} \quad h_v^{(k+1)} = \sigma \left( W^{(k)} \left[ h_v^{(k)} \concat \text{mean} \left( \{ h_u^{(k)}, \forall u \in N(v) \} \right) \right] \right).
\end{split}
\end{equation}

\textbf{Graph Isomorphism Network (GIN)} \cite{gnn_GIN} formally explores the connection between GNNs and Weisfeiler-Lehman (WL) graph isomorphism test \cite{wl_kernel}. Authors show that generally, GNNs are at most as powerful as WL test in distinguishing graph structures and show examples of structures that  previous architectures such as GCN or GraphSAGE are unable to distinguish. However, those GNNs have considerable advantages over WL test: it focuses only on graph topology, while GNNs make use of graph features, and additionally, they learn not only to distinguish graphs structures but to embed them, so that the distance between similar graphs in the embedding space will be small. The GIN architecture aims to be as powerful as WL test at discriminating between graph structures, and at the same time provide advantages of GNNs, in particular spatial ConvGNNs. Authors prove that this can be achieved in a very simple way: neighborhood aggregation and readout functions need to be injective. As MLPs are universal function approximators \cite{nn_universal_approximators,nn_universal_approximators_2} and can therefore approximate injective functions, they can be used here. The authors show that single-layer MLP has insufficient capabilities in this regard and that 2-layer MLP is required. Neighborhood vectors are summed before transforming them with MLP. The equation for update becomes:
\begin{equation}
h_v^{(k+1)} = \sigma \left( \text{MLP}^{(k)} \left( \left( 1 + \varepsilon^{(k)} \right) \cdot h_v^{(k)} + \sum_{u \in N(v)} h_u^{(k)} \right) \right).
\end{equation}

$\varepsilon$ can be either a learnable parameter, a fixed constant, or set to 0, each option less powerful than the previous one. However, the authors show that setting $\varepsilon = 0$, while making the network less powerful, makes it generalize better. With this simplification, the update can be efficiently represented in matrix form as:
\begin{equation}
H_v^{(k)} = \sigma \left( \text{MLP}^{(k)} ((A + I) \cdot X) \right).
\end{equation}

The attention mechanism \cite{attention_bahdanau,attention_self_attention,attention_is_all_you_need} is basically a learnable weighting scheme, allowing nodes to \q{attend} to some neighbors, which are deemed more interesting and informative, more than others. Those networks are inherently spatial, as nodes attend to their neighbors. This can be thought of as a generalization of static weight in GCN $\frac{1}{\sqrt{d(u) d(v)}}$, which acts as a normalizing constant there; in attentional GNNs this is learned, rather than set explicitly. 

Arguably the most well-known attention-based GNN is \textbf{Graph Attention Network (GAT)} \cite{attentional_gat}. In this model, every two neighboring nodes attend to each other, using the arbitrary attention function $attention(u, v)$. Additionally, the node attends to itself (similar to a self-loop). To obtain sufficient expressive power, features are linearly transformed into higher-level space before computing attention scores, and attention scores are normalized with the softmax function afterward to form a probability distribution for each node. For clarity, we omit the $(k)$ mark below, but all relevant equations concern a single layer, e.g. $W$ means $W^{(k)}$.
\begin{equation}
\begin{split}
& \tilde{h}_u = W h_u \quad\quad \tilde{h}_v = W h_v \\
& e_{uv} = attention \left( \tilde{h}_u, \tilde{h}_v \right) \\
& \alpha_{uv} = \text{softmax}_v(e_{uv}) = \frac{\exp(e_{uv})}{\sum_{v \in N(u) \cup \{u\}} \exp(e_{uv})}
\end{split}
\end{equation}

GAT model is agnostic to the choice of the attention mechanism. The authors of \cite{attentional_gat} chose Bahdanau attention \cite{attention_bahdanau}, also known as additive score function, in the form of:
\begin{equation}
attention_{add}(\tilde{h}_u, \tilde{h}_v) = \text{LeakyReLU} \left( a^T \left[ \tilde{h}_u \concat \tilde{h}_v \right] \right),
\end{equation}

\noindent
where $a$ is a vector of attention weights (for the given layer) and LeakyReLU is a smooth approximation of ReLU, with negative input slope $\alpha$ (in \cite{attentional_gat} $\alpha = 0.2$ is used).
\begin{equation}
\text{LeakyReLU} = 
\begin{cases}
x & x \geq 0 \\
\alpha * x & x < 0
\end{cases} 
\end{equation}

Normalized attention scores are applied to linearly transformed node features, and the final update equation for the node in GAT becomes:
\begin{equation}
\begin{split}
& h_u^{(k + 1)} = \sigma \left( \sum_{v \in N(u) \cup \{ u \}} \alpha_{uv} W^{(k)} h_v^{(k)}
\right) \\
& \alpha_{uv} = \frac{\exp \left( \text{LeakyReLU} \left( a^T \left[ W^{(k)} h_u^{(k)} \concat W^{(k)} h_v^{(k)} \right] \right) \right)}{\sum_{v \in N(u) \cup \{u\}} \exp \left (\text{LeakyReLU} \left( a^T \left[ W^{(k)} h_u^{(k)} \concat W^{(k)} h_v^{(k)} \right] \right) \right)}
\end{split}
\end{equation}

To stabilize the learning process, multihead attention \cite{attention_is_all_you_need} is used in GAT. Multiple attention mechanisms, called heads in this context, are used in parallel and can be either concatenated or averaged. In GAT concatenation is used in all layers except the last one, where averaging is used, in order to keep the proper dimensionality (since otherwise, $T$ attention heads with concatenation result in $T \cdot D$ output dimensionality). This does not necessarily result in increased computational cost, since when using multiple attention heads the output dimensionality stays the same, while accuracy is often higher \cite{attention_is_all_you_need}.

In Equation \ref{equation_multihead_attention}, following \cite{attentional_gat}, we mark concatenation of series of vectors with $\concat$ operator.

\begin{equation}
\begin{split}
& h_u^{(k + 1)} = \concat_{t=1}^T \sigma \left( \sum_{v \in N(u) \cup \{ u \}} \alpha_{uv} W^{(k)} h_v^{(k)} \right) \\
& h_u^{(K)} = \sigma \left( \frac{1}{T} \sum_{t=1}^{T} \sum_{v \in N(u) \cup \{ u \}} \alpha_{uv} W^{(k)} h_v^{(k)} \right)
\end{split}
\end{equation}
\label{equation_multihead_attention}

\begin{figure}
\makebox[\textwidth][c]{\includegraphics[width=\textwidth]{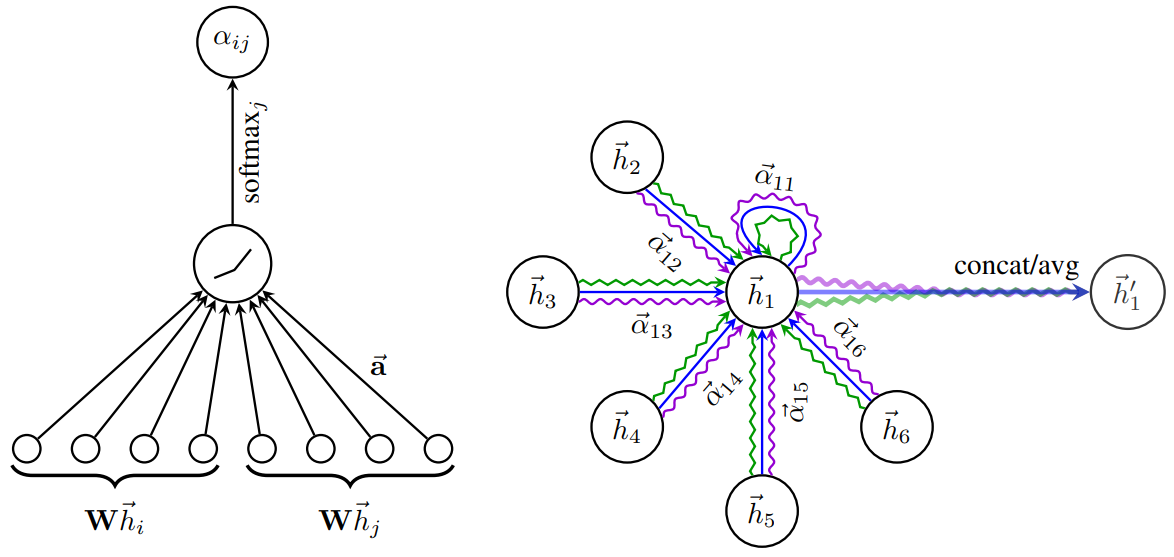}}
\caption{Left: GAT attention mechanism. Right: illustration of multihead attention for $K = 3$. Source: \cite{attentional_gat}}
\label{fig_minimal}
\end{figure}

\subsubsection{Readout layer}
\label{section_theory_gnns_aggregation}

Convolutional layers exchange information between neighboring nodes. While this propagates information in the graph (using its structure), it still requires gathering global information for graph classification. This is achieved through $READOUT$ function, gathering information from all node embeddings and flattening them into a single graph embedding vector, which can be then classified. This operation is also sometimes called global pooling \cite{pytorch_geometric}.

Most often, simple and permutation invariant functions are used: \textbf{mean}, \textbf{max} and \textbf{sum}. The selections found in the literature often seem arbitrary, however, the authors of GIN \cite{gnn_GIN} analyze this and prove that sum is the most powerful function of the three, being able to distinguish graph structures that are indistinguishable for mean and max functions. Some works suggest that using more than one function is beneficial, as their results can be concatenated, and more information is extracted, e.g. \cite{pooling_gpool,pooling_sagpool} use mean and max, while \cite{convgnn_graph_unet} uses all three.

More advanced readout schemes also exist, which aim to aggregate the information from all nodes in a more sophisticated way, weighting their influence. The examples are SortPool \cite{pooling_sort_pool}, which sorts the nodes and transforms the output using 1D CNN (due to the loss of injectivity because of the sorting), global attention \cite{pooling_global_attention_pool} using the attention mechanism for weighting nodes, and Set2Set (S2S) \cite{pooling_set2set} based on the sequence-to-sequence model for learning on sets. Those techniques are more computationally involved and are outside the scope of this thesis.

\subsubsection{Skip connections}

Using skip connections (also called shortcut connections), i.e. passing the input, or output of previous layers, directly to further layers, has been a useful tool in CNNs since the introduction of residual connections in ResNet \cite{skip_resnet}. In the context of GNNs skip connections are especially useful for tasks that show homophily, i.e. where each node is strongly related to the features of its local neighborhood \cite{graph_representation_learning}. They are also used to prevent oversmoothing and vanishing gradients \cite{introduction_to_graph_neural_networks}.

Residual connection is the simplest form of skip connection, where layer implements $f(x) + x$ instead of plain $f(x)$. The authors of \cite{skip_resnet} hypothesize that training networks with access to identity mapping makes it easier to train the network. This notion has been formalized and explained in \cite{skip_resnet_visualization}, where it is shown that using residual connections greatly simplifies the surface of the loss function and therefore greatly reduces the number of local minima, making training easier. In GNNs this approach has been used e.g. in GraphSAGE \cite{convgnn_graphsage} and ResGCN \cite{skip_deep_gcn}.

However, many layers may contain different information useful for final classification. For example, early layers contain more \q{raw} features, since they cover a small graph radius, using just the node and its direct neighborhood. On the other hand, layer layers are more smoothed and cover larger subgraphs, with larger radii. We may want to capture structure information at many layers of smoothing. Simple residual connections cannot solve this problem, since even when a particular layer uses a skip connection and sends an unchanged $x$ signal further, all subsequent layers must use this output, i.e. any later representation cannot decide to use skipped or non-skipped outputs from previous layers.

The problems outlined above have been studied in \cite{skip_jumping_knowledge}, where \textbf{Jumping Knowledge (JK)} model is proposed. Authors show that GNNs have problems with varying neighborhood sizes and differentiating nodes with varying structural roles in the graph. They propose a more adaptive scheme, where networks can learn representations of different order for particular graph substructures. This is done through a type of skip connection called jump connection, where the output of each layer is passed directly to the end, just before the readout. This way we have access to the rich information about different graph substructures, and can also adapt the effective size of the neighborhood for each node. Three schemes of aggregation are proposed:
\begin{enumerate}
    \item Max pooling - for each node, we perform global max pooling, i.e. we gather its representations at different layers and select the maximal value for each feature among all layers. This should maximize the usage of both local features from early layers and more globally aggregated information from later layers. This scheme is node adaptive and has a very low computational cost. Note that other global pooling functions can also be used in a very similar way.

    \item Concatenation - a very straightforward, but not injective, way of combining vectors is concatenating them. This approach is not node adaptive, i.e. it does not differentiate between individual node structural roles well. However, it optimizes for combining the subgraph features, which may work better for small and more regular graphs. It has higher computational cost than max variant, since later MLP network has to work with larger input.

    \item LSTM pooling - a bidirectional LSTM with an attention mechanism is used to learn attention scores, and features at different levels are combined using those weights. It is a node adaptive method, as attention scores are different for each node. This approach is particularly effective for large, complex graphs, but is susceptible to overfitting. The computational and memory costs are high, since a complex bidirectional RNN with backpropagation through time has to be computed.
\end{enumerate}

\begin{figure}[h]
    \makebox[\textwidth][c]{\includegraphics[width=0.5\textwidth]{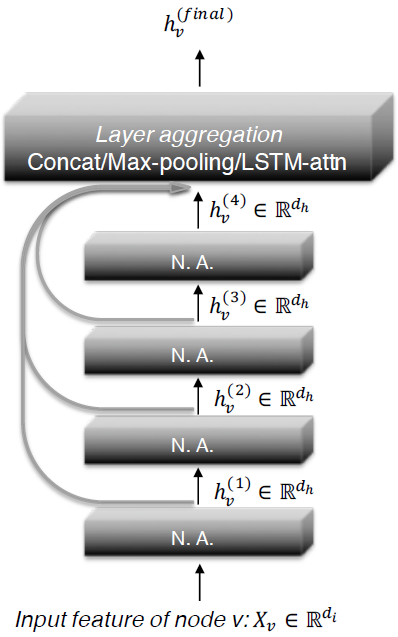}}
    \caption{The visualization of Jumping Knowledge architecture. \q{N.A.} stands for neighborhood aggregation, i.e. graph convolution. For graph classification, readout has to be applied after the last layer shown. Source: \cite{skip_jumping_knowledge}}
    \label{skip_jumping_knowledge_graphics}
\end{figure}

Since jumping connections excel at effectively combining representations at different scales, they are of great importance for graph classification, since it relies on extracting hierarchical information about the graph substructures. It allows merging of both more local and global information for each node.

GIN \cite{gnn_GIN} is based on differentiating graph substructures, and for this reason, should greatly benefit from jumping connections. Summation is used as a readout for each layer, and JK concatenation scheme is used.

\subsubsection{Baseline architectures}
\label{section_theory_gnns_baselines}

Having proper baselines is very important for model assessment. Model performance does not exist in a void, it should be always compared to both simple and strong baselines. This way we can understand how much improvement we actually obtained using new methods.

Baselines for graph classification have been created over the years. They are based on the observation that graph data is very unique, consisting of both node features and graph structural information, and therefore requires dedicated algorithms. While they are simple, they often achieve results comparable with GNNs, or even outperform them. This signifies the importance of datasets sufficiently large for GNNs, proper architectures able to extract task-relevant features, and hyperparameter tuning, to be able to outperform baselines.

In \cite{evaluation_gnn_fair_comparison}, two approaches have been proposed. \textbf{Molecular Fingerprints (MFPs)} is a simple, feature-based baseline aimed at chemical graph classification. It is based on molecular fingerprints, a group of methods used in chemistry to vectorize molecules; those classical methods, however, use hand-crafted features, while this baseline is learnable end-to-end since it is similar to a very simple GNN. The algorithm applies global sum pooling and then applies a single layer MLP with ReLU activation. This works particularly well for molecular datasets since they typically use one-hot encoded features, where each feature is an atom type. The sum of such features is injective, and the sum is simply the count of particular atoms in a whole molecule graph. This approach should not be mistaken with molecular fingerprints for feature extraction, which are graph descriptors and may use other classifiers than MLP. Another proposed technique, inspired by deep learning on sets, is called \textbf{Deep Multisets}. It is based on learning permutation-invariant functions on node sets and is aimed at datasets with no node features (e.g. social datasets) or with continuous features (e.g. protein datasets). First, node vectors are transformed with a single layer MLP, followed by global sum pooling, and another single-layer MLP for classification. Those approaches do not leverage graph topology at all. For this reason, if GNNs are unable to match or outperform those baselines, it means that either the structural information is not very important for the problem, or that the architecture is not able to use it properly. As shown in \cite{evaluation_gnn_fair_comparison}, most architectures struggle to match the results of those methods on popular, small datasets.

\subsection{Graph descriptors}
\label{section_theory_descriptors}

A great advantage of GNNs is using node features and smoothing neighborhood representations. However, they do not make explicit use of graph topology, whereas graph kernels use graph topology exclusively and are able to achieve a good performance, sometimes surpassing GNNs \cite{evaluation_gnn_fair_comparison,gnn_local_degree_profile}. To explicitly describe the graph structure, node, edge, and graph-level descriptors can be used, or domain-specific descriptors such as molecular fingerprints. For graph classification, node and edge descriptors are aggregated, typically as histograms. LDP \cite{gnn_local_degree_profile}, described in Section \ref{section_theory_gnns_baselines}, is a descriptor-based method. It also shows that features can be calculated not only for nodes themselves but also for their neighborhoods, better describing their structural roles in the graph. The authors of LDP also mention that they experimented with a few other node-level and edge-level descriptors, but did not get a consistent improvement (e.g. closeness centrality or Fiedler vector). A particularly important group of features are centrality measures, which can describe the importance of specific nodes, e.g. the existence of large central stems in molecules, from which the branching parts start.

\textbf{Local Degree Profile (LDP)} \cite{gnn_local_degree_profile} has been designed as a baseline for GNNs to be measured against. It is a graph descriptor-based method, leveraging only simple topology-based features describing the local neighborhood of vertices. For each node, 5 simple structural features are computed: degree, minimal neighbor degree, maximal neighbor degree, average neighbor degree, and standard deviation of neighbors' degrees. Those features summarize the degree information about the node and its neighborhood. Each feature is then aggregated with either a histogram (number of bins is a hyperparameter) or an empirical distribution function (EDF). This creates a graph-level summary, which is then classified with an SVM (of course, any other algorithm can be used here). The authors also experiment with adding all-pairs shortest paths aggregation to the final vector, getting about 2\% improvement on chemical datasets. Open source code for LDP \cite{local_degree_profile_github} also shows that the authors added the sum of neighbors' degrees as a feature, which was not described in the paper. This method is shown to be able to closely match or outperform (sometimes quite significantly) graph kernels and GNNs.

\textbf{Clustering coefficient} \cite{descriptor_vertex_clustering_coeff} measures neighborhood connectivity, indicating how close is it to a complete graph. For a single node $v$ it is defined as a proportion of the number of edges in its neighborhood to the total possible number of edges that could exist in that neighborhood:
\begin{equation}
Clust_{coef}(v) = \frac{2 * |\{ e_{ij} \}|}{d(v) (d(v) - 1)},
\end{equation}

\noindent
where $i, j \in N(v)$, $d(v) = |N(v)|$, $e_{ij} = (i, j) \in E_G$ are edges in the neighborhood of $v$. Alternatively, we can define it as a fraction of possible triangles through a node \cite{czech_doktorat}:
\begin{equation}
Clust_{coef}(v) = \frac{2T(v)}{|N(v)| (|N(v)| - 1)},
\end{equation}

\noindent
where $T(v)$ is the number of triangles through a node $v$.

\textbf{Betweenness centrality} \cite{descriptor_betweenness_centrality} measures how much influence a node has over the flow of information in a graph. It finds \q{bridge} or \q{central} nodes. For node $v$ is defined as a fraction of all-pairs shortest paths that go through $v$:
\begin{equation}
C_B(v) = \sum_{s,t \in V} \frac{\sigma(s, t|v)}{\sigma(s, t)},
\end{equation}

\noindent
where $\sigma(s, t)$ is the number of shortest paths between nodes $s$ and $t$, and $\sigma(s, t)$ is the number of those paths that pass through node $v$.

\textbf{Closeness centrality} \cite{descriptor_closeness_centrality} is a measure of node centrality. It measures the closeness of a node to all other nodes in terms of shortest paths lengths and is defined as:
\begin{equation}
C_C = \frac{|V| - 1}{\sum_{u \in V} dist(u, v)}.
\end{equation}

\textbf{PageRank} \cite{descriptor_vertex_pagerank} is a well-known descriptor measuring the influence of nodes in the network. Originally made for ranking websites, it assigns high influence to nodes connected to other important nodes, measuring the \q{core} elements of the graph. For undirected graphs, it is defined as:
\begin{equation}
PR(v) = \frac{1-d}{|V|} + d \sum_{u \in N(v)} \frac{PR(v)}{d(v)},
\end{equation}

\noindent
where $d$ is a damping factor, a constant of value typically around $0.85$.

\begin{figure}
    \makebox[\textwidth][c]{\includegraphics[width=0.5\textwidth]{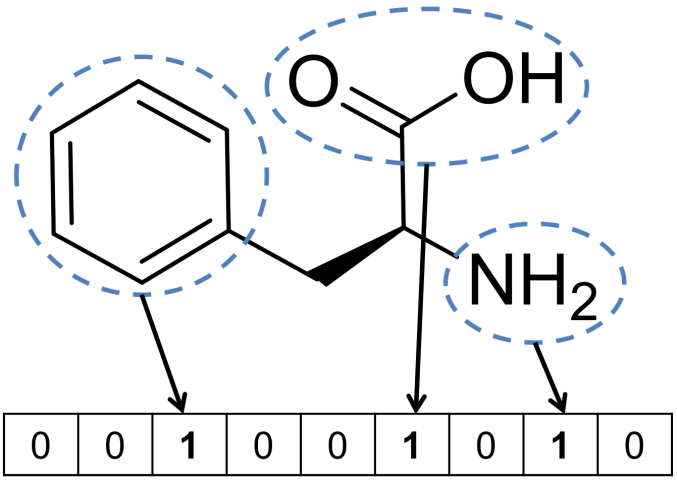}}
    \caption{General scheme of hashed binary fingerprints. Subgraphs are identified, hashed and the corresponding bit in a bit vector dictionary is set. Source: \cite{fingerprints_general_graphics}}
    \label{fingerprint_image}
\end{figure}

\textbf{Extended Connectivity Fingerprints (ECFP)} \cite{fingerprints_ECFP}, also called circular fingerprints or Morgan fingerprints, is a molecule descriptor calculated using hashing graph substructures. Hydrogen atoms and bonds to hydrogens are ignored during calculation. Firstly, each atom in a molecule gets a unique integer identifier, using e.g. number of \q{heavy} (non-hydrogen) neighbors, the atomic number, or atomic mass. Then a given number of rounds of iterative updating happen (see Figure \ref{ecfp_graphics} for visualization), up to a given radius $k$, which is typically 2 or 3 (resulting in ECFP4 and ECFP6 fingerprint, respectively, since traditionally diameter is used instead of radius in the name). In each round, an extended neighborhood is considered, and all unique substructures are given distinct identifiers. This process is quite similar to the Weisfeiler-Lehman isomorphism test since it is a variant of Morgan's algorithm, which was created to solve the molecule isomorphism problem (similar to the graph isomorphism problem of WL-test). Finally, all identifiers are hashed into a fixed-length vector. There are two variants of ECFP fingerprints: bit vector (binary), where only the existence or absence of a feature at a given position is saved, and count vector, where all occurrences are accounted for. Note that due to hash collisions this mapping is not unequivocal and reversible, but this is not a problem in practice for graph classification. This algorithm is able to detect large and complex subgraph structures, especially in ECFP6 variant (with radius 3).

\begin{figure}
    \makebox[\textwidth][c]{\includegraphics[width=0.5\textwidth]{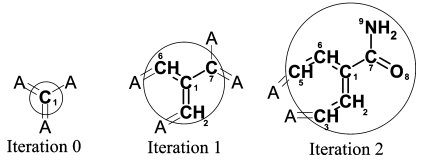}}
    \caption{Visualization of ECFP fingerprint. Each iteration increases the explored subgraph radius by 1, up to a given maximum. Source: \cite{fingerprints_ECFP}}
    \label{ecfp_graphics}
\end{figure}

\begin{figure}
    \makebox[\textwidth][c]{\includegraphics[width=0.75\textwidth]{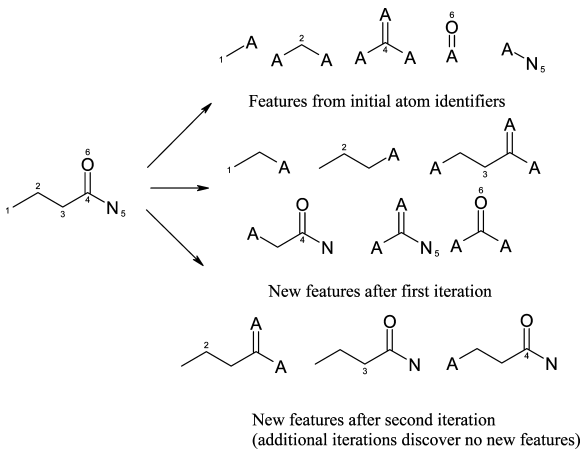}}
    \caption{Features discovered by ECFP algorithm for butyramide molecule. Source: \cite{fingerprints_ECFP}}
    \label{ecfp_graphics}
\end{figure}

\textbf{RDKit fingerprints} \cite{fingerprints_RDKit,fingerprints_RDKit_code, fingerprints_RDKit_presentation} are inspired by publicly available descriptions of a proprietary Daylight fingerprint. They are hashed substructure fingerprints, similar to Morgan fingerprints - first a set of subgraph structures is identified, then they are hashed into a fixed-length vector. It can be a bit or count vector, similarly to ECFP. First, all subgraphs in a molecule are identified, enumerating them as paths up to a given length, typically between 5 and 7. Each subgraph is hashed and a relevant bit in a dictionary is set (or count is increased). Individual subgraphs are hashed based on their bonds, using atomic numbers, atom degrees, bond types, and atom and bond aromaticity. In comparison to atom-centric ECFP fingerprints, the RDKit fingerprints generate subgraphs using walks, which enables them to discover and remember structures with more complex topology, such as path-like subgraphs.

\begin{figure}
    \makebox[\textwidth][c]{\includegraphics[width=\textwidth]{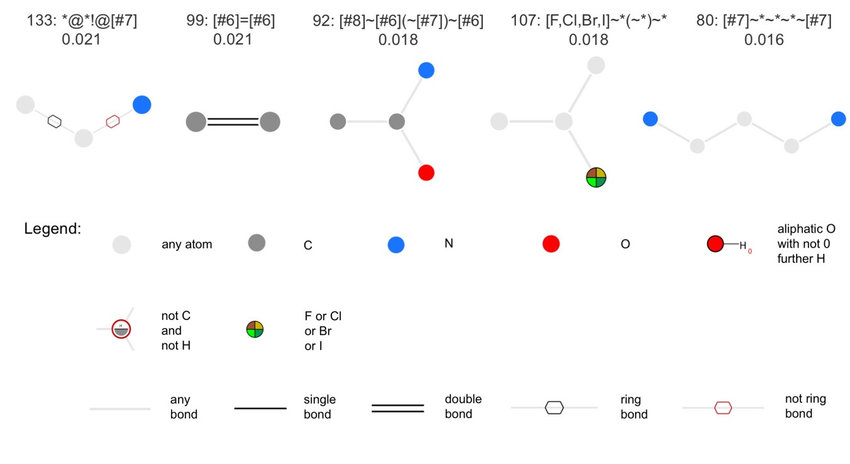}}
    \caption{Visualization of selected MACCS keys using SMARTSviewer. Above each diagram, the first line reports the index of the respective MACCS key and its SMARTS pattern. Source: \cite{fingerprints_MACCS_graphics}}
    \label{MACCS_graphics}
\end{figure}

\textbf{MACCS keys} \cite{fingerprints_MACCS}, also known as MDL keys \cite{fingerprints_cheminformatics_book} (after the company that developed them), is a set of hand-crafted binary features. They consist of 166 descriptors that have been selected using expert knowledge, which are detected in a molecule, creating a fixed-length bit vector (0 for the absence of a feature, 1 for existence). Those are structural keys, i.e. they are not hashed like ECFP fingerprints. Features are mostly defined using SMARTS patterns \cite{fingerprints_SMARTS}, which are similar to regular expressions (regexes) but targeted towards SMILES strings. Calculating them is available as a part of multiple open source packages, which also list SMARTS patterns, e.g. in RDKit \cite{fingerprints_MACCS_code}. Apart from a few atom counts (e.g. iodine I or sulfur S), a vast majority are groups and hierarchical structures, e.g. carbonyl groups C=O (common in organic compounds and typically a part of larger functional groups) or aromatic rings (participating in many organic reactions). This means that this descriptor can detect predetermined hierarchical structures that can be problematic to catch for other methods, since the chances of learning topologically complex or large substructures are low for limited datasets. For examples of some MACCS keys and their SMARTS patterns, see Figure \ref{MACCS_graphics}.

Using any descriptor-based method basically performs feature extraction for a graph, generating a vector - graph embedding. While in GNNs the classification of those embeddings is done through MLP, because of gradient flow and end-to-end learning, in descriptor-based methods it is not required. We can use any type of classifier, and in practice \textbf{Random Forest (RF)} is a very popular choice. It typically gives very good results and is not very sensitive to hyperparameters choice, which reduces the computational cost of hyperparameter optimization. It also handles high-dimensional vectors generated by Morgan or RDKit fingerprints well. It is very scalable, as all trees can be trained in parallel, which is a considerable advantage over another popular choice, kernel SVM, which has computational complexity between $O(^2)$ and $O(n^3)$. Similarly to neural networks, class weighting scheme can be used in RF, in two modes: globally, when class weight is computed from its proportion in the whole datasets, or adaptively per subsample, when we compute class weights separately in each tree based on the proportion of classes in its bootstrap sample. The minimal number of samples to perform a split can be tuned to prevent overfitting.

\subsection{Evaluation methodology}
\label{section_theory_evaluation}

Every proper evaluation of the machine learning model requires dividing the available dataset into three parts: training, validation, and testing. We train the model on the training set, making the labels available during training. A validation set is used for model selection, checking the model's performance with given hyperparameters, and labels are not used to train the model, but rather to steer the process of tuning and prevent overfitting to the training data. The test set is used only once, after the entire training and model selection, to assess the performance of the final model. The matter of how to choose those sets is not straightforward, since while properties of the particular dataset influence this choice, we should use the same benchmarking process for all datasets for a fair comparison.

The two most popular approaches are the holdout method and cross-validation (CV) \cite{ESL}. In the \textbf{holdout method}, we simply divide the available dataset randomly into training/validation/test parts, typically in proportions $80\% / 10\% / 10\%$. This works well for large datasets, where the number of test samples is large enough to get a proper evaluation of the model. \textbf{Cross-validation (CV)} divides the dataset randomly into $K$ equal-sized, non-overlapping parts called folds. A fold acts as a test set, and all the other parts are merged into a training set, and this process is repeated for all folds. This way we have $K$ test scores. A validation set is created each time from the training set, and we can use either holdout or another CV, where $K'$ folds are validation sets. The latter approach is called \textbf{nested CV}. Cross-validation has a high computational cost, as $K$ models have to be trained (or $K * K'$ for nested CV), and for this reason, it is used only for the small datasets. It is also recommended in those cases since the single test set would be too small and would give biased results. In addition, using CV allows for computing the standard deviations of the test results providing a practical measure of the model stability.

\begin{figure}
    \makebox[\textwidth][c]{\includegraphics[width=\textwidth]{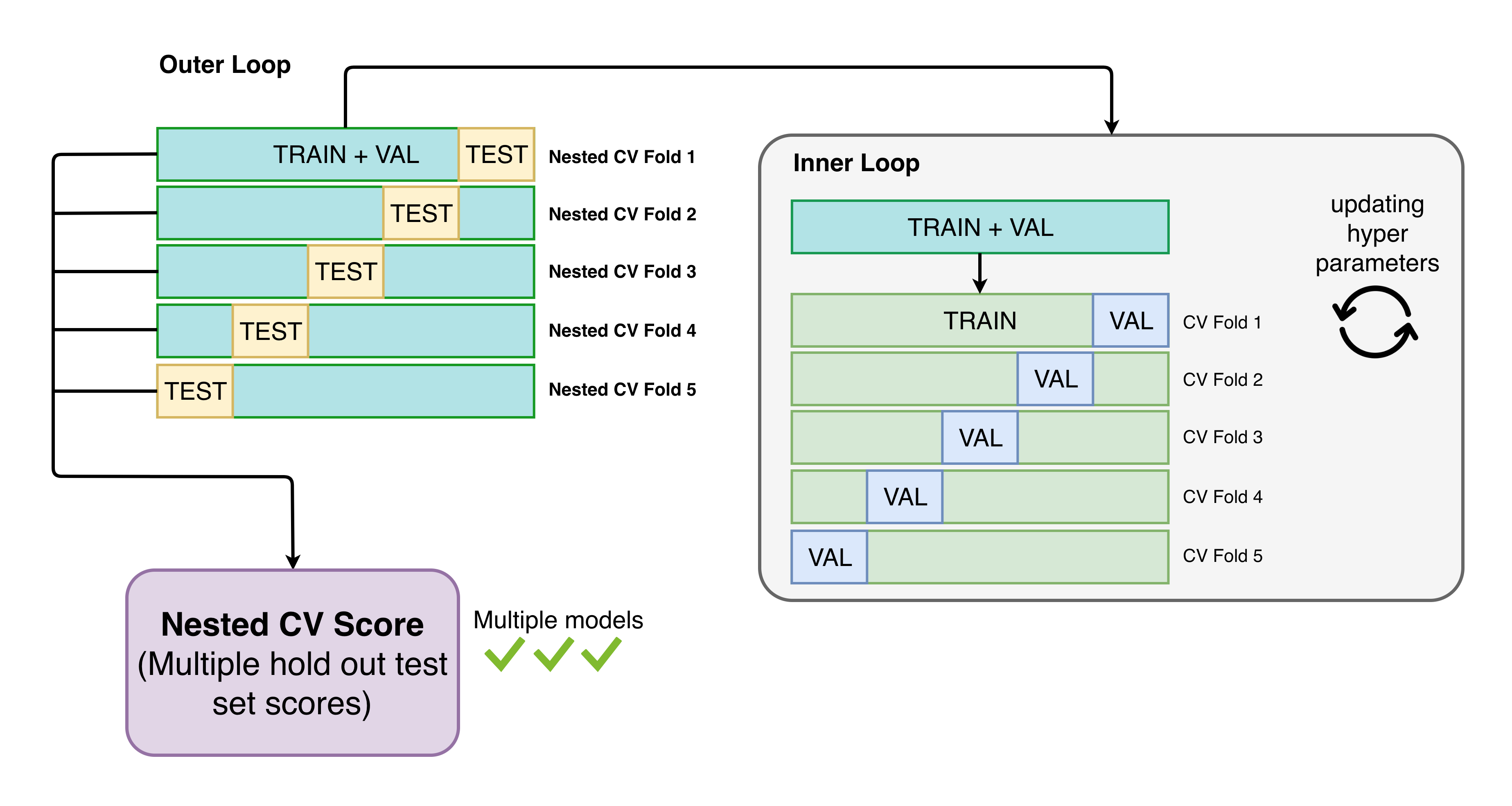}}
    \caption{Visualization of nested cross-validation procedure. The source of complexity, nested loop over folds, is clearly visible. This is 5 outer folds, 5 inner folds variant, resulting in training 25 models. Source: \cite{nested_cv_graphics}}
    \label{nested_cv_graphics}
\end{figure}

In the literature on graph classification, both approaches were used. The reason for using cross-validation is that older graph classification datasets are quite small, having from as few as 180 samples in the case of the MUTAG \cite{datasets_MUTAG} dataset, up to about 1000 samples, e.g. in the case of the PROTEINS \cite{datasets_PROTEINS} dataset. However, in some works only validation scores are provided, e.g. in GIN paper \cite{gnn_GIN} (10-fold CV). Such results are always biased, giving more optimistic results than on a proper test set. This is a major problem in comparing the results between works, as the authors of \cite{evaluation_gnn_fair_comparison} have highlighted.

Small datasets cause many issues. Having to use cross-validation, or nested cross-validation, may increase the computational cost so much that the training and evaluation procedure takes as long as for large datasets. At the same time, the small sizes of single validation and test sets make the model comparison hard. For example, if the test set consists of 100 samples (which for many older datasets is quite normal), each sample changes accuracy by $1\%$. For a perspective, in graph classification, the gain of $2-3\%$ accuracy is often a considerable improvement. Smaller training sets mean easy overfitting, penalizing more sophisticated and often very interesting architectures. They also require shorter training, which makes the initial random initialization of weight much more important. Many of those problems have been described in recent works \cite{datasets_moleculenet,evaluation_gnn_fair_comparison,datasets_ogb}. The authors of \cite{evaluation_gnn_fair_comparison} propose a pipeline for evaluation, using static training/validation/test splits, CV for testing, and holdout for validation. Additionally, in order to minimize the effect of unlucky random initialization, they retrain the model 3 times for testing, and for each test fold report the average of those three runs. We argue that this tries to fix the symptoms, not the problem - too small datasets. Large graph classification datasets, consisting of thousands or tens of thousands of graphs, are currently available, e.g. many datasets from the TUDatasets collection \cite{datasets_TUDataset}, molecular benchmark MoleculeNet \cite{datasets_moleculenet}, and Open Graph Benchmark (OGB) \cite{datasets_ogb}. Those datasets often come with pre-selected validation and test sets, offering a unified, fair comparison between models. They typically use the holdout method, since they are large enough.

Holdout and CV are general machine learning tools for model assessment. Those methods, using randomized division into subsets, assume the i.i.d. property, i.e. that samples (data points) are independently and identically distributed. Since they have been created for tabular data, which typically has this property, it works well in general. However, this often does not work that well for molecular property prediction, which is probably the most common task in graph classification, as shown by the authors of MoleculeNet \cite{datasets_moleculenet}. Purely random division, used e.g. in MUTAG or PROTEINS datasets, may cause very similar molecules to be in both training and validation/test set, which makes the task easier.

In real-world applications, when the laboratory experiments are used to verify the ground truth for new molecules, a \textbf{time split} is used - a dataset created up to a given date is a training set, and newer data serves as a test set. The validation set is typically the part of the newest data from the training set. This also causes a phenomenon known as a distribution shift, where newer data follows a slightly different distribution compared to the historical data. This is unavoidable in real-world datasets, and as such should be reflected in model evaluation. However, this approach requires time labels to be available for all molecules, which is almost never the case. In MoleculeNet, which is already a large and varied collection, only a single dataset has this information available.

Instead, we can use a \textbf{scaffold split} \cite{evaluation_scaffold_split} (specifically Bemis-Murcko scaffolds), which takes into consideration the structure of molecules. A molecular scaffold reduces the chemical structure of a compound to its core components, essentially by removing all side chains and only keeping ring systems and parts that link together ring systems. Data is divided into equivalence classes, i.e. groups of samples, where each group has the same scaffold. Training, validation, and test sets are created by merging those groups so that all molecules with a particular scaffold are in only one subset. This creates an out-of-distribution prediction problem, where the model is forced to learn the underlying properties of molecules to achieve good test performance because the distribution of training and test data is sufficiently different.

\begin{figure}
    \makebox[\textwidth][c]{\includegraphics[width=0.75\textwidth]{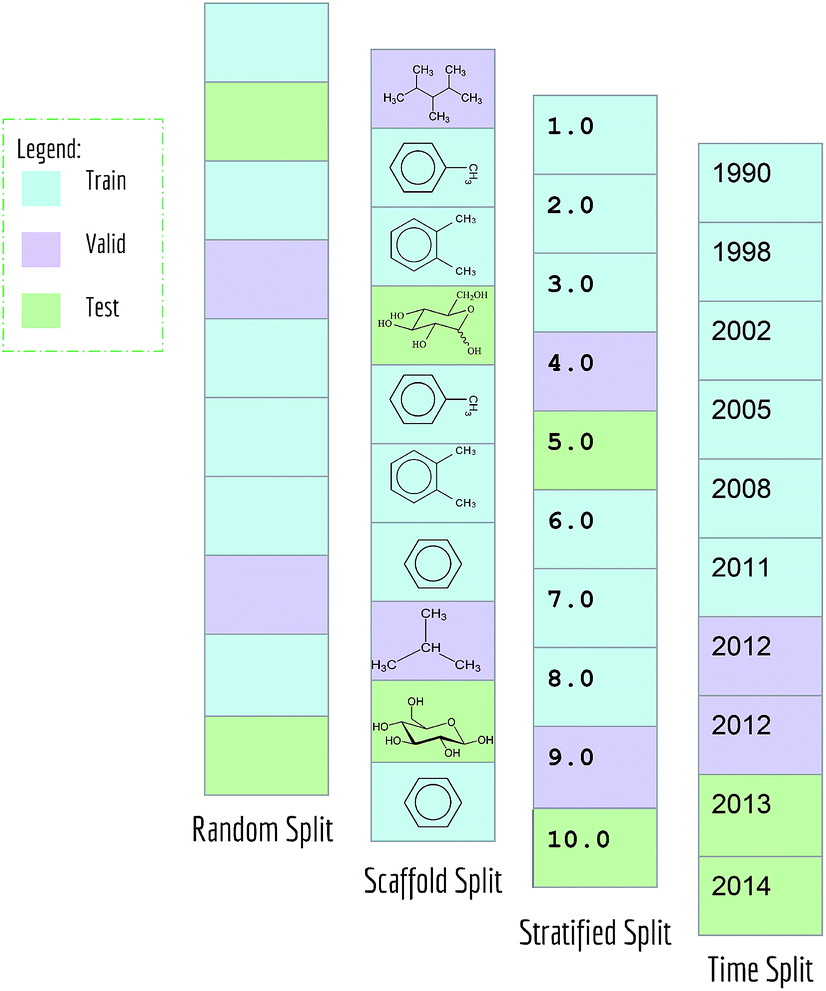}}
    \caption{Comparison of train-validation-test split methods. Random split disregards the nature of the data. Scaffold split relies on chemical structure of molecules. Stratified random sampling for regression and time split are shown, but are not used in this thesis. Source: \cite{datasets_moleculenet}}
    \label{split_methods_comparison_graphics}
\end{figure}

Graph classification datasets are often highly imbalanced, i.e. the number of samples from different classes is very different. It is not uncommon for binary classification to have $90\%$ of the samples from the negative class (e.g. non-active compound) and just $10\%$ from the positive class (e.g. compound active against HIV). Additionally, we are often interested in the minority class. This requires the appropriate design of algorithm, training/validation/test splitting, and metrics used for evaluation. Neural networks of any kind naturally lend themselves to imbalanced problems, as they minimize loss function, which can be weighted by class proportion, so that we put more weight on classes with fewer samples. Random sampling would use the majority class, threatening that validation or test sets could even not contain the minority class at all. For this reason, \textbf{stratified sampling} is used in those cases. It ensures that while samples are random, the class distribution is preserved.

The model selection is typically done through loss minimization since during training neural networks minimize it and as such it's the most natural way. However, on test set we are interested in measuring the actual performance of the model, e.g. how good is it in classifying previously unseen samples. Different metrics measure the quality of the model in different ways. In particular, we are interested in metrics that work well in the case of imbalanced datasets. Many such metrics can be used, but the most common is the \textbf{Area Under Receiver Operating Characteristic Curve (\text{AUROC}, ROC AUC, or AUC)}. For binary classification with positive class $1$ and negative class $0$, we denote:
\begin{itemize}
    \item $P$ - number of positive class samples
    \item $N$ - number of negative class samples
    \item $TP$ - true positives, the number of samples from positive class with model prediction $1$
    \item $TN$ - true negatives, the number of samples from positive class with model prediction $0$
    \item $FP$ - false positives, the number of samples from the positive class incorrectly classified as $0$ by the model
    \item $FN$ - false negatives, the number of samples from the negative class incorrectly classified as $1$ by the model
\end{itemize}

The AUROC is defined as an area under the ROC curve, which is created by plotting the recall, or true positive rate (TPR), against the false positive rate (FPR), at various positive class probability thresholds. It measures the probability that a randomly chosen positive class sample will be assigned a higher probability (of being from the positive class) than a randomly chosen sample from the negative class. TPR and FPR are defined as:
\begin{equation}
\begin{split}
& TPR = \frac{TP}{P} = \frac{TP}{TP + FN} \\
& FPR = \frac{FP}{N} = \frac{FP}{FP + TN}
\end{split}
\end{equation}

The AUROC metric has many desirable properties for measuring efficiency of graph classification models on imbalanced datasets. It is a single number, which makes it a convenient measure to compare models, compared to e.g. ROC or PRC (Precision-Recall) curves. Additionally, it is sensitive to the class imbalance that may be present in the dataset, compared to e.g. accuracy. For example, for a dataset with 90\% negative class and 10\% positive class, accuracy will be 90\%, which does not account for class imbalance in any way, while AUROC will be low if the model does not discriminate the positive class well. At the same time, it is not biased towards models that perform well on the minority class at the expense of the majority class \cite{imbalanced_learning_book}.

Calculating the AUROC value requires outputs of positive class probability, not only a binary 0/1 decisions from the classifiers. This is not a major limitation, however, since most algorithms support outputting probabilities, e.g. neural networks, Random Forests or SVMs (through Platt scaling). Moreover, this requires the classifier to not only output good class values, but also to be sure of its predictions, i.e. output low probabilities for negative class samples and high probabilities for positive class samples.

\newpage

\section{Experiments, results, and discussion}
\label{section_experiments}

In this section, we present the experiment design, results, and discussion. Experimental setup, including datasets, models and their configurations, is described in Section \ref{section_experiments_setup}. Results are presented and discussed in Section \ref{section_experiments_results_and_discussion}.

\subsection{Experimental setup}
\label{section_experiments_setup}

\subsubsection{Datasets}
\label{section_experiments_setup_datasets}

\begin{figure}[h!]
\makebox[\textwidth][c]{\includegraphics[width=0.94\textwidth]{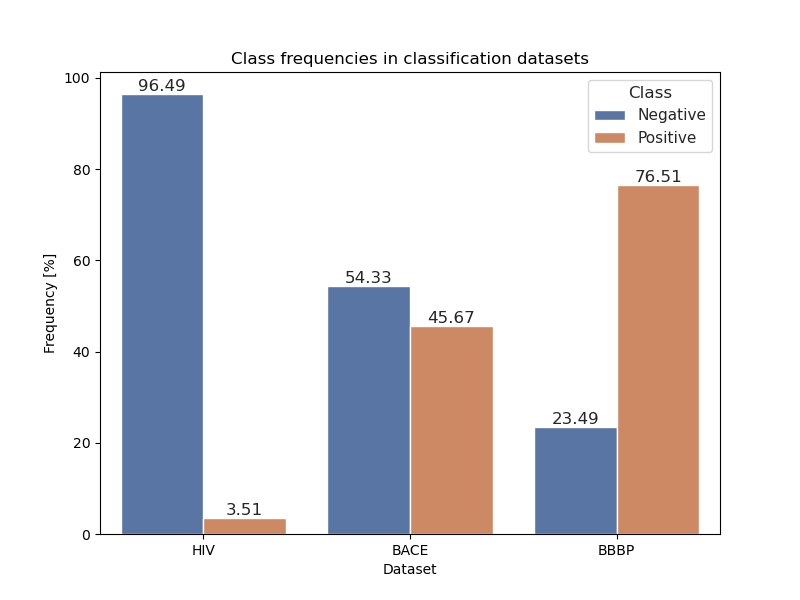}}
\caption{Positive and negative class percentages in classification datasets used.}
\label{datasets_classification_graphics}
\end{figure}

\begin{figure}[h!]
\makebox[\textwidth][c]{\includegraphics[width=0.94\textwidth]{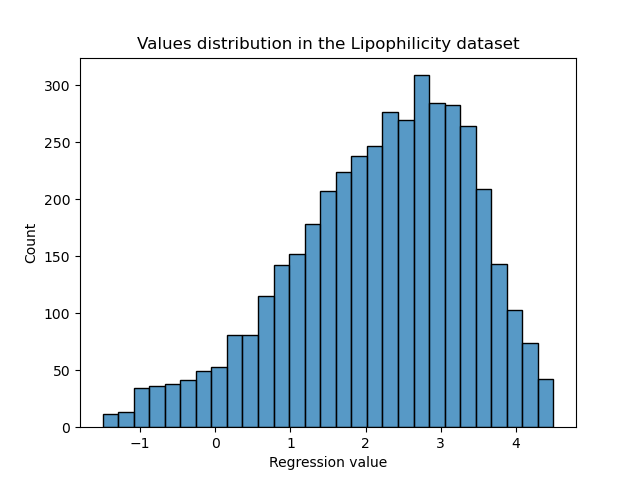}}
\caption{Regression values distribution in the Lipophilicity dataset.}
\label{datasets_regression_graphics}
\end{figure}

Because of the problems with small datasets described in Section \ref{section_theory_evaluation}, we decided to use datasets from MoleculeNet \cite{datasets_moleculenet}. The datasets from MoleculeNet have reasonable size and have predetermined training, validation and test split, available through the OGB interface. Those splits are often calculated using scaffold splitting, making them more challenging. Additionally, we limit ourselves to the datasets having at least 1500 graphs. This allows us to use the holdout method, without incurring the large computational cost of cross-validation. We select a varied collection of datasets. Their basic statistics are summarized in Table \ref{table_datasets}. The descriptions below closely follow MoleculeNet paper \cite{datasets_moleculenet}. The visualizations of class frequency (for classification datasets) and target value distribution (for regression dataset) are provided in Figure \ref{datasets_classification_graphics} and Figure \ref{datasets_regression_graphics}.

\begin{table}[h!]
\centering
\begin{tabular}{|p{2cm}|l|l|l|l|l|l|} 
\hline
\textbf{Category} & \textbf{Dataset} & \textbf{\# tasks} & \textbf{\# graphs} & \textbf{Task} & \textbf{Split} & \textbf{Metric}   \\ 
\hline
Physical \newline chemistry          & Lipophilicity & 1        & 4200      & Regression     & Random   & RMSE     \\ 
\hline
\multirow{2}{*}{Biophysics} & HIV           & 1        & 41127     & Classification & Scaffold & ROC-AUC  \\ 
\cline{2-7}
                            & BACE          & 1        & 1513      & Classification & Scaffold & ROC-AUC  \\ 
\hline
Physiology                  & BBBP          & 1        & 2039      & Classification & Scaffold & ROC-AUC  \\
\hline
\end{tabular}
\caption{Datasets used and their basic statistics, taken from \cite{datasets_moleculenet}.}
\label{table_datasets}
\end{table}

\textbf{Lipophilicity} dataset aims to predict lipophilicity - the ability of a chemical compound to dissolve in fats and oils. It is important for drugs, as it affects solubility and membrane permeability. The measure to predict is the octanol/water distribution coefficient $\log{D}$ at pH $7.4$. It uses random split, as it is a regression dataset.

\textbf{HIV} dataset presents the task of predicting whether the molecules are active against HIV. Molecules inhibiting HIV replication are crucial for developing anti-HIV drugs. Scaffold split is used, as we are interested in discovering new HIV inhibitors, and therefore have to evaluate previously unseen molecular structures. It is by a large margin the largest and the most imbalanced dataset used.

\textbf{BACE} dataset provides binary binding/non-binding results for a set of inhibitors of human $\beta$-secretase 1 (BACE1). Drugs that block this enzyme (BACE inhibitors), as they could prevent the buildup of beta-amyloid peptides in neurons, which are the major drivers of Alzheimer’s disease pathogenesis \cite{alzheimer_disease}. Scaffold split is used, for the similar reason as in HIV dataset.

\textbf{BBBP} dataset concerns prediction of the blood-brain barrier penetration (BBBP) ability of small molecules. The membrane separating circulating blood and brain extracellular fluid blocks most drugs, hormones, and neurotransmitters. Any drugs that aim to directly affect the central nervous system have to penetrate that barrier. Scaffold splitting is also used, similarly to two previous datasets.

\newpage

\subsubsection{Data preprocessing}
\label{section_experiments_setup_preprocessing}

Since GNNs work by neighborhood aggregation, they operate on node features. For this reason, choosing proper featurization is essential for GNN performance. Fortunately, for chemical datasets many atomic features have been studied and have become a standard in molecular graph classification, in part because of the OGB initiative \cite{datasets_ogb}. They are one-hot encoded, since neighborhood aggregation works feature-wise, and also because GIN makes such assumption about node features \cite{gnn_GIN}. Those features can be easily calculated by standard RDKit software \cite{rdkit} based on SMILES strings. The full list of features is available online as part of OGB package \cite{ogb_features} and they have been summarized in Table \ref{table_gnn_molecular_features}.

\begin{table}[h!]
\begin{tabularx}{\textwidth}{|l|X|}
\hline
\textbf{Feature}                        & \textbf{Values}                                       \\ \hline
Atomic number                           & Integer in $[1, 118]$ or misc                       \\ \hline
Chirality                               & Tetrahedral-CW, tetrahedral-CCW, unspecified or other \\ \hline
Degree                                  & Integer in $[0, 10]$ or misc                        \\ \hline
Formal charge                           & Integer in $[-5, 5]$ or misc                        \\ \hline
Number of connected  hydrogen atoms     & Integer in $[0, 8]$ or misc                         \\ \hline
Number of radical electrons             & Integer in $[0, 4]$ or misc                         \\ \hline
Hybridization                           & SP, SP2, SP3, SP3D, SP3D2 or misc                     \\ \hline
Aromaticity                             & True or false                                         \\ \hline
Part of a ring                          & True or false                                         \\ \hline
\end{tabularx}
\caption{Features extracted in OGB for molecular datasets, taken from \cite{ogb_features}.}
\label{table_gnn_molecular_features}
\end{table}

Nevertheless, using those raw features would result in huge dimensionality and extremely sparse features. In many datasets there would be potentially many constant features, as, e.g. molecules with very high degree atoms (9 or 10) are rare. For this reason, we can instead embed those features, similarly to embedding input words in NLP models \cite{ogb_encoder,pytorch_embeddings}. This results in smaller, more dense input vectors. Features can be embedded independently and summed, creating a fixed length, continuous representation \cite{ogb_encoder}. In PyTorch, the embedding class also linearly transforms features, which allows creating a weighted summation, where weights are learned through backpropagation.

For all GNN models, we use embedded atomic features. Embedding dimensionality is the same as the number of convolutional channels for a given model. Molecular Fingerprints and Deep Multisets have been proposed for datasets with pre-calculated features \cite{evaluation_gnn_fair_comparison}, so we check raw features, but we also propose using embedded features and evaluate this approach, for the reasons outlined above. While this approach results in lower interpretability in case of MFPs, it may improve the classification results.

We do not perform any other kind of preprocessing. We use dataset splits as provided by OGB data loaders, without filtering any molecules.

\subsubsection{Architectures}
\label{section_experiments_setup_architectures}

As baselines, we use:
\begin{itemize}
    \item Molecular Fingerprints (MFPs)\cite{evaluation_gnn_fair_comparison}
    \item Deep Multisets \cite{evaluation_gnn_fair_comparison}
    \item SGC
\end{itemize}

For MFPs and Deep Multisets, we treat layer sizes in MLPs and learning rate as hyperparameters. We check MFPs and Deep Multisets both without and with atom embeddings, as separate models. For SGC, tune use the same hyperparameters as for other GNNs (see below).

We focus on four most well-known GNN architectures: GCN, GraphSAGE, GIN, and GAT. They are varied, since they follow different design principles. Firstly, we perform a set of basic experiments, checking \q{flat} architectures, i.e. just $K$ layers stacked one after another, then the readout function and MLP. The MLP is always a single linear layer, with either softmax activation (classification) or no activation (regression), which is typical for GNNs. Additionally, for all GNN-based approaches we tune the following hyperparameters: the number of convolutional layers, size (number of channels) of convolutional layers, readout function, dropout in convolutional layers, learning rate. For all other settings, we use PyTorch Geometric model defaults, e.g. ReLU activation and no normalization between layers (we found during initial experiments that adding batch normalization either does not improve performance or harms it).

We also check the Jumping Knowledge architectures. For each one of the four basic GNN architectures, we aggregate the outputs of layers with JK: max, concatenation and LSTM. We do not treat the aggregation type as a hyperparameter, but rather check 12 different architectures, e.g. GCN + JK max, GCN + JK concatenation, GCN + JK LSTM. For LSTM, we use default PyTorch Geometric model defaults.

We evaluate Local Degree Profile (LDP) as a purely topological graph descriptor approach. We check four variants:
\begin{enumerate}
    \item Original - 5 node features proposed as a baseline
    \item Extended - with neighbors degree sum and shortest path descriptors
    \item With additional features - we add clustering coefficient, betweenness centrality, closeness centrality and PageRank to the extended variant features
    \item With additional features and feature selection - we take the extended variant and treat using (or not) each one of additional 4 features as a binary hyperparameter
\end{enumerate}
We treat the number of bins and whether to normalize the histograms as hyperparameters. 

For fingerprints-based approaches, we check:
\begin{enumerate}
    \item Morgan fingerprints, hyperparameters: radius, number of bits per radius. The total number of bits passed to RDKit for fingerprint calculation is simply a multiplication of those two. We choose this approach instead of optimizing the number of bits directly because this approach gave better results in preliminary experiments.
    \item RDKit fingerprints, hyperparameters: max path length, number of bits
    \item MACCS keys, no additional hyperparameters
    \item Concatenation of Morgan, RDKit and MACCS fingerprints, simultaneously tuning hyperparameters for all of them
\end{enumerate}

Inspired by OGB leaderboard architectures and motivated by preliminary experiments, we apply subsampling the negative class in the training data for LDP and fingerprints on HIV dataset, which has very significant negative class dominance. We treat the percentage of retained negative class samples as a hyperparameter. For both LDP and fingerprints, we use Random Forest as a classifier/regressor, and we treat minimal number of samples for split as a hyperparameter. For classification, we use entropy minimization as a split criterion, and for regression we use squared error minimization. For classification tasks, we also treat the class weighting approach (none, balanced, balanced subsample) as a hyperparameter.

For both LDP and fingerprints, we also drop constant features and heavily correlated features (Pearson correlation coefficient 0.95 or higher). Feature can be constant, e.g. for MACCS keys, when a particular feature is not present in the dataset, which can happen, since they are predetermined. For concatenation of fingerprints, we can get potentially many heavily correlated features, since particular fingerprints can extract the same substructures.

We summarize hyperparameter optimization configurations in Section \ref{section_experiments_setup_selection_evaluation}. Note that for GNNs on a large HIV dataset, we check larger convolution sizes.

Out of all models in this section, our original propositions are:
\begin{itemize}
    \item SGC - as a baseline in the context of graph classification
    \item Molecular Fingerprints and Deep Multisets - the combination with node features embedding
    \item LDP - using well selected additional features, and combining it with feature selection
    \item using concatenation of multiple fingerprints - while used before for HIV dataset, there has been no through study of this method on multiple datasets
\end{itemize}

\subsubsection{Model training, selection, and evaluation}
\label{section_experiments_setup_selection_evaluation}

To implement our methods, we use:
\begin{itemize}
    \item loading MoleculeNet datasets: OGB \cite{datasets_ogb}
    \item GNNs: PyTorch \cite{pytorch} and PyTorch Geometric \cite{pytorch_geometric}
    \item LDP and graph descriptors: PyTorch Geometric \cite{pytorch_geometric}, NetworkX \cite{networkx}, and Scikit-learn \cite{scikit-learn}
    \item fingerprint-based methods: RDKit \cite{rdkit} and Scikit-learn \cite{scikit-learn}
    \item hyperparameter optimization: Optuna \cite{optuna}
\end{itemize}

To train GNNs, we optimize cross-entropy loss (classification) or mean squared error (regression), using the Adam optimizer. We train for 3000 epochs max, with early stopping, if we do not get better validation result (AUROC for classification, RMSE for regression) for set number of rounds, we end the training. The patience parameter is:
\begin{itemize}
    \item GNN models - 100 rounds
    \item MFPs and Deep Multisets - 500 rounds, we find that improves results
\end{itemize}

The batch size is:
\begin{itemize}
    \item GNN models - 64, a typical value in the literature
    \item MFPs and Deep Multisets - full batch training for all datasets except HIV, where we use 4096; we find that this greatly stabilizes training
\end{itemize}

Those settings have been inspired by the literature, especially by \cite{evaluation_gnn_fair_comparison}, and validated during preliminary experiments. For heavily imbalanced classification datasets (HIV and BBBP) we use class weighting, as implemented in the Scikit-learn library \cite{sklearn_class_weights}.

Since the size of many hyperparameter grids are large, and training the neural network-based models (Molecular Fingerprints, Deep Multisets, GNNs) takes very long time, we cannot use grid search for hyperparameter optimization. Additionally, many parameters are inherently continuous or ordered (e.g. learning rate, number of layers) and discretizing them may have a negative effect. Instead of using simple random search, we use the Bayesian optimization, which is commonly used in recent works because of the excellent performance and being able to detect good sets of parameters in just a few iterations. Specifically, we use the Tree Parzen Estimator (TPE) sampling, implemented in the Optuna \cite{optuna} library, which allows extremely efficient optimization. In the literature, using about 20-30 iterations seems the most popular, but to ensure good exploration of hyperparameter space, we use 40 trials.

For the models using Random Forest (LDP, fingerprints), the model training is extremely fast (seconds or at most minutes). Moreover, all parameters are discrete or can be easily discretized. For this reason, we use grid search in most cases, and fingerprints parameters are all categorically distributed (since we check all possibilities anyway, Optuna will not benefit from specifying more specific distribution type). The only exception is the LDP with additional features and feature selection, since the feature selection has an exponential complexity ($2^4$ possibilities for 4 features). This, in conjunction with other hyperparameters, results in a very large grid, so in this case, we use Bayesian optimization with 100 iterations.

In all cases of hyperparameter selection, we select the model with highest AUROC or lowest RMSE, depending on the dataset task.

For LDP and fingerprints models using RF, we optimize hyperparameters for feature extraction and for classifier / regressor in a 4-step process:
\begin{enumerate}
    \item We use RF with default Scikit-learn parameters (apart from split criterion) to choose optimal feature extraction (LDP or fingerprint) hyperparameters. For HIV dataset, negative class sampling rate is also tuned here.
    \item The features are calculated for the entire dataset using selected hyperparameters, creating a regular tabular dataset.
    \item We tune RF hyperparameters on transformed dataset (without modifying training / validation / test splits).
    \item The final classifier is trained on the training dataset, using 1000 trees and determined RF hyperparameters.
\end{enumerate}

Hyperparameter ranges are summarized in Table \ref{table_hyperparameters}.

\newpage

\begin{table}[h!]
\begin{tabularx}{\textwidth}{|p{2cm}|X|}
\hline
\textbf{Architecture} & \textbf{Hyperparameters} 
\\ \hline

GNN & 
\textit{Number of channels}: [64, 512] (HIV) or [64, 256] (other datasets), discrete integer distribution with step 64 \newline
\textit{Number of convolutional layers}: [1, 5], discrete integer distribution with step 1 \newline
\textit{Readout}: ["mean", "max", "sum"], categorical distribution \newline
\textit{Dropout}: [0.0, 0.3], discrete real distribution with step 0.1 \newline
\textit{Learning rate}: [1e-5, 1e-3], continuous real distribution \newline
 \newline
Only for GAT architectures: \newline
\textit{Number of attention heads}: [2, 8], discrete integer distribution with step 2
\\ \hline

Molecular Fingerprints, Deep \newline Multisets &
\textit{MLP layer size}: [64, 512], discrete integer distribution with step 64 \newline
\textit{Learning rate}: [1e-5, 1e-3], continuous real distribution \newline
\\ \hline

LDP &
\textit{Number of bins}: [10, 100], discrete integer distribution with step 10 \newline
\textit{Normalize distributions}: [false, true], categorical distribution \newline

Only for additional features and feature selection variant: \newline
\textit{Add clustering coefficient}: [false, true], categorical distribution \newline
\textit{Add betweenness centrality}: [false, true], categorical distribution \newline
\textit{Add closeness centrality}: [false, true], categorical distribution \newline
\textit{Add PageRank}: [false, true], categorical distribution \newline
 \newline
Only for HIV: \newline
\textit{Negative class sample size}: [0.3, 0.5], discrete real distribution with step 0.1
\\ \hline

Fingerprints &
For Morgan fingerprints: \newline
\textit{Radius}: [2, 3], categorical distribution \newline
\textit{Bits per radius}: [1024, 2048, 4096], categorical distribution \newline
 \newline
For RDKit fingerprints: \newline
\textit{Max path length}: [5, 6, 7], categorical distribution \newline
\textit{Number of bits}: [1024, 2048, 4096], categorical distribution \newline
 \newline
Only for HIV: \newline
\textit{Negative class sample size}: [0.3, 0.4, 0.5], categorical distribution
\\ \hline

Random \newline Forest &
\textit{Minimal samples for split}: [2, 10], discrete integer distribution with step 2 \newline
\newline
Only for classification datasets: \newline
\textit{Class weight}: [none, "balanced", "balanced\_subsample"], categorical distribution
\\ \hline

\end{tabularx}
\caption{Hyperparameters ranges and distributions.}
\label{table_hyperparameters}
\end{table}

\pagebreak
 
\subsection{The results and discussion}
\label{section_experiments_results_and_discussion}

Since the tables with full results are very large, they are presented in Appendix \ref{appendix_full_results_tables}. Here, we refer to selected, interesting subsets of results for easier analysis. They have been broken down into separate sections, as they analyze the results from different points of view. In all cases, the learning rate has been rounded for readability.

In case of GNN models with JK LSTM on HIV dataset, we could not allocate enough memory on GPU, as it required over 17 GB, even when batch size was reduced to 1. We mark them as OOM.

\subsubsection{Overall models performance comparison}

\begin{table}[h!]
\centering
\begin{tabular}{|l|c|c|c|c|}
\hline
\multicolumn{1}{|c|}{\textbf{Model}}                                                         & \textbf{HIV}   & \textbf{BACE}  & \textbf{BBBP}  & \textbf{Lipophilicity} \\ \hline
GCN                                                                                          & 73.51          & 70.58          & 68.61          & 0.774                  \\ \hline
GraphSAGE                                                                                    & 74.14          & \textbf{84.51} & 63.09          & 0.777                  \\ \hline
GIN                                                                                          & 74.07          & 68.84          & 62.58          & 0.779                  \\ \hline
GAT                                                                                          & 75.86          & 77.69          & 67.76          & 0.753                  \\ \hline
SGC                                                                                          & 75.56          & 63.24          & 66.21          & 0.884                  \\ \hline
GCN + JK max                                                                                 & 75.36          & 80.96          & 64.53          & 0.763                  \\ \hline
GraphSAGE + JK max                                                                           & 78.18          & 79.85          & 66.98          & 0.776                  \\ \hline
GIN + JK max                                                                                 & 72.00          & 65.15          & 66.38          & 0.747                  \\ \hline
GAT + JK max                                                                                 & 71.75          & 75.73          & 68.49          & 0.734                  \\ \hline
GCN + JK concat                                                                              & 78.72          & 81.03          & 69.01          & \textbf{0.715}         \\ \hline
GraphSAGE + JK concat                                                                        & 77.98          & 80.93          & 63.57          & 0.732                  \\ \hline
GIN + JK concat                                                                              & 74.60          & 65.61          & 67.89          & 0.777                  \\ \hline
GAT + JK concat                                                                              & 76.00          & 67.76          & 66.67          & 0.808                  \\ \hline
GCN + JK LSTM                                                                                & OOM            & 81.08          & 65.99          & \textbf{0.721}         \\ \hline
GraphSAGE + JK LSTM                                                                          & OOM            & 80.39          & 64.53          & \textbf{0.726}         \\ \hline
GIN + JK LSTM                                                                                & OOM            & 70.34          & 60.16          & 0.735                  \\ \hline
GAT + JK LSTM                                                                                & OOM            & 76.09          & 66.08          & 0.791                  \\ \hline
MFPs                                                                                         & 72.10          & 79.30          & 66.17          & 1.022                  \\ \hline
MFPs + atom encoding                                                                         & 73.12          & 78.78          & 68.19          & 0.992                  \\ \hline
Deep Multisets                                                                               & 72.75          & 71.81          & 62.86          & 0.848                  \\ \hline
Deep Multisets + atom encoding                                                               & 71.13          & 80.93          & 65.46          & 0.837                  \\ \hline
LDP                                                                                          & 71.57          & 80.16          & 62.77          & 1.006                  \\ \hline
LDP extended                                                                                 & 72.65          & 81.02          & 65.06          & 0.966                  \\ \hline
LDP with additional features                                                                 & 75.35          & 81.43          & 67.11          & 0.965                  \\ \hline
\begin{tabular}[c]{@{}l@{}}LDP with additional features\\ and feature selection\end{tabular} & 74.06          & 80.82          & 66.94          & 0.966                  \\ \hline
Morgan                                                                                       & 79.40          & \textbf{86.26} & 67.16          & 0.857                  \\ \hline
RDKit                                                                                        & \textbf{79.61} & 83.86          & \textbf{69.10} & 0.935                  \\ \hline
MACCS                                                                                        & \textbf{80.49} & 82.78          & \textbf{70.54} & 0.883                  \\ \hline
Fingerprints concatenation                                                                   & \textbf{82.25} & \textbf{86.17} & \textbf{70.47} & 0.820                  \\ \hline
\end{tabular}
\caption{The best models summary. We mark the 3 best models for each dataset in bold.}
\label{table_best_models_summary}
\end{table}

See Table \ref{table_best_models_summary}. On 3 out of 4 datasets, the fingerprint-based models are clear winners. For HIV and BBBP all 3 best models are fingerprint-based, and for BACE two of the best ones, with GraphSAGE achieving 2\% less AUROC. For Lipophilicity, however, the GNNs combined with JK are clearly better. Purely topological LDP methods were far from the best in all cases. We conclude that fingerprint-based methods, due to simple, yet smart and effective domain-specific feature engineering, still have to be considered for production-grade systems using graph classification.

In all cases, the baseline architectures were not in the top 3. This means that problems are nontrivial, datasets are reasonably chosen, their test sets have been chosen appropriately, and that other methods have enough data and ability to use it. Outperforming baselines has been a problem on older datasets, so this validates our experimental design.

On HIV dataset, all four fingerprint-based models achieve clearly better results than all GNN-based models, having 1.5\% higher AUROC than the best GNN model (GraphSAGE + JK max). This means that for this problem the extraction of larger graph substructures, which is problematic for GNNs, is crucial. Additionally, this is troubling for GNNs, as this is the largest dataset and the lack of training data is not a problem here. The low performance of GIN architecture, which aims to better recognize graph substructures, is a bit disappointing.

On BACE dataset, the Morgan fingerprint achieves the best result, and the second result is the concatenation of fingerprints. This fingerprint is atom-centric, describing the neighborhood information well. The best GNN model, GraphSAGE, also relies on each by itself, more than, e.g. GCN. This may mean that in this dataset, the crucial information lies in a few important nodes in each molecule.

Three best models on BBBP dataset are also fingerprint-based, so the situation looks similar to the HIV dataset. However, there the MACCS and RDKit fingerprints are best, as is their concatenation. This may mean that there are only a few crucial substructures that are enough for good results, and that they are part of MACCS features.

Lipophilicity dataset is the only one where GNNs have consistent advantage over other methods. This is the only physical chemistry dataset, so this may mean that physical property relies mostly on small radius neighborhood features, which GNNs easily extract. All three best architectures use jumping knowledge, so the information needed may be flat (not hierarchical, since fingerprints do not perform well), but not strictly local.

\subsubsection{Baselines performance}

We compare the baselines to the best GNN model, the best LDP model and the best fingerprint-based model for each dataset.

\begin{table}[h!]
\centering
\begin{tabular}{|l|c|c|c|c|}
\hline
\multicolumn{1}{|c|}{\textbf{Model}} & \textbf{HIV}   & \textbf{BACE}  & \textbf{BBBP}  & \textbf{Lipophilicity} \\ \hline
Best GNN                             & 78.18          & 84.51          & 68.61          & 0.715                  \\ \hline
Best LDP                             & 75.35          & 81.43          & 67.11          & 0.965                  \\ \hline
Best fingerprint-based               & 82.25          & 86.26          & 70.54          & 0.820                  \\ \hline
SGC                                  & \textbf{75.56} & 63.24          & \textbf{66.21} & \textbf{0.884}         \\ \hline
MFPs                                 & 72.10          & \textbf{79.30} & \textbf{66.17} & 1.022                  \\ \hline
MFPs + atom encoding                 & \textbf{73.12} & \textbf{78.78} & \textbf{68.19} & 0.992                  \\ \hline
Deep Multisets                       & \textbf{72.75} & 71.81          & 62.86          & \textbf{0.848}         \\ \hline
Deep Multisets + atom encoding       & 71.13          & \textbf{80.93} & 65.46          & \textbf{0.837}         \\ \hline
\end{tabular}
\caption{Best model in each group vs baselines comparison. We mark the 3 best baseline models for each dataset in bold. }
\label{table_baselines}
\end{table}

See Table \ref{table_baselines}. It is not clear which baseline method performs best. SGC and MFPs are the best models on 1 dataset each, and Deep Multisets is best on two. Additionally, the SGC model is either the best or close to other baselines, except for the BACE dataset, where it performs much worse. This may also confirm our hypothesis from the previous section that the few nodes are most important here, since SGC does not differentiate between nodes well. MFPs and Deep Multisets can indeed do this, and the second model comes close to the best LDP one.

Among all four datasets, only on BBBP the baseline comes reasonably close to the best regular model, being about 2\% worse than the best fingerprint-based method. However, the best baseline is able to outperform at least one group of other models on 3 datasets. Only on BACE it cannot, and it still comes close to the LDP model. This shows that using varied approaches is required for those tasks, and strong baselines are very much needed as reference points.

Our idea to apply SGC to graph classification task seems reasonable, based on the results. On HIV, it was better than previously proposed baselines, including the LDP method. On BBBP it also achieved reasonable performance. This method is also very fast, making it a good reference point, especially for GNNs.

Adding atom encoding to Molecular Fingerprints and Deep Multisets is a success. It improved results on 3 out of 4 datasets for both methods. Improvements range from about 1\% AUROC or 0.1-0.2 RMSE to as high as over 9\% in case of Deep Multisets on BACE dataset. This model is actually the best baseline on this dataset, rivaling the best LDP model. This means that with standard OGB features, the baselines created for old datasets with limited atom information do not employ richer information well by themselves. Adding the embedding layer and therefore providing lower-dimensional, continuous features for those baselines typically makes them stronger and more suitable for modern datasets.

\subsubsection{GNN models performance}

\begin{table}[h!]
\centering
\begin{tabular}{|l|c|c|c|c|}
\hline
\multicolumn{1}{|c|}{\textbf{Model}} & \textbf{HIV}   & \textbf{BACE}  & \textbf{BBBP}  & \textbf{Lipophilicity} \\ \hline
GCN & 73.51          & 70.58          & \textbf{68.61} & 0.774                  \\ \hline
GraphSAGE & 74.14          & \textbf{84.51} & 63.09          & 0.777                  \\ \hline
GIN & 74.07          & 68.84          & 62.58          & 0.779                  \\ \hline
GAT & 75.86 & 77.69          & 67.76          & 0.753                  \\ \hline
SGC & 75.56 & 63.24          & 66.21          & 0.884                  \\ \hline
GCN + JK max & 75.36          & 80.96          & 64.53          & 0.763                  \\ \hline
GraphSAGE + JK max & \textbf{78.18} & 79.85          & 66.98          & 0.776                  \\ \hline
GIN + JK max & 72.00 & 65.15          & 66.38          & 0.747                  \\ \hline
GAT + JK max & 71.75 & 75.73          & \textbf{68.49} & 0.734         \\ \hline
GCN + JK concat & \textbf{78.72} & \textbf{81.03} & \textbf{69.01} & \textbf{0.715}         \\ \hline
GraphSAGE + JK concat & \textbf{77.98} & 80.93          & 63.57          & 0.732         \\ \hline
GIN + JK concat & 74.60 & 65.61          & 67.89          & 0.777                  \\ \hline
GAT + JK concat & 76.00 & 67.76          & 66.67          & 0.808                  \\ \hline
GCN + JK LSTM & OOM & \textbf{81.08} & 65.99          & \textbf{0.721} \\ \hline
GraphSAGE + JK LSTM & OOM & 80.39          & 64.53          & \textbf{0.726} \\ \hline
GIN + JK LSTM & OOM & 70.34          & 60.16          & 0.735 \\ \hline
GAT + JK LSTM & OOM & 76.09          & 66.08          & 0.791 \\ \hline
\end{tabular}
\caption{GNN models comparison. We mark the 3 best models for each dataset in bold.}
\label{table_gnn_models}
\end{table}

Seeing GNN results in Table \ref{table_gnn_models}, we notice that only on BACE dataset the best GNN model is a simple flat GNN without Jumping Knowledge skip connections. However, on BBBP, the simple GCN model was second best. On 2 datasets the best model is GCN with JK concatenation, and on HIV the best model is GraphSAGE with JK max. No JK LSTM model was the best, but they were in top 3 on BACE and Lipophilicity. However, on BACE dataset, the difference between GCN + JK LSTM and GCN + JK concat was practically negligible. On HIV dataset, all JK LSTM models resulted in OOM errors, which shows that this approach is not scalable. The results on BACE and BBBP show that checking simple models without any Jumping Knowledge is also required for graph classification, and blindly adding this mechanism may obscure the information from the last layers, when the problem requires more knowledge about the further nodes. The results of LSTM aggregator are unsatisfying, as it is the only architecture overall that could not be computed on large dataset. Since graph data is becoming more available and datasets larger, and LSTM results are not particularly spectacular, we can reasonably suggest omitting the LSTM aggregator for graph classification. The choice between JK max and concat, however, is much less obvious. The GCN + JK concat combination, however, seems especially potent, achieving the best result on 3 datasets and third best (very closely to the second place) on the 4th dataset, which is the most consistent advantage among all GNN models.

\begin{table}[h!]
\centering
\begin{tabular}{|l|c|c|c|c|}
\hline
\multicolumn{1}{|c|}{\textbf{Model}} & \textbf{HIV}   & \textbf{BACE}  & \textbf{BBBP}  & \textbf{Lipophilicity} \\ \hline
GCN       & 73.51          & \textbf{70.58} & \textbf{68.61} & \textbf{0.774} \\ \hline
GraphSAGE & \textbf{74.14} & \textbf{84.51} & 63.09          & \textbf{0.777} \\ \hline
GIN       & 74.07          & 68.84          & 62.58          & 0.779          \\ \hline
GAT       & \textbf{75.86} & \textbf{77.69} & \textbf{67.76} & \textbf{0.753} \\ \hline
SGC       & \textbf{75.56} & 63.24          & \textbf{66.21} & 0.884          \\ \hline
\end{tabular}
\caption{GNN flat models (without JK) comparison. We mark the 3 best models for each dataset in bold.}
\label{table_gnn_models_flat_only}
\end{table}

Focusing only on architectures without JK (see Table \ref{table_gnn_models_flat_only}), we can see that GAT architecture achieves consistently good performance, being always the best or second-best model. This signifies the potential of attentional architectures for graph classification. However, it should be kept in mind that in the comparison between all GNNs GAT-based architectures were not particularly powerful. Nevertheless, the GAT architecture has been designed purely with node classification in mind \cite{attentional_gat}, so this result is very good, particularly in comparison with GIN. This may be because attention weighting allows GAT to ignore less important atoms to a large extent, instead focusing on a few most important small functional groups in a molecule.

GIN performs surprisingly bad overall. No GIN-based model is in the top 3 on any dataset, both overall and among flat models only. Since this architecture is the only one among those which were designed for graph classification specifically, it may suggest that complex neighbor aggregation scheme through MLP is not that required in practice for molecular classification. Not even the strong theoretical guarantees and being equal to the WL-test in terms of power to distinguish graph structures \cite{gnn_GIN} are helpful here. In fact, on HIV and BBBP datasets, the baseline SGC method performs better. This shows that independent benchmarking on several datasets and introducing strong baselines is very important even for theoretically sound models.

The simple GCN and GraphSAGE models perform very well. Despite simple neighborhood aggregation schemes, they are consistently good both in flat architectures and in conjunction with Jumping Knowledge. The basic GraphSAGE model is the best one on the BACE dataset among all compared models. Its good performance can be attributed to focusing separately on the node and its neighborhood during aggregation. This shows that checking both GCN and GraphSAGE as baselines for novel GNN models is important.

\begin{table}[h!]
\centering
\begin{tabular}{|l|c|c|c|c|}
\hline
\multicolumn{1}{|c|}{\textbf{Model}}                                                         & \textbf{HIV}   & \textbf{BACE}  & \textbf{BBBP}  & \textbf{Lipophilicity} \\ \hline
GCN + JK max          & 75.36          & \textbf{80.96} & 64.53          & 0.763                  \\ \hline
GraphSAGE + JK max    & \textbf{78.18} & 79.85          & 66.98          & 0.776                  \\ \hline
GIN + JK max          & 72.00 & 65.15          & 66.38          & 0.747                  \\ \hline
GAT + JK max          & 71.75 & 75.73          & \textbf{68.49} & \textbf{0.734}         \\ \Xhline{1.5pt}
GCN + JK concat       & \textbf{78.72} & \textbf{81.03} & \textbf{69.01} & \textbf{0.715}         \\ \hline
GraphSAGE + JK concat & 77.98 & 80.93          & 63.57          & 0.732                  \\ \hline
GIN + JK concat       & 74.60 & 65.61          & 67.89          & 0.777                  \\ \hline
GAT + JK concat       & 76.00 & 67.76          & 66.67          & 0.808                  \\ \Xhline{1.5pt}
GCN + JK LSTM         & OOM & \textbf{81.08} & 65.99          & \textbf{0.721} \\ \hline
GraphSAGE + JK LSTM   & OOM & 80.39          & 64.53          & 0.726 \\ \hline
GIN + JK LSTM         & OOM & 70.34          & 60.16          & 0.735 \\ \hline
GAT + JK LSTM         & OOM & 76.09          & \textbf{66.08} & 0.791 \\ \hline
\end{tabular}
\caption{GNN JK models comparison. We mark the best model for each JK approach for each dataset in bold.}
\label{table_gnn_models_jk_only}
\end{table}

Lastly, we analyze the JK approaches (see Table \ref{table_gnn_models_jk_only}). GCN with JK concat is the most potent model, heavily outperforming other concatenation models on all datasets, except for BACE. Among JK max models, there is no clear winner. LSTM aggregator performs consistently similar to alternatives on BACE and BBBP, while getting OOM errors on HIV. We conclude that considering Jumping Knowledge with either max or concatenation aggregation is a powerful tool for graph classification, having very low computational and memory cost, while often improving results by a large margin.

\subsubsection{LDP models performance}

\begin{table}[h!]
\centering
\begin{tabular}{|l|c|c|c|c|}
\hline
\multicolumn{1}{|c|}{\textbf{Model}} & \textbf{HIV} & \textbf{BACE} & \textbf{BBBP} & \textbf{Lipophilicity} \\ \hline
LDP                          & 71.57                   & 80.16          & 62.77          & 1.006                  \\ \hline
LDP extended                 & 72.65                   & 81.02          & 65.06          & 0.966                  \\ \hline
LDP with additional features & \textbf{75.35}          & \textbf{81.43}          & \textbf{67.11}          & \textbf{0.965}                  \\ \hline
\begin{tabular}[c]{@{}l@{}}LDP with additional features\\ and feature selection\end{tabular} & 74.06          & 80.82          & 66.94          & 0.966                  \\ \hline
\end{tabular}
\caption{LDP models comparison. We mark the best model for each dataset in bold }
\label{table_ldp_results}
\end{table}

Among LDP models, we have a clear winner (see Table \ref{table_ldp_results}), the LDP with additional features. It achieves the best performance across all datasets. The results on Lipophilicity are tied among all models except for the basic LDP, so for physical chemistry the shortest paths information and additional graph features may not be particularly helpful. On HIV and BBBP, the gains from additional features are significant, about 3\% and 2\%, respectively. This validates our approach and shows that adding well selected graph descriptors can capture information important from a perspective of a whole molecule, when used in an inherently local method.

The basic LDP method, as proposed in the original paper \cite{gnn_local_degree_profile}, is the worst among LDP models. Adding the shortest paths, which was researched in the context of chemical graph classification, also helps significantly for MoleculeNet datasets. On Lipophilicity all methods using more than basic features achieve basically a tie, so this information can sometimes capture as much information as other, additional descriptors.

Feature selection did not help, and always achieved worse performance than using all additional features. This is a very good thing, for two reasons. First, it means that we have chosen the features well and their information does not overlap. Secondly, the feature selection is costly, and this means that we can omit that step and just add all 4 features.

Based on those observations, we can confidently say that we have improved the LDP method. Our baseline is stronger in all cases, does not lose generality (additional features are general and apply to graphs in any domain), and is highly scalable, since features we have chosen are well known and easily computable even for very large graphs.

\subsubsection{Fingerprint models performance}

\begin{table}[]
\centering
\begin{tabular}{|l|c|c|c|c|}
\hline
\multicolumn{1}{|c|}{\textbf{Model}} & \textbf{HIV} & \textbf{BACE} & \textbf{BBBP} & \textbf{Lipophilicity} \\ \hline
Morgan                     & 79.40          & \textbf{86.26} & 67.16          & 0.857         \\ \hline
RDKit                      & 79.61          & 83.86          & 69.10          & 0.935                  \\ \hline
MACCS                      & 80.49 & 82.78          & \textbf{70.54} & 0.883                  \\ \hline
Fingerprints concatenation & \textbf{82.25} & 86.17 & 70.47 & \textbf{0.820}         \\ \hline
\end{tabular}
\caption{Fingerprint models comparison. We mark the best model for each dataset in bold.}
\label{table_fingerprints_results}
\end{table}

The fingerprint-based models achieve very good performance overall among all models. The performance of individual fingerprints, however, strongly vary between datasets (see Table \ref{table_fingerprints_results}). Morgan fingerprint, the most commonly used fingerprint among three ones used, and the most commonly used as a baseline against GNNs, is the best only on one dataset, and the second best one (concatenated fingerprint) has very similar score. MACCS fingerprint is the best on BBBP and the best among individual fingerprints on HIV. RDKit fingerprint has the worst performance of the three, but outperforms MACCS on BACE and Morgan on BBBP.

All fingerprints perform badly on Lipophilicity dataset. No single fingerprint is able to outperform the proposed strong baseline, Deep Multisets with atom encoding. Only concatenated fingerprint can achieve that, but it also falls short in comparison with GNNs. This may indicate that this approach performs well for predicting higher level properties of molecules, such as biophysical or physiological properties, but for low level tasks of physical chemistry other approaches are more suitable. Chemical indices, for example, have been originally created for that application, and they are both simpler and extract more global information than fingerprints.

Our proposed fingerprints concatenation is the most robust fingerprint. It has the best performance on 2 datasets and is a very close second on the other two. Since the best individual fingerprint varies heavily, combining them gives a better result in general. We also combine this feature extraction with Random Forest, which can measure feature importance and perform feature selection, effectively combining features from multiple fingerprints. We can recommend using this approach as a general purpose tool and as a strong method for molecular classification above physical chemistry level, able to outperform or at least rival GNNs.

\subsubsection{Learning curves analysis}

For neural network models, we can also perform analysis of their learning curves. This provides additional insights into their stability, tendency to overfit and ease of training.

Our main observation is that training GNNs, even on large datasets, is often extremely unstable, compared to CNNs (based on our experience). This also applies to baseline neural network models, Molecular Fingerprint and Deep Multisets. While learning rates used were low, going as low as $1e-5$, we experienced chaotic changes during training and validation, or even complete divergence. Interestingly, this behavior has been visible only on classification datasets, while on the Lipophilicity datasets stability was much better.

Because we performed hyperparameter optimization, we got 40 learning curves per model. As this number is far too great for a practical analysis, we chose a selection of plots exhibiting behaviors mentioned above and present them below. Note that all learning curves presented below plot metric value against epoch number, i.e. they are not loss curves. For classification (AUROC metric) the higher, the better, and for Lipophilicity dataset (RMSE metric) the lower, the better. 

\begin{figure}[h!]
\makebox[\textwidth][c]{\includegraphics[width=0.8\textwidth]{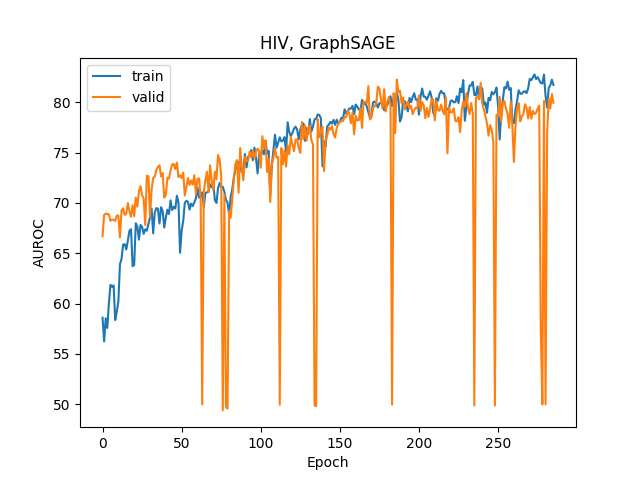}}
\caption{Unstable behavior of GraphSAGE on HIV.}
\label{image_hiv_graphsage_chaotic}
\end{figure}

\begin{figure}[h!]
\makebox[\textwidth][c]{\includegraphics[width=0.8\textwidth]{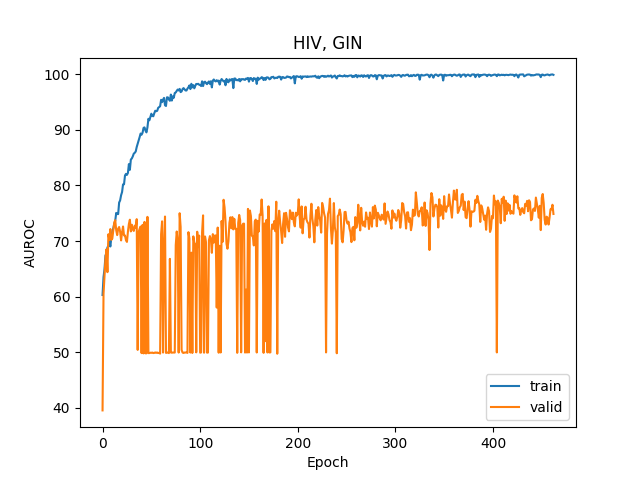}}
\caption{Unstable behavior of GIN on HIV.}
\label{image_hiv_gin_chaotic}
\end{figure}

\begin{figure}[h!]
\makebox[\textwidth][c]{\includegraphics[width=0.8\textwidth]{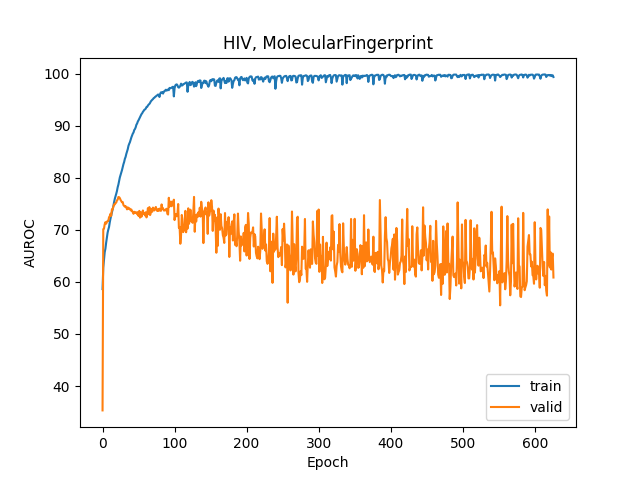}}
\caption{Unstable behavior of Molecular Fingerprint on HIV.}
\label{image_hiv_mfps_chaotic}
\end{figure}

\begin{figure}[h!]
\makebox[\textwidth][c]{\includegraphics[width=0.8\textwidth]{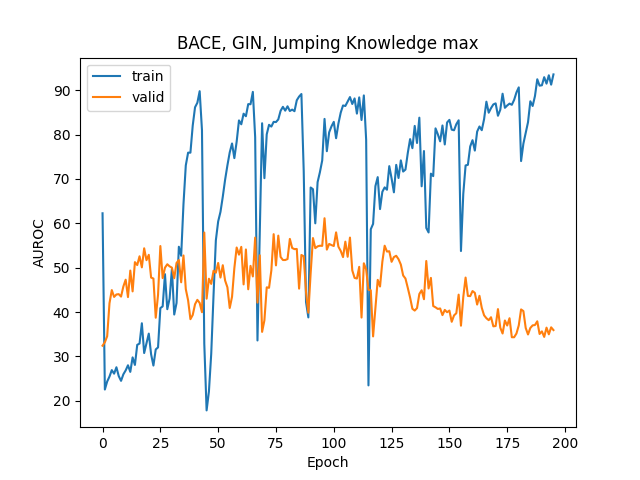}}
\caption{Unstable behavior of GIN + JK max on BACE.}
\label{image_bace_gin_chaotic}
\end{figure}

\begin{figure}[h!]
\makebox[\textwidth][c]{\includegraphics[width=0.8\textwidth]{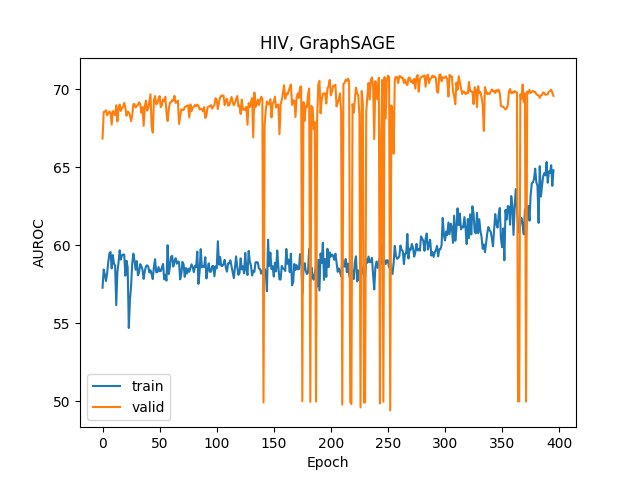}}
\caption{Divergent training of GraphSAGE on HIV.}
\label{image_hiv_graphsage_divergent}
\end{figure}

\begin{figure}[h!]
\makebox[\textwidth][c]{\includegraphics[width=0.8\textwidth]{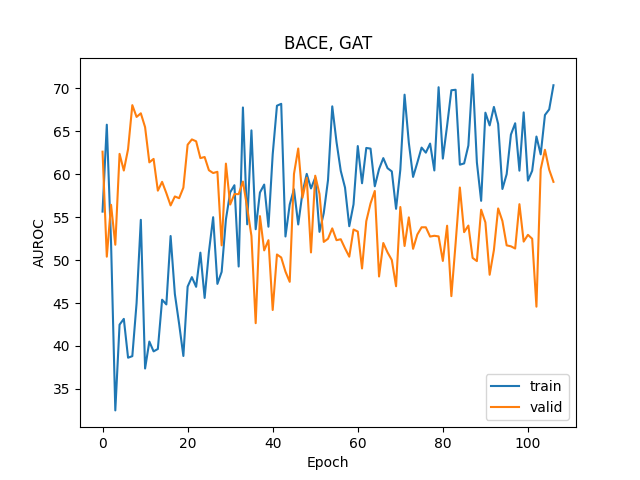}}
\caption{Divergent training of GAT on BACE.}
\label{image_bace_gat_divergent}
\end{figure}

\begin{figure}[h!]
\makebox[\textwidth][c]{\includegraphics[width=0.8\textwidth]{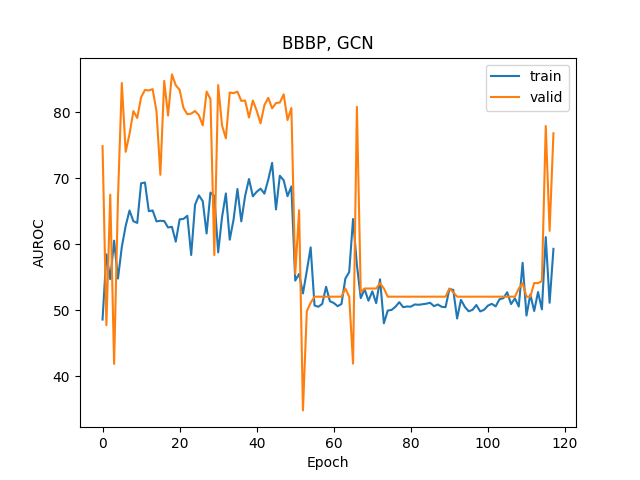}}
\caption{Divergent training of GCN on BBBP.}
\label{image_bbbp_gcn_divergent}
\end{figure}

The figures \ref{image_hiv_graphsage_chaotic} - \ref{image_bace_gin_chaotic} present unstable behavior of GNNs and baseline NN-based models. Figures \ref{image_hiv_graphsage_divergent} - \ref{image_bbbp_gcn_divergent} present examples of divergent training.

\newpage

\clearpage

\begin{figure}[h!]
\makebox[\textwidth][c]{\includegraphics[width=0.85\textwidth]{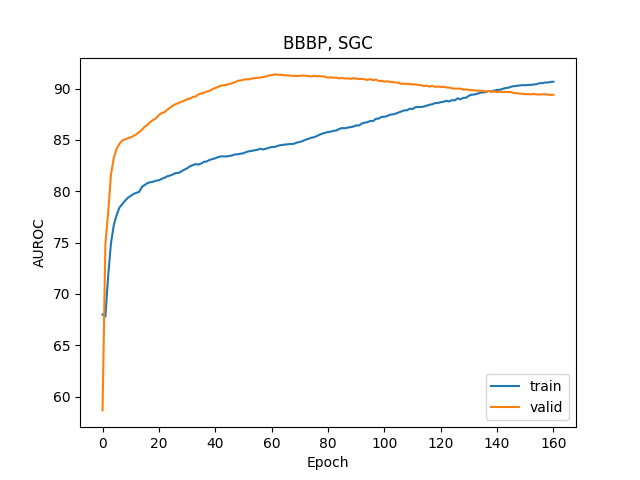}}
\caption{Validation curve above training curve, training SGC on BBBP.}
\label{image_bbbp_sgc_valid_over_train}
\end{figure}

\begin{figure}[h!]
\makebox[\textwidth][c]{\includegraphics[width=0.85\textwidth]{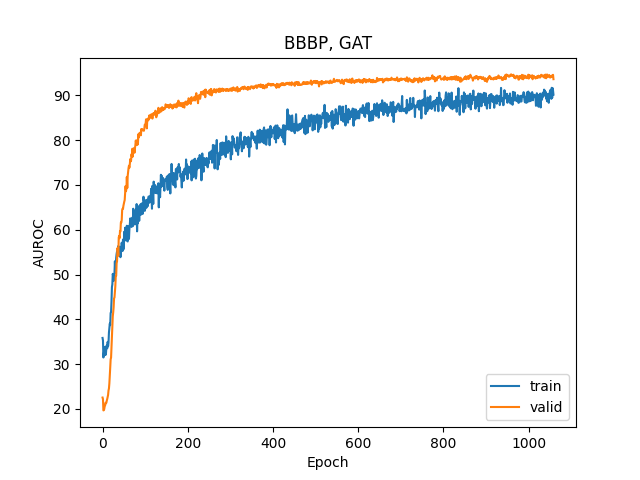}}
\caption{Validation curve above training curve, training GAT on BBBP.}
\label{image_bbbp_gat_valid_over_train}
\end{figure}

Another surprising behavior which we experienced is that the validation curve consistently better than the training curve, i.e. validation scores are better than training scores. This can be explained by the fact that we are dealing with out-of-distribution problems, and initial random weights and further training favor the predictions of the validation dataset. See Figure \ref{image_bbbp_sgc_valid_over_train} and Figure \ref{image_bbbp_gat_valid_over_train}.

\clearpage

\begin{figure}[h!]
\makebox[\textwidth][c]{\includegraphics[width=0.8\textwidth]{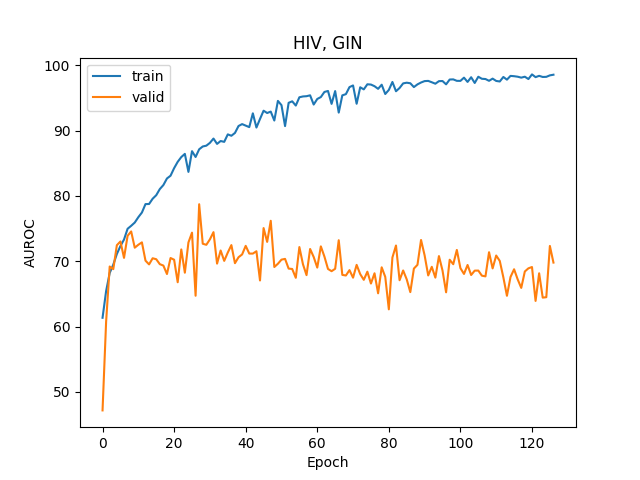}}
\caption{Overfitting, training GIN on HIV.}
\label{image_hiv_gin_overfitting}
\end{figure}

\begin{figure}[h!]
\makebox[\textwidth][c]{\includegraphics[width=0.8\textwidth]{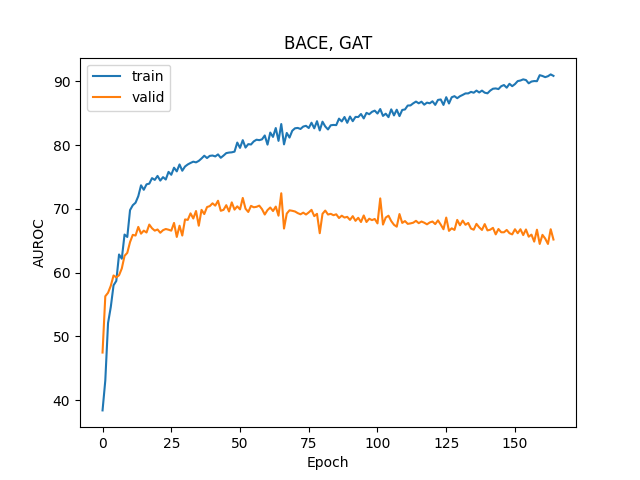}}
\caption{Overfitting, training GAT on BACE.}
\label{image_bace_gat_overfitting}
\end{figure}

\begin{figure}[h!]
\makebox[\textwidth][c]{\includegraphics[width=0.8\textwidth]{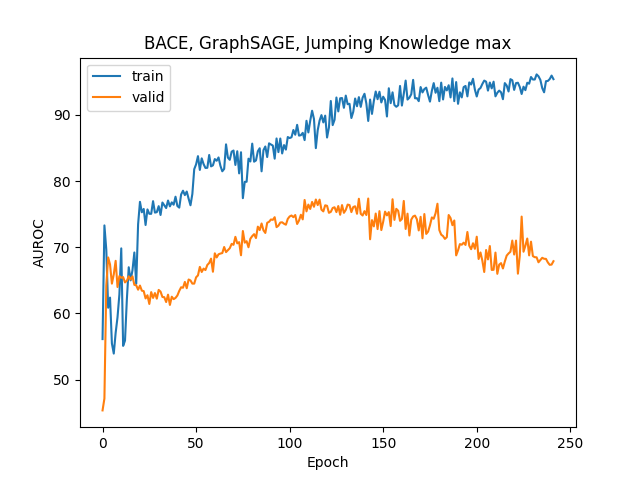}}
\caption{Overfitting, training GraphSAGE + JK max on BACE.}
\label{image_bace_graphsage_overfitting}
\end{figure}

Overfitting, often heavy, was visible during many trainings. While for BACE, BBBP or Lipophilicity this could be explained by relatively small dataset sizes, we also saw that behavior on large HIV dataset. Initial experiments with reducing learning rate did not achieve anything, since the learning rates are small to begin with. GNNs may overfit on local substructures, majority classes (in case of imbalanced datasets) and similar molecules, which will probably also present a challenge in the future. See Figures \ref{image_hiv_gin_overfitting} - \ref{image_bace_graphsage_overfitting}.

\clearpage

\begin{figure}[h!]
\makebox[\textwidth][c]{\includegraphics[width=0.8\textwidth]{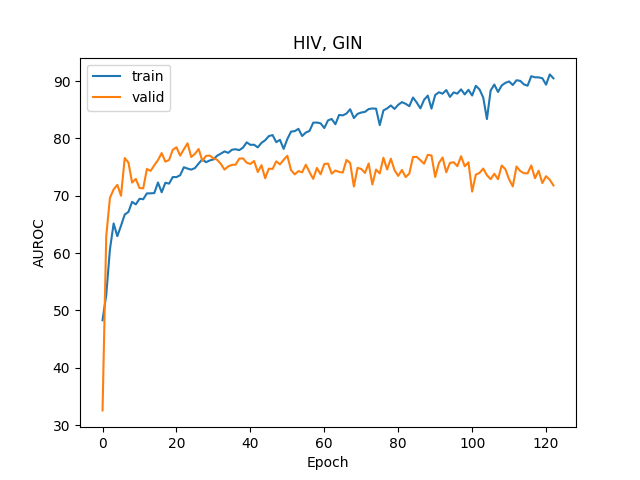}}
\caption{Well-behaved learning curves, training GIN on Lipophilicity.}
\label{image_hiv_gin_nice}
\end{figure}

While overfitting was present to some extent across almost all trainings, some GNNs trained well, especially on Lipophilicity dataset. The parameters for those cases, however, seemed quite random. For examples, see Figures \ref{image_hiv_graphsage_nice} - \ref{image_hiv_gin_nice}.

\begin{figure}[h!]
\makebox[\textwidth][c]{\includegraphics[width=0.8\textwidth]{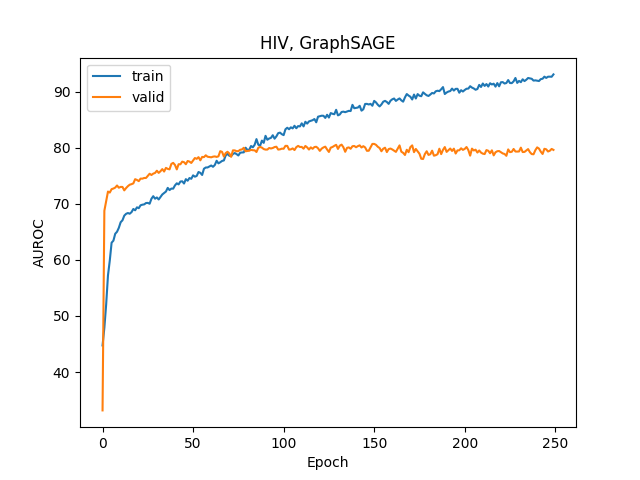}}
\caption{Well-behaved learning curves, training GraphSAGE on HIV.}
\label{image_hiv_graphsage_nice}
\end{figure}

\begin{figure}[h!]
\makebox[\textwidth][c]{\includegraphics[width=0.8\textwidth]{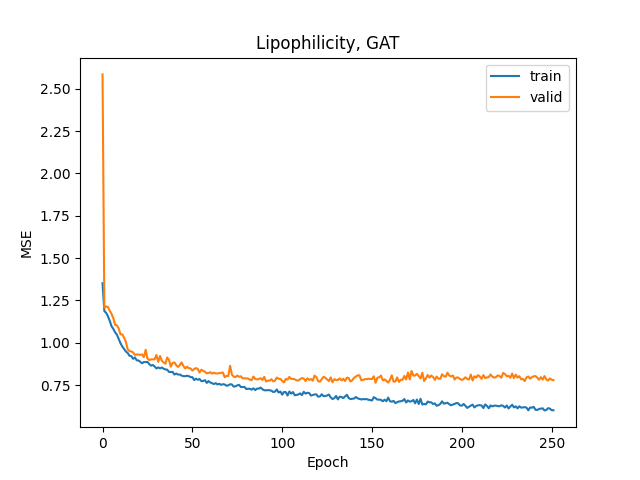}}
\caption{Well-behaved learning curves, training GAT on Lipophilicity.}
\label{image_hiv_graphsage_nice}
\end{figure}

\subsubsection{Hyperparameters analysis}

In this section we analyze hyperparameters chosen by Optuna for GNNs, LDP and fingerprint-based models. This is important, since it is a basis for selecting hyperparameter ranges on similar datasets.

\begin{figure}[h!]
\makebox[\textwidth][c]{\includegraphics[width=0.8\textwidth]{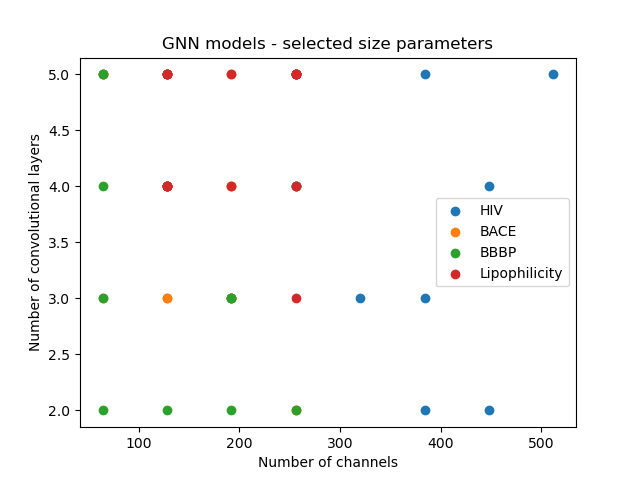}}
\caption{Visualization of GNN model sizes. Note that some dots on this scatter plot can overlap each other.}
\label{image_gnn_model_sizes}
\end{figure}

First, we focus on GNN model size, measured by number of convolutional layers and number of channels in each layer, as shown in Figure \ref{image_gnn_model_sizes}. On BBBP dataset, small networks are preferred. Lipophilicity dataset, where GNNs achieve the best results, often uses deep architectures (4-5 layers, where 5 was the maximal depth), and the number of channels varies almost uniformly between 128 and 256. We note that multiple networks hit the upper limit of 256 channels, which was imposed in order to avoid catastrophic overfitting. Nevertheless, we still experienced heavy overfitting. Therefore, it may be reasonable to reduce the maximal number of channels even more. For HIV dataset, where we allowed maximum of 512 channels, because of the large dataset size, the large number of channels was indeed often selected, but with greatly varied number of layers, even as low as 2.

\begin{figure}[h!]
\makebox[\textwidth][c]{\includegraphics[width=0.8\textwidth]{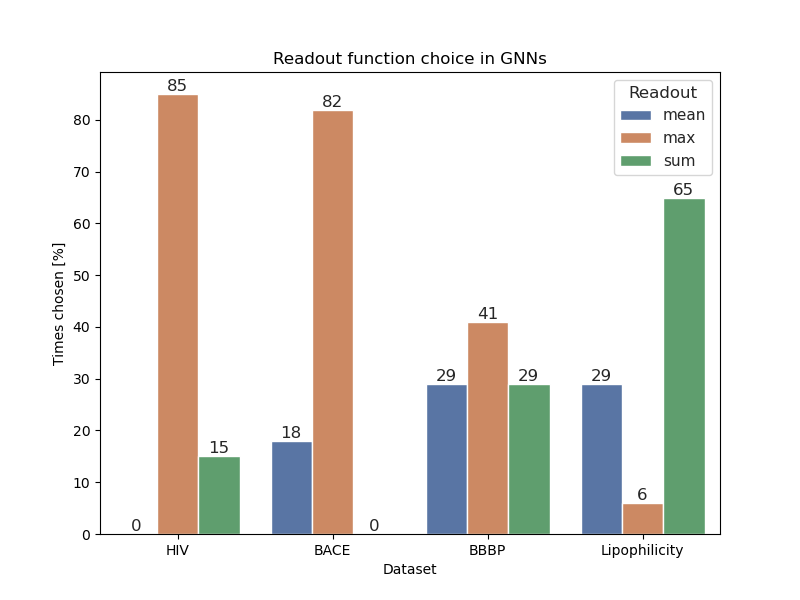}}
\caption{Visualization of GNN readout functions.}
\label{image_gnn_readout_choices}
\end{figure}

As shown in Figure \ref{image_gnn_readout_choices}, readout function choice varies by the dataset. Max clearly dominates on HIV and BACE datasets. This may mean that only the presence (or lack thereof) of a particular feature in a molecule is the most important for this task. This may indicate that, e.g. some small functional groups, which GNN can extract as a part of message passing between neighbors, are very important for those tasks. On BBBP, the distribution is almost uniform. On Lipophilicity, the max function is, in turn, almost unused, but mean and sum functions are most often used. Those results happen despite sum function being provably more powerful in distinguishing graph structures \cite{gnn_GIN}, but this indicates that this property is not that useful on those tasks.

\begin{figure}[h!]
\makebox[\textwidth][c]{\includegraphics[width=0.8\textwidth]{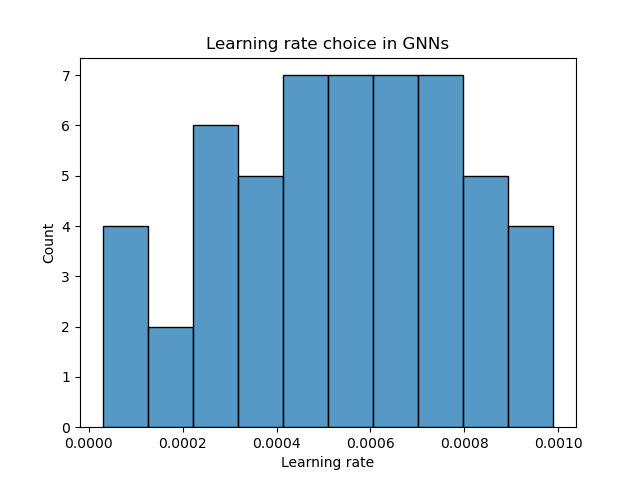}}
\caption{Visualization of GNN learning rates.}
\label{image_gnn_learning_rates}
\end{figure}

The histogram of learning rates in Figure \ref{image_gnn_learning_rates} has been prepared jointly for all datasets, as values are similar across all of them. Distribution is between normal and uniform, with the majority of values lying in the middle. This shows that our choice of range for this hyperparameter was reasonable, since if small or large values would dominate, that would indicate that range was too short and should extend towards smaller / larger learning rates.

Overall, we can conclude that our hyperparameter ranges across all datasets made sense, and various architectures made use of different fragments of the available parameter space. For GNNs, since each experiment was allowed 40 trials, they could all search through reasonably many hyperparameter combinations, while being steered towards better ones by Bayesian optimization in Optuna. Variability on plots for GNNs shows that tuning their hyperparameters is a nontrivial task, and choice of advanced Bayesian methods is a good one.

\clearpage

\section{Thesis summary}
\label{section_summary}

In this work, we addressed the problem of supervised learning on graphs. We compared multiple modern techniques, including Graph Neural Networks (GNNs), molecular fingerprints and graph descriptor-based methods. Here we summarize the results and future directions.

\subsection{Key findings and conclusions}

In this work, we designed a fair evaluation experimental setup for GNNs and performed extensive experiments on graph classification and regression. The experiments were followed by thorough analysis, showing strengths and weaknesses of both particular architectures and graph networks in general. We show problems sometimes rarely talked about in GNN literature - not only overfitting, but also training instability or divergence. Jumping Knowledge model is shown to provide great improvements for graph classification, and LSTM aggregation is shown to be unreliable and poorly scalable. This will allow future works to focus on reduced model space, considering max and concatenation Jumping Knowledge variants.

We also used fingerprints, which turned out to give consistently better results than GNNs for molecular property prediction. The proposed concatenation of multiple fingerprints, aimed to increase model robustness and eliminate the need of fingerprint selection, is proven to work well and generalize to multiple datasets, while still having computational cost much lower than GNNs. This highlights the problems with the novel GNN approach, and leads to many questions and future work. The main advantage of molecular fingerprints is extracting the hierarchical structures of graphs, which is problematic for inherently flat GNNs.

Local Degree Profile (LDP) has also been used as a model based purely on graph topology, and our proposed improvements to this model resulted in much better results. This resulting model can be used as a stronger baseline for arbitrary graph classification. We also checked other baselines, such as Simple Graph Convolution (SGC), which were not used for graph classification before, and gives good results as a baseline. Molecular Fingerprints (MFPs) and Deep Multisets were also used, and we proposed enhancing the two latter architectures with a node embedding layer, which improves results and results in stronger baseline. Work on simple, yet strong baselines is a basis of fair evaluation, and as such those advancements should greatly benefit future graph classification benchmarks, especially on new or small/medium-sized datasets.

\subsection{Future work}

Multiple datasets show that extracting hierarchical information through molecular fingerprints is crucial for good performance. While Jumping Knowledge allows this to a certain extent, other improvements to GNNs can be considered in this regard. Pooling layers \cite{pooling_pooling_strategies}, such as DiffPool \cite{pooling_diffpool}, gPool \cite{pooling_gpool} or SAGPool \cite{pooling_sagpool} emerged recently as graph analogues of pooling layers in CNNs. They face many problems, however, as creating subgraphs hierarchy is a particularly hard task, since it has to consider graph structure, node attributes, and also possibly edge attributes. Other issues such as computational cost \cite{pooling_diffpool}, or subpar performance in some cases \cite{pooling_gpool} also have to be solved. We expect much development in this area, and it is a crucial area for future work on molecular graph property prediction.

Alternatively, instead of choosing one method, multiple ones can be combined in various ways. For example, graph embeddings from the readout layer can be combined with molecular descriptors or fingerprints \cite{swanson_thesis}, incorporating hierarchical or domain-specific knowledge into GNN framework. Selection of such additional descriptors, and particularly doing so automatically, remains a challenge and promising area of development.

Further development of strong baselines remains an area of active research. As more sophisticated models are proposed, we are in even stronger need of simple, yet well-performing baseline algorithms. One way of obtaining such results would be to create ensemble algorithms from baselines proposed in this thesis, which could improve robustness.

\clearpage

\bibliographystyle{abbrv}
\bibliography{bibliography}

\appendix

\newpage

\section{Full results tables}
\label{appendix_full_results_tables}

\subsection{HIV dataset}

\begin{table}[h!]
\centering
\begin{tabularx}{\textwidth}{|X|c|c|c|c|}
\hline
\multicolumn{1}{|c|}{\textbf{Model}}                                               & \multicolumn{1}{c|}{\textbf{\begin{tabular}[c]{@{}c@{}}AUROC\\ validation\end{tabular}}} & \multicolumn{1}{c|}{\textbf{\begin{tabular}[c]{@{}c@{}}AUROC \\ test\end{tabular}}} & \textbf{\begin{tabular}[c]{@{}c@{}}MLP \\ layer size\end{tabular}} & \textbf{\begin{tabular}[c]{@{}c@{}}Learning\\ rate\end{tabular}} \\ \hline
Molecular Fingerprints & 76.59 & 72.10 & 192 & 6.3e-04 \\ \hline
Molecular Fingerprints + atom embedding & 77.47 & 73.12 & 384 & 4.2e-04 \\ \hline
Deep Multisets & 81.19 & 72.75 & 512 & 3.9e-04 \\ \hline
Deep Multisets + atom embedding & 81.19 & 71.13 & 64 & 9.6e-04 \\ \hline
\end{tabularx}
\caption{Baseline models results, HIV dataset. Metric is AUROC - the higher, the better.}
\end{table}

\begin{landscape}
\begin{table}
\centering
\begin{xltabular}{\textwidth}{|l|c|c|c|c|c|c|c|c|}
\hline
\multicolumn{1}{|c|}{\textbf{Model}}                                                        & \textbf{\begin{tabular}[c]{@{}c@{}}AUROC\\ validation\end{tabular}} & \textbf{\begin{tabular}[c]{@{}c@{}}AUROC \\ test\end{tabular}} & \textbf{Number of channels} & \textbf{Number of conv. layers} & \textbf{Readout} & \textbf{Dropout} & \textbf{\begin{tabular}[c]{@{}c@{}}Learning\\ rate\end{tabular}} & \textbf{Heads} \\ \hline
GCN                                                                                         & 83.73                                                               & 73.51                                                          & 512                  & 5                                                                  & max              & 0.0              & 3.2e-04                                                          & -              \\ \hline
GraphSAGE                                                                                   & 83.73                                                               & 74.14                                                          & 256                  & 4                                                                  & max              & 0.0              & 4.2e-04                                                          & -              \\ \hline
GIN                                                                                         & 83.56                                                               & 74.07                                                          & 448                  & 2                                                                  & max              & 0.3              & 3.4e-05                                                          & -              \\ \hline
GAT                                                                                         & 85.31                                                               & 75.86                                                          & 320                  & 3                                                                  & sum              & 0.1              & 5.1e-04                                                          & 8              \\ \hline
SGC                                                                                         & 79.98                                                               & 75.56                                                          & 384                  & 2                                                                  & max              & 0.0              & 3.2e-04                                                          & -              \\ \hline
GCN + JK max                                                                                & 85.42                                                               & 75.36                                                          & 384                  & 5                                                                  & max              & 0.0              & 7.1e-04                                                          & -              \\ \hline
GraphSAGE + JK max                                                                          & 83.85                                                               & 78.18                                                          & 128                  & 4                                                                  & max              & 0.2              & 3.6e-04                                                          & -              \\ \hline
GIN + JK max & 84.32 & 72.00 & 512 & 5 & max & 0.0 & 2.5e-04 & - \\ \hline
GAT + JK max & 87.44 & 71.75 & 512 & 5 & max & 0.1 & 3.1e-04 & 4 \\ \hline
GCN + JK concat & 84.09 & 78.72 & 128 & 2 & max & 0.2 & 8.9e-04 & - \\ \hline
GraphSAGE + JK concat & 86.03 & 77.98 & 192 & 4 & max & 0.3 & 1.8e-04 & - \\ \hline
GIN + JK concat & 82.63 & 74.60 & 384 & 3 & sum & 0.1 & 6.8e-04 & - \\ \hline
GAT + JK concat & 85.69 & 76.00 & 448 & 4 & max & 0.1 & 5.3e-04 & 6 \\ \hline
GCN + JK LSTM & OOM & - & - & - & - & - & - & - \\ \hline
GraphSAGE + JK LSTM & OOM & - & - & - & - & - & - & - \\ \hline
GIN + JK LSTM & OOM & - & - & - & - & - & - & - \\ \hline
GAT + JK LSTM & OOM & - & - & - & - & - & - & - \\ \hline
\end{xltabular}
\caption{GNN models results, HIV dataset. Metric is AUROC - the higher, the better. Models that resulted in Out-Of-Memory (OOM) errors have been marked with OOM in the first column.}
\end{table}

\begin{table}[]
\centering
\begin{tabular}{|l|c|c|c|c|c|c|c|c|}
\hline
\multicolumn{1}{|c|}{\textbf{Model}}                                                         & \textbf{\begin{tabular}[c]{@{}c@{}}AUROC\\ validation\end{tabular}} & \textbf{\begin{tabular}[c]{@{}c@{}}AUROC \\ test\end{tabular}} & \textbf{\begin{tabular}[c]{@{}c@{}}Number\\ of bins\end{tabular}} & \textbf{Normalize} & \textbf{\begin{tabular}[c]{@{}c@{}}Negative class \\ sample size\end{tabular}} & \textbf{Class weight}                                        & \textbf{\begin{tabular}[c]{@{}c@{}}Minimal samples\\ for split\end{tabular}} & \textbf{\begin{tabular}[c]{@{}c@{}}Selected\\ features\end{tabular}}                                                                               \\ \hline
LDP                                                                                          & 99.75                                                               & 71.57                                                          & 50                                                                & False              & 0.4                                                                            & Balanced subsample & 2                                                                            & -                                                                                                                                                  \\ \hline
LDP extended                                                                                 & 99.79                                                               & 72.65                                                          & 50                                                                & False              & 0.3                                                                            & Balanced                                                     & 2                                                                            & -                                                                                                                                                  \\ \hline
LDP with additional features                                                                 & 99.90                                                               & 75.35                                                          & 30                                                                & True               & 0.5                                                                            & None                                                         & 2                                                                            & -                                                                                                                                                  \\ \hline
\begin{tabular}[c]{@{}l@{}}LDP with additional features\\ and feature selection\end{tabular} & 99.84                                                               & 74.06                                                          & 20                                                                & False              & 0.5                                                                            & None                                                         & 2                                                                            & \begin{tabular}[c]{@{}c@{}}Clustering coefficient: false\\ Betweenness centrality: true\\ Closeness centrality: true\\ PageRank: true\end{tabular} \\ \hline
\end{tabular}
\caption{LDP models results, HIV dataset. Metric is AUROC - the higher, the better.}
\end{table}

\begin{table}[]
\centering
\begin{tabular}{|l|l|l|c|c|c|c|}
\hline
\multicolumn{1}{|c|}{\textbf{Model}} & \multicolumn{1}{c|}{\textbf{\begin{tabular}[c]{@{}c@{}}AUROC\\ validation\end{tabular}}} & \multicolumn{1}{c|}{\textbf{\begin{tabular}[c]{@{}c@{}}AUROC \\ test\end{tabular}}} & \textbf{\begin{tabular}[c]{@{}c@{}}Fingerprint\\ hyperparameters\end{tabular}}                                       & \textbf{\begin{tabular}[c]{@{}c@{}}Negative class\\ sample size\end{tabular}} & \textbf{Class weight} & \textbf{\begin{tabular}[c]{@{}c@{}}Minimal samples \\ for split\end{tabular}} \\ \hline
Morgan                               & 84.17                                                                                    & 79.40                                                                               & Radius: 2, bits per radius: 2048                                                                                     & 0.4                                                                           & Balanced              & 8                                                                             \\ \hline
RDKit                                & 78.46                                                                                    & 79.61                                                                               & Max path length: 6, number of bits: 4096                                                                             & 0.3                                                                           & Balanced subsample    & 2                                                                             \\ \hline
MACCS                                & 83.05                                                                                    & 80.49                                                                               & -                                                                                                                    & 0.5                                                                           & Balanced              & 8                                                                             \\ \hline
Fingerprints concatenation                        & 81.33                                                                                    & 82.25                                                                               & \begin{tabular}[c]{@{}c@{}}Radius: 3, bits per radius: 2048,\\ max path length: 5, number of bits: 4096\end{tabular} & 0.5                                                                           & None                  & 8                                                                             \\ \hline
\end{tabular}
\caption{Fingerprints models results, HIV dataset. Metric is AUROC - the higher, the better.}
\end{table}

\end{landscape}

\newpage

\subsection{BACE dataset}

\begin{table}[h!]
\centering
\begin{tabularx}{\textwidth}{|X|c|c|c|c|}
\hline
\multicolumn{1}{|c|}{\textbf{Model}}                                               & \multicolumn{1}{c|}{\textbf{\begin{tabular}[c]{@{}c@{}}AUROC\\ validation\end{tabular}}} & \multicolumn{1}{c|}{\textbf{\begin{tabular}[c]{@{}c@{}}AUROC \\ test\end{tabular}}} & \textbf{\begin{tabular}[c]{@{}c@{}}MLP \\ layer size\end{tabular}} & \textbf{\begin{tabular}[c]{@{}c@{}}Learning\\ rate\end{tabular}} \\ \hline
Molecular Fingerprints                                                             & 75.51                                                                                    & 79.30                                                                               & 192                                                                & 7.2e-04                                                          \\ \hline
Molecular Fingerprints + atom embedding & 75.09                                                                                    & 78.78                                                                               & 256                                                                & 4.3e-04                                                          \\ \hline
Deep Multisets                                                                     & 75.09                                                                                    & 71.81                                                                               & 128                                                                & 7.1e-04                                                          \\ \hline
Deep Multisets + atom embedding          & 76.33                                                                                    & 80.93                                                                               & 64                                                                 & 4.9e-04                                                          \\ \hline
\end{tabularx}
\caption{Baseline models results, BACE dataset. Metric is AUROC - the higher, the better.}
\end{table}

\begin{landscape}
\begin{table}[]
\centering
\begin{tabular}{|l|c|c|c|c|c|c|c|c|}
\hline
\multicolumn{1}{|c|}{\textbf{Model}} & \textbf{\begin{tabular}[c]{@{}c@{}}AUROC\\ validation\end{tabular}} & \textbf{\begin{tabular}[c]{@{}c@{}}AUROC \\ test\end{tabular}} & \textbf{Number of channels} & \textbf{Number of conv. layers} & \textbf{Readout} & \textbf{Dropout} & \textbf{\begin{tabular}[c]{@{}c@{}}Learning\\ rate\end{tabular}} & \textbf{Heads} \\ \hline
GCN                                  & 78.29                                                                                    & 70.58                                                                               & 192                         & 5                               & max              & 0.2              & 9.7e-04                                                          & -              \\ \hline
GraphSAGE                            & 78.57                                                                                    & 84.51                                                                               & 256                         & 2                               & max              & 0.0              & 3.9e-04                                                          & -              \\ \hline
GIN                                  & 77.50                                                                                    & 68.84                                                                               & 256                         & 2                               & max              & 0.3              & 2.3e-04                                                          & -              \\ \hline
GAT                                  & 82.23                                                                                    & 77.69                                                                               & 128                         & 3                               & max              & 0.2              & 4.8e-04                                                          & 8              \\ \hline
SGC                                  & 75.71                                                                                    & 63.24                                                                               & 256                         & 5                               & max              & 0.2              & 2.3e-04                                                          & -              \\ \hline
GCN + JK max                         & 75.12                                                                                    & 80.96                                                                               & 64                          & 3                               & max              & 0.3              & 5.1e-04                                                          & -              \\ \hline
GraphSAGE + JK max                   & 79.54                                                                                    & 79.85                                                                               & 192                         & 4                               & max              & 0.1              & 4.6e-04                                                          & -              \\ \hline
GIN + JK max                         & 81.30                                                                                    & 65.15                                                                               & 192                         & 4                               & max              & 0.1              & 5.4e-04                                                          & -              \\ \hline
GAT + JK max                         & 79.81                                                                                    & 75.73                                                                               & 256                         & 5                               & mean             & 0.0              & 7.9e-04                                                          & 4              \\ \hline
GCN + JK concat                      & 76.48                                                                                    & 81.03                                                                               & 64                          & 5                               & max              & 0.0              & 5.5e-04                                                          & -              \\ \hline
GraphSAGE + JK concat                & 80.84                                                                                    & 80.93                                                                               & 128                         & 4                               & max              & 0.1              & 7.5e-04                                                          & -              \\ \hline
GIN + JK concat                      & 82.01                                                                                    & 65.61                                                                               & 192                         & 3                               & max              & 0.0              & 6.0e-04                                                          & -              \\ \hline
GAT + JK concat                      & 80.76                                                                                    & 67.76                                                                               & 256                         & 5                               & mean             & 0.0              & 4.9e-04                                                          & 8              \\ \hline
GCN + JK LSTM                        & 77.72                                                                                    & 81.08                                                                               & 128                         & 3                               & max              & 0.1              & 8.1e-04                                                          & -              \\ \hline
GraphSAGE + JK LSTM                  & 80.43                                                                                    & 80.39                                                                               & 64                          & 5                               & max              & 0.1              & 6.4e-04                                                          & -              \\ \hline
GIN + JK LSTM                        & 80.51                                                                                    & 70.34                                                                               & 128                         & 4                               & max              & 0.0              & 5.1e-04                                                          & -              \\ \hline
GAT + JK LSTM                        & 78.53                                                                                    & 76.09                                                                               & 192                         & 3                               & mean             & 0.0              & 6.6e-04                                                          & 2              \\ \hline
\end{tabular}
\caption{GNN models results, BACE dataset. Metric is AUROC - the higher, the better.}
\end{table}

\begin{table}[]
\centering
\begin{tabular}{|l|c|c|c|c|c|c|c|}
\hline
\multicolumn{1}{|c|}{\textbf{Model}}                                                         & \textbf{\begin{tabular}[c]{@{}c@{}}AUROC\\ validation\end{tabular}} & \textbf{\begin{tabular}[c]{@{}c@{}}AUROC \\ test\end{tabular}} & \textbf{\begin{tabular}[c]{@{}c@{}}Number\\ of bins\end{tabular}} & \textbf{Normalize} & \textbf{Class weight}                                         & \textbf{\begin{tabular}[c]{@{}c@{}}Minimal samples\\ for split\end{tabular}} & \textbf{\begin{tabular}[c]{@{}c@{}}Selected\\ features\end{tabular}}                                                                                \\ \hline
LDP                                                                                          & 95.65                                                               & 80.16                                                          & 10                                                                & False              & Balanced subsample & 2                                                                            & -                                                                                                                                                   \\ \hline
LDP extended                                                                                 & 95.73                                                               & 81.02                                                          & 70                                                                & True               & None                                                          & 2                                                                            & -                                                                                                                                                   \\ \hline
LDP with additional features                                                                 & 95.84                                                               & 81.43                                                          & 50                                                                & False              & Balanced subsample & 2                                                                            & -                                                                                                                                                   \\ \hline
\begin{tabular}[c]{@{}l@{}}LDP with additional features\\ and feature selection\end{tabular} & 95.91                                                               & 80.82                                                          & 20                                                                & False              & Balanced subsample & 2                                                                            & \begin{tabular}[c]{@{}c@{}}Clustering coefficient: true\\ Betweenness centrality: true\\ Closeness centrality: false\\ PageRank: false\end{tabular} \\ \hline
\end{tabular}
\caption{LDP models results, BACE dataset. Metric is AUROC - the higher, the better.}
\end{table}

\begin{table}[]
\centering
\begin{tabular}{|l|l|l|c|c|c|}
\hline
\multicolumn{1}{|c|}{\textbf{Model}} & \multicolumn{1}{c|}{\textbf{\begin{tabular}[c]{@{}c@{}}AUROC\\ validation\end{tabular}}} & \multicolumn{1}{c|}{\textbf{\begin{tabular}[c]{@{}c@{}}AUROC \\ test\end{tabular}}} & \textbf{\begin{tabular}[c]{@{}c@{}}Fingerprint\\ hyperparameters\end{tabular}}                                       & \textbf{Class weight} & \textbf{\begin{tabular}[c]{@{}c@{}}Minimal samples \\ for split\end{tabular}} \\ \hline
Morgan                               & 73.33                                                                                    & 86.26                                                                               & Radius: 2, bits per radius: 1024                                                                                     & Balanced              & 10                                                                            \\ \hline
RDKit                                & 71.13                                                                                    & 83.86                                                                               & Max path length: 5, number of bits: 2048                                                                             & None                  & 6                                                                             \\ \hline
MACCS                                & 71.33                                                                                    & 82.78                                                                               & -                                                                                                                    & Balanced              & 8                                                                             \\ \hline
Fingerprints concatenation                        & 73.15                                                                                    & 86.17                                                                               & \begin{tabular}[c]{@{}c@{}}Radius: 2, bits per radius: 2048,\\ max path length: 5, number of bits: 4096\end{tabular} & Balanced subsample    & 10                                                                            \\ \hline
\end{tabular}
\caption{Fingerprints models results, BACE dataset. Metric is AUROC - the higher, the better.}
\end{table}

\end{landscape}

\newpage

\subsection{BBBP dataset}

\begin{table}[h!]
\centering
\begin{tabularx}{\textwidth}{|X|c|c|c|c|}
\hline
\multicolumn{1}{|c|}{\textbf{Model}}                                               & \multicolumn{1}{c|}{\textbf{\begin{tabular}[c]{@{}c@{}}AUROC\\ validation\end{tabular}}} & \multicolumn{1}{c|}{\textbf{\begin{tabular}[c]{@{}c@{}}AUROC \\ test\end{tabular}}} & \textbf{\begin{tabular}[c]{@{}c@{}}MLP \\ layer size\end{tabular}} & \textbf{\begin{tabular}[c]{@{}c@{}}Learning\\ rate\end{tabular}} \\ \hline
Molecular Fingerprints                                                             & 93.05                                                                                    & 66.17                                                                               & 192                                                                & 6.6e-04                                                          \\ \hline
Molecular Fingerprints + atom embedding & 95.01                                                                                    & 68.19                                                                               & 64                                                                 & 9.6e-05                                                          \\ \hline
Deep Multisets                                                                     & 95.44                                                                                    & 62.86                                                                               & 128                                                                & 6.4e-04                                                          \\ \hline
Deep Multisets + atom embedding          & 96.69                                                                                    & 65.46                                                                               & 64                                                                 & 9.2e-04                                                          \\ \hline
\end{tabularx}
\caption{Baseline models results, BBBP dataset. Metric is AUROC - the higher, the better.}
\end{table}

\begin{landscape}
\begin{table}[]
\centering
\begin{tabular}{|l|c|c|c|c|c|c|c|c|}
\hline
\multicolumn{1}{|c|}{\textbf{Model}} & \textbf{\begin{tabular}[c]{@{}c@{}}AUROC\\ validation\end{tabular}} & \textbf{\begin{tabular}[c]{@{}c@{}}AUROC \\ test\end{tabular}} & \textbf{Number of channels} & \textbf{Number of conv. layers} & \textbf{Readout} & \textbf{Dropout} & \textbf{\begin{tabular}[c]{@{}c@{}}Learning\\ rate\end{tabular}} & \textbf{Heads} \\ \hline
GCN                                  & 94.29                                                                                    & 68.61                                                                               & 192                         & 3                               & mean             & 0.0              & 6.5e-04                                                          & -              \\ \hline
GraphSAGE                            & 94.01                                                                                    & 63.09                                                                               & 192                         & 3                               & mean             & 0.1              & 1.8e-04                                                          & -              \\ \hline
GIN                                  & 94.75                                                                                    & 62.58                                                                               & 256                         & 4                               & max              & 0.0              & 7.2e-04                                                          & -              \\ \hline
GAT                                  & 95.39                                                                                    & 67.76                                                                               & 64                          & 5                               & sum              & 0.2              & 8.8e-04                                                          & 6              \\ \hline
SGC                                  & 92.17                                                                                    & 66.21                                                                               & 256                         & 2                               & max              & 0.1              & 2.8e-04                                                          & -              \\ \hline
GCN + JK max                         & 95.07                                                                                    & 64.53                                                                               & 128                         & 5                               & sum              & 0.1              & 3.7e-05                                                          & -              \\ \hline
GraphSAGE + JK max                   & 95.35                                                                                    & 66.98                                                                               & 64                          & 3                               & mean             & 0.3              & 2.7e-04                                                          & -              \\ \hline
GIN + JK max                         & 95.43                                                                                    & 66.38                                                                               & 256                         & 5                               & max              & 0.0              & 7.3e-04                                                          & -              \\ \hline
GAT + JK max                         & 94.02                                                                                    & 68.49                                                                               & 64                          & 2                               & max              & 0.1              & 6.7e-04                                                          & 6              \\ \hline
GCN + JK concat                      & 92.54                                                                                    & 69.01                                                                               & 192                         & 3                               & mean             & 0.0              & 6.6e-04                                                          & -              \\ \hline
GraphSAGE + JK concat                & 95.31                                                                                    & 63.57                                                                               & 64                          & 4                               & mean             & 0.2              & 3.1e-04                                                          & -              \\ \hline
GIN + JK concat                      & 93.72                                                                                    & 67.89                                                                               & 128                         & 4                               & max              & 0.2              & 9.4e-04                                                          & -              \\ \hline
GAT + JK concat                      & 94.17                                                                                    & 66.67                                                                               & 256                         & 4                               & sum              & 0.3              & 2.9e-05                                                          & 2              \\ \hline
GCN + JK LSTM                        & 93.62                                                                                    & 65.99                                                                               & 128                         & 5                               & sum              & 0.1              & 8.9e-05                                                          & -              \\ \hline
GraphSAGE + JK LSTM                  & 95.04                                                                                    & 64.53                                                                               & 128                         & 2                               & max              & 0.2              & 3.7e-04                                                          & -              \\ \hline
GIN + JK LSTM                        & 94.20                                                                                    & 60.16                                                                               & 192                         & 2                               & max              & 0.2              & 5.1e-04                                                          & -              \\ \hline
GAT + JK LSTM                        & 94.20                                                                                    & 66.08                                                                               & 64                          & 5                               & sum              & 0.3              & 4.7e-04                                                          & 8              \\ \hline
\end{tabular}
\caption{GNN models results, BBBP dataset. Metric is AUROC - the higher, the better.}
\end{table}

\begin{table}[]
\centering
\begin{tabular}{|l|l|l|c|c|c|c|c|}
\hline
\multicolumn{1}{|c|}{\textbf{Model}}                                                         & \multicolumn{1}{c|}{\textbf{\begin{tabular}[c]{@{}c@{}}AUROC\\ validation\end{tabular}}} & \multicolumn{1}{c|}{\textbf{\begin{tabular}[c]{@{}c@{}}AUROC \\ test\end{tabular}}} & \textbf{\begin{tabular}[c]{@{}c@{}}Number\\ of bins\end{tabular}} & \textbf{Normalize} & \textbf{Class weight} & \textbf{\begin{tabular}[c]{@{}c@{}}Minimal samples\\ for split\end{tabular}} & \textbf{\begin{tabular}[c]{@{}c@{}}Selected\\ features\end{tabular}}                                                                                \\ \hline
LDP                                                                                          & 100.00                                                                                   & 62.77                                                                               & 40                                                                & False              & None                  & 4                                                                            & -                                                                                                                                                   \\ \hline
LDP extended                                                                                 & 100.00                                                                                   & 65.06                                                                               & 20                                                                & False              & None                  & 4                                                                            & -                                                                                                                                                   \\ \hline
LDP with additional features                                                                 & 100.00                                                                                   & 67.11                                                                               & 80                                                                & False              & None                  & 4                                                                            & -                                                                                                                                                   \\ \hline
\begin{tabular}[c]{@{}l@{}}LDP with additional features\\ and feature selection\end{tabular} & 100.00                                                                                   & 66.94                                                                               & 70                                                                & False              & None                  & 2                                                                            & \begin{tabular}[c]{@{}c@{}}Clustering coefficient: false\\ Betweenness centrality: true\\ Closeness centrality: false\\ PageRank: true\end{tabular} \\ \hline
\end{tabular}
\caption{LDP models results, BBBP dataset. Metric is AUROC - the higher, the better.}
\end{table}

\begin{table}[]
\centering
\begin{tabular}{|l|l|l|c|c|c|}
\hline
\multicolumn{1}{|c|}{\textbf{Model}} & \multicolumn{1}{c|}{\textbf{\begin{tabular}[c]{@{}c@{}}AUROC\\ validation\end{tabular}}} & \multicolumn{1}{c|}{\textbf{\begin{tabular}[c]{@{}c@{}}AUROC \\ test\end{tabular}}} & \textbf{\begin{tabular}[c]{@{}c@{}}Fingerprint\\ hyperparameters\end{tabular}}                                       & \textbf{Class weight} & \textbf{\begin{tabular}[c]{@{}c@{}}Minimal samples \\ for split\end{tabular}} \\ \hline
Morgan                               & 95.47                                                                                    & 67.16                                                                               & Radius: 3, bits per radius: 4096                                                                                     & None                  & 8                                                                             \\ \hline
RDKit                                & 95.25                                                                                    & 69.10                                                                               & Max path length: 7, number of bits: 4096                                                                             & None                  & 6                                                                             \\ \hline
MACCS                                & 95.88                                                                                    & 70.54                                                                               & -                                                                                                                    & None                  & 4                                                                             \\ \hline
Fingerprints concatenation                        & 96.45                                                                                    & 70.47                                                                               & \begin{tabular}[c]{@{}c@{}}Radius: 3, bits per radius: 2048,\\ max path length: 7, number of bits: 1024\end{tabular} & None                  & 10                                                                            \\ \hline
\end{tabular}
\caption{Fingerprints models results, BBBP dataset. Metric is AUROC - the higher, the better.}
\end{table}

\end{landscape}

\newpage

\subsection{Lipophilicity dataset}

\begin{table}[h!]
\centering
\centering
\begin{tabularx}{\textwidth}{|X|c|c|c|c|}
\hline
\multicolumn{1}{|c|}{\textbf{Model}}                                               & \multicolumn{1}{c|}{\textbf{\begin{tabular}[c]{@{}c@{}}AUROC\\ validation\end{tabular}}} & \multicolumn{1}{c|}{\textbf{\begin{tabular}[c]{@{}c@{}}AUROC \\ test\end{tabular}}} & \textbf{\begin{tabular}[c]{@{}c@{}}MLP \\ layer size\end{tabular}} & \textbf{\begin{tabular}[c]{@{}c@{}}Learning\\ rate\end{tabular}} \\ \hline
Molecular Fingerprints                                                             & 1.069                                                                                   & 1.022                                                                              & 320                                                                & 9.3e-04                                                          \\ \hline
Molecular Fingerprints + atom embedding & 0.982                                                                                   & 0.992                                                                              & 320                                                                & 9.3e-04                                                          \\ \hline
Deep Multisets                                                                     & 0.853                                                                                   & 0.848                                                                              & 128                                                                & 1.8e-05                                                          \\ \hline
Deep Multisets + atom embedding          & 0.843                                                                                   & 0.837                                                                              & 256                                                                & 9.9e-04                                                          \\ \hline
\end{tabularx}
\caption{Baseline models results, Lipophilicity dataset. Metric is RMSE - the lower, the better.}
\end{table}

\begin{landscape}
\begin{table}[]
\centering
\begin{tabular}{|l|c|c|c|c|c|c|c|c|}
\hline
\multicolumn{1}{|c|}{\textbf{Model}} & \multicolumn{1}{c|}{\textbf{\begin{tabular}[c]{@{}c@{}}RMSE\\ validation\end{tabular}}} & \multicolumn{1}{c|}{\textbf{\begin{tabular}[c]{@{}c@{}}RMSE \\ test\end{tabular}}} & \textbf{Number of channels} & \textbf{Number of conv. layers} & \textbf{Readout} & \textbf{Dropout} & \textbf{\begin{tabular}[c]{@{}c@{}}Learning\\ rate\end{tabular}} & \textbf{Heads} \\ \hline
GCN                                  & 0.714                                                                                   & 0.774                                                                              & 128                         & 5                               & sum              & 0.2              & 7.5e-04                                                          & -              \\ \hline
GraphSAGE                            & 0.715                                                                                   & 0.777                                                                              & 256                         & 5                               & sum              & 0.3              & 9.9e-04                                                          & -              \\ \hline
GIN                                  & 0.714                                                                                   & 0.779                                                                              & 128                         & 4                               & mean             & 0.1              & 4.9e-04                                                          & -              \\ \hline
GAT                                  & 0.740                                                                                   & 0.753                                                                              & 256                         & 4                               & mean             & 0.1              & 4.5e-04                                                          & 6              \\ \hline
SGC                                  & 0.895                                                                                   & 0.884                                                                              & 128                         & 4                               & max              & 0.1              & 2.1e-04                                                          & -              \\ \hline
GCN + JK max                         & 0.750                                                                                   & 0.763                                                                              & 128                         & 4                               & sum              & 0.1              & 7.9e-04                                                          & -              \\ \hline
GraphSAGE + JK max                   & 0.741                                                                                   & 0.776                                                                              & 256                         & 5                               & sum              & 0.2              & 7.0e-04                                                          & -              \\ \hline
GIN + JK max                         & 0.724                                                                                   & 0.747                                                                              & 128                         & 4                               & mean             & 0.2              & 2.6e-04                                                          & -              \\ \hline
GAT + JK max                         & 0.732                                                                                   & 0.734                                                                              & 256                         & 5                               & mean             & 0.1              & 8.9e-04                                                          & 6              \\ \hline
GCN + JK concat                      & 0.733                                                                                   & 0.715                                                                              & 128                         & 5                               & sum              & 0.1              & 6.4e-04                                                          & -              \\ \hline
GraphSAGE + JK concat                & 0.743                                                                                   & 0.732                                                                              & 192                         & 4                               & sum              & 0.1              & 8.7e-04                                                          & -              \\ \hline
GIN + JK concat                      & 0.726                                                                                   & 0.777                                                                              & 256                         & 3                               & sum              & 0.2              & 8.4e-04                                                          & -              \\ \hline
GAT + JK concat                      & 0.725                                                                                   & 0.808                                                                              & 256                         & 5                               & mean             & 0.0              & 9.9e-04                                                          & 8              \\ \hline
GCN + JK LSTM & \multicolumn{1}{c|}{0.750} & \multicolumn{1}{c|}{0.721} & 128 & 5 & sum & 0.1 & 7.4e-04 & - \\ \hline
GraphSAGE + JK LSTM & \multicolumn{1}{c|}{0.745} & \multicolumn{1}{c|}{0.726} & 128 & 5 & sum & 0.1 & 5.7e-04 & - \\ \hline
GIN + JK LSTM & \multicolumn{1}{c|}{0.697} & \multicolumn{1}{c|}{0.735} & 192 & 5 & sum & 0.2 & 8.6e-04 & - \\ \hline
GAT + JK LSTM & \multicolumn{1}{c|}{0.762} & \multicolumn{1}{c|}{0.791} & 256 & 5 & sum & 0.1 & 6.7e-04 & 6 \\ \hline
\end{tabular}
\caption{GNN models results, Lipophilicity dataset. Metric is RMSE - the lower, the better.}
\end{table}

\begin{table}[]
\centering
\begin{tabular}{|l|l|l|c|c|c|c|}
\hline
\multicolumn{1}{|c|}{\textbf{Model}}                                                         & \multicolumn{1}{c|}{\textbf{\begin{tabular}[c]{@{}c@{}}RMSE\\ validation\end{tabular}}} & \multicolumn{1}{c|}{\textbf{\begin{tabular}[c]{@{}c@{}}RMSE \\ test\end{tabular}}} & \textbf{\begin{tabular}[c]{@{}c@{}}Number\\ of bins\end{tabular}} & \textbf{Normalize} & \textbf{\begin{tabular}[c]{@{}c@{}}Minimal samples\\ for split\end{tabular}} & \textbf{\begin{tabular}[c]{@{}c@{}}Selected\\ features\end{tabular}}                                                                              \\ \hline
LDP                                                                                          & 0.456                                                                                   & 1.006                                                                              & 10                                                                & False              & 2                                                                            & -                                                                                                                                                 \\ \hline
LDP extended                                                                                 & 0.447                                                                                   & 0.966                                                                              & 10                                                                & True               & 2                                                                            & -                                                                                                                                                 \\ \hline
LDP with additional features                                                                 & 0.438                                                                                   & 0.965                                                                              & 10                                                                & True               & 2                                                                            & -                                                                                                                                                 \\ \hline
\begin{tabular}[c]{@{}l@{}}LDP with additional features\\ and feature selection\end{tabular} & 0.502                                                                                   & 0.966                                                                              & 10                                                                & True               & 8                                                                            & \begin{tabular}[c]{@{}c@{}}Clustering coefficient: true\\ Betweenness centrality: true\\ Closeness centrality: true\\ PageRank: true\end{tabular} \\ \hline
\end{tabular}
\caption{LDP models results, Lipophilicity dataset. Metric is RMSE - the lower, the better.}
\end{table}

\begin{table}[]
\centering
\begin{tabular}{|l|l|l|c|c|}
\hline
\multicolumn{1}{|c|}{\textbf{Model}} & \multicolumn{1}{c|}{\textbf{\begin{tabular}[c]{@{}c@{}}RMSE\\ validation\end{tabular}}} & \multicolumn{1}{c|}{\textbf{\begin{tabular}[c]{@{}c@{}}RMSE \\ test\end{tabular}}} & \textbf{\begin{tabular}[c]{@{}c@{}}Fingerprint\\ hyperparameters\end{tabular}}                                       & \textbf{\begin{tabular}[c]{@{}c@{}}Minimal samples \\ for split\end{tabular}} \\ \hline
Morgan                               & 0.928                                                                                   & 0.857                                                                              & Radius: 3, bits per radius: 2048                                                                                     & 2                                                                             \\ \hline
RDKit                                & 0.980                                                                                   & 0.935                                                                              & Max path length: 5, number of bits: 4096                                                                             & 6                                                                             \\ \hline
MACCS                                & 0.970                                                                                   & 0.883                                                                              & -                                                                                                                    & 4                                                                             \\ \hline
Fingerprints concatenation                        & 0.920                                                                                   & 0.820                                                                              & \begin{tabular}[c]{@{}c@{}}Radius: 3, bits per radius: 1024,\\ max path length: 5, number of bits: 4096\end{tabular} & 2                                                                             \\ \hline
\end{tabular}
\caption{Fingerprints models results, Lipophilicity dataset. Metric is RMSE - the lower, the better.}
\end{table}

\end{landscape}

\end{document}